\pdfoutput=1

\documentclass[11pt]{article}

\usepackage[preprint]{acl}

\usepackage{times}
\usepackage{latexsym}
\usepackage{amsmath}
\usepackage{amssymb}
\usepackage{mathtools}
\usepackage{booktabs}
\usepackage{subcaption}
\usepackage{tabularx}
\usepackage{multirow}
\usepackage{multicol}
\newcommand*{\Scale}[2][4]{\scalebox{#1}{$#2$}}%
\usepackage[T1]{fontenc}

\usepackage[utf8]{inputenc}

\usepackage{microtype}

\usepackage{inconsolata}

\usepackage{graphicx}
\usepackage{subcaption}
\usepackage{caption}
\usepackage{enumitem}
\usepackage[normalem]{ulem}

\title{Understanding LLMs' Cross-Lingual Context Retrieval:\\How Good It Is And Where It Comes From}

\newcommand*{\affaddr}[1]{#1} 
\newcommand*{\affmark}[1][*]{\textsuperscript{#1}}

\author{
\textbf{Changjiang Gao}\affmark[$\clubsuit$], \textbf{Hankun Lin}\affmark[$\clubsuit$], \textbf{Xin Huang}\affmark[$\Diamond$], \\
\textbf{Xue Han}\affmark[$\Diamond$], \textbf{Junlan Feng}\affmark[$\Diamond$], \textbf{Chao Deng}\affmark[$\Diamond$], \textbf{Jiajun Chen}\affmark[$\clubsuit$], \textbf{Shujian Huang}\affmark[$\clubsuit$]\thanks{Corresponding author} \\
\affaddr{\affmark[$\clubsuit$]National Key Laboratory for Novel Software Technology, Nanjing University} \\
\affaddr{\affmark[$\Diamond$]China Mobile Research Beijing, China} \\
\texttt{\{gaocj,linhk\}@smail.nju.edu.cn, \{chenjj,huangsj\}@nju.edu.cn} \\
\texttt{\{huangxinyjy,hanxueai,fengjunlan,dengchao\}@chinamobile.com}
}

\begin{document}
\maketitle
\begin{abstract}
Cross-lingual context retrieval (extracting contextual information in one language based on requests in another) is a fundamental aspect of cross-lingual alignment, but the performance and mechanism of it for large language models (LLMs) remains unclear. In this paper, we evaluate the cross-lingual context retrieval of over 40 LLMs across 12 languages, using cross-lingual machine reading comprehension (xMRC) as a representative scenario. Our results show that post-trained open LLMs show strong cross-lingual context retrieval ability, comparable to closed-source LLMs such as GPT-4o, and their estimated oracle performances greatly improve after post-training. Our mechanism analysis shows that the cross-lingual context retrieval process can be divided into two main phases: question encoding and answer retrieval, which are formed in pre-training and post-training respectively. The phasing stability correlates with xMRC performance, and the xMRC bottleneck lies at the last model layers in the second phase, where the effect of post-training can be evidently observed. Our results also indicate that larger-scale pretraining cannot improve the xMRC performance. Instead, larger LLMs need further multilingual post-training to fully unlock their cross-lingual context retrieval potential.
\footnote{Our code and results are available at \url{https://github.com/NJUNLP/Cross-Lingual-Context-Retrieval}}

\end{abstract}
\begin{figure}[ht]
    \centering
    \begin{subfigure}{0.45\textwidth}
        \includegraphics[width=\textwidth]{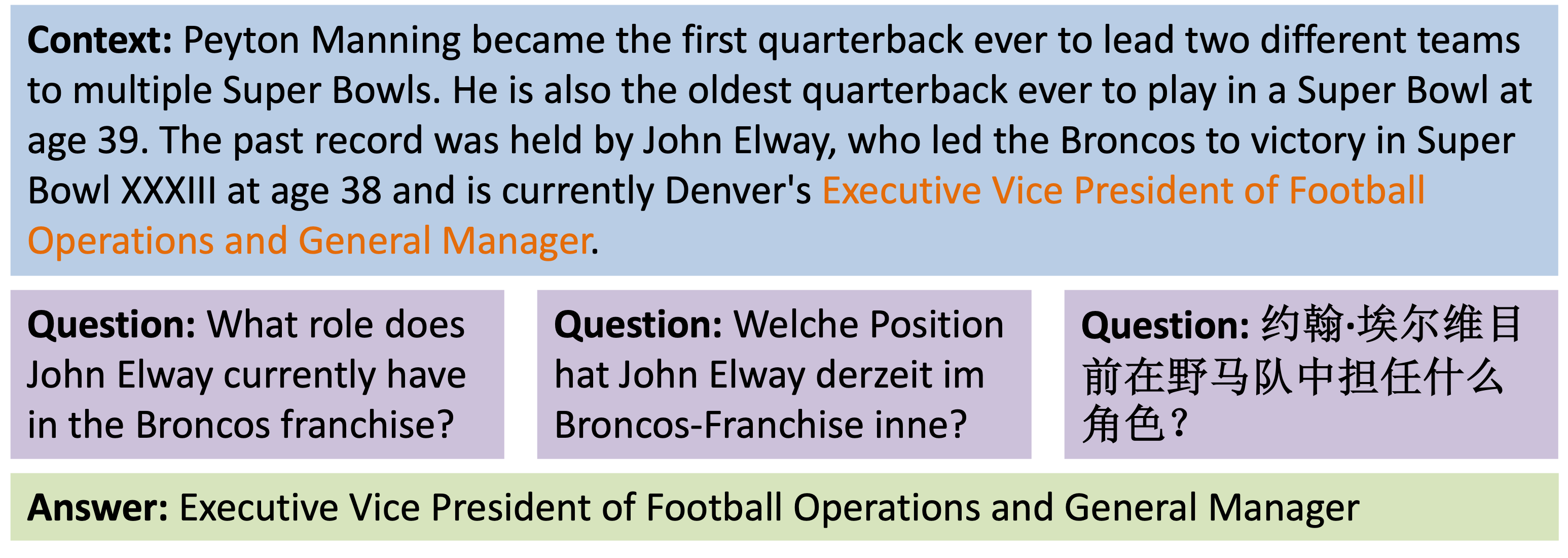}
        \caption{en-x} 
        \label{en-x_samples}
    \end{subfigure}
    \hfill
    \begin{subfigure}{0.45\textwidth}
        \includegraphics[width=\textwidth]{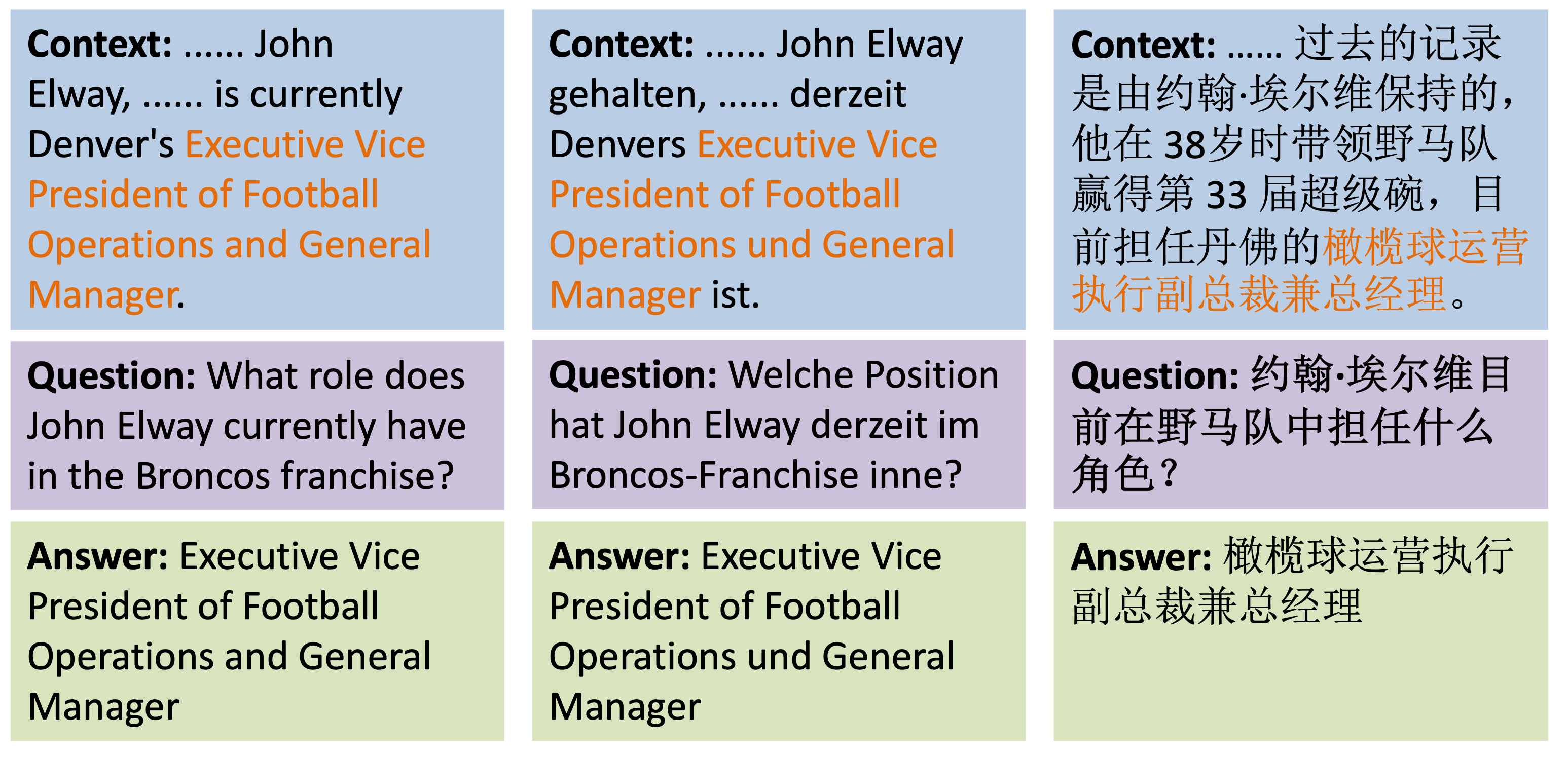}
        \caption{x-x} 
        \label{x-x_samples}
    \end{subfigure}
    \caption{Examples of our en-x and x-x testing scenarios. The figures show examples in English (en), German (de), and Chinese (zh).}
    \label{scenarios}
\end{figure}

\section{Introduction}
Since the rise of Large language models (LLMs), many models have demonstrated their strong capability in various NLP tasks \cite{changSurveyEvaluationLarge2024}, e.g. ChatGPT\footnote{\url{https://chatgpt.com}}, Claude\footnote{\url{https://claude.ai}}, Gemini \cite{geminiteam2024geminifamilyhighlycapable}, LLaMA \cite{grattafiori2024llama3herdmodels}, Qwen \cite{qwen2025qwen25technicalreport}, DeepSeek \cite{deepseekai2024deepseekv3technicalreport}, etc. 
However, due to the dominance of English training data, most of these LLMs show their best performance in English \cite{laiChatGPTEnglishComprehensive2023}. 
To improve their performance and efficiency in non-English languages, cross-lingual alignment has become a major research topic for multilingual LLMs \cite{qiCrossLingualConsistencyFactual2023,wang-etal-2024-seaeval}, which encourages LLMs to share capabilities across languages.
For example, given requests with the same semantics but in different languages, LLMs should give consistent answers.

Such alignment has been shown challenging when the task requires recalling trained knowledge \cite{gao-etal-2024-multilingual,hu2025largelanguagemodelscrosslingual}. Thus, a follow-up question is that, when the knowledge is presented in the context in one language (e.g. English), could LLMs retrieve it when answering requests in the another language? Note that this is different from multilingual context retrieval (the context and the request are in the same language) and cross-lingual information retrieval (retrieving queried text from a database). However, evaluation and mechanism analysis of this ability in LLMs are not fully explored.


In this paper, we evaluate cross-lingual context retrieval of SOTA multilingual LLMs, and analyze the mechanism of it. We use cross-lingual machine reading comprehension (xMRC) \cite{cui-etal-2019-cross} as a simplified but representative scenario because: (1) It is a representative NLP task where models need to retrieve cross-lingual knowledge within the context; (2) The target knowledge to be retrieved is entirely within the context, so minimal extra knowledge recall is needed; (3) The model only has to copy part of the context as the answer, simplifying evaluation and mitigating other factors such as low-resource language generation errors.

Furthermore, we conduct in-depth analysis of the performance bottleneck, oracle performance, and mechanism of LLMs doing the xMRC task, using suitable tools and specially designed metrics. Specifically, we propose a hypothesis of two-phased xMRC process in LLMs, and verify our hypothesis with experimental evidence.

Our main findings are:
\begin{itemize} [nosep,itemsep=2pt,leftmargin=0.2cm]
    \item Post-trained open-source LLMs, especially 7-9B versions, show strong xMRC ability, catching up with closed-source models. Larger models show higher English MRC performance, but larger gap between English and non-English.
    \item Post-training significantly improves the estimated oracle performance to almost saturate in all tested languages, setting space for improvement of real performance, while the effect of extended pretraining is minor.
    \item The xMRC process can be divided into two phases within the model: question encoding and answer retrieval. The former forms in pre-training, and its stability correlates with the base model capability, as well as xMRC performance; the latter forms in post-training, and serves as a bottleneck.
    \item One reason for the larger language gap observed on larger post-trained open-source LLMs might be insufficient and language-biased post training.
\end{itemize}

\section{Methods}
\subsection{Evaluation methods}
\subsubsection{Scenarios}
We use English to non-English (en-x) as a typical and common cross-lingual scenario, where the context and answer are in English, and the question is in non-English, because it avoids the requirement of non-English generation fluency that is not related to cross-lingual context retrieval, and keeps the answers always in English which makes answer verification easier and fairer.
Meanwhile, non-English monolingual (x-x) is used as comparison to ablate the effect of non-English understanding. Figure \ref{scenarios} shows examples of the testing scenarios.




\subsubsection{Metrics}
\paragraph{Performance metrics.} We use the F1 score to evaluate xMRC performance, as adopted by XQuAD~\cite{artetxe-etal-2020-cross}. Exact match (EM) is not used because it tends to under-estimate the performance due to unnecessary source text in the answers (see Appendix \ref{appendix:example_unnecessary}).
\paragraph{Cross-lingual performance metrics.} To measure performance alignment between English and non-English, we calculate the average performance on all the non-English languages divided by English (en-x/en-en, x-x/en-en). 

\subsubsection{Performance bottleneck analysis}
\paragraph{Error type ablation.} 
Based on observation of model responses, we distinguish error types as: 
\begin{itemize} [nosep,itemsep=2pt,leftmargin=0.2cm]
    \item \textbf{Language}: answering in \texttt{x} instead of \texttt{en};
    \item \textbf{Generation}: generation errors including gibberish, refusal and blank answers;
    \item \textbf{Content}: meaningful but incorrect answers in the correct language;
\end{itemize}

For example, the language error rate for test setting \texttt{en-x} is calculated as:
\[E_\mathrm{lang}(\mathrm{en, x})=\sum_{\mathclap{(r,a)\in W_{\mathrm{en,x}}}}\frac{\Scale[0.9]{\mathbb I(\mathrm{Lang}(r)=\mathrm{en})\mathbb I(\mathrm{Lang}(a)=\mathrm{x})}}{|W_{\mathrm{en,x}}|}\]

where $r$ is the reference answer, $a$ is the model prediction, and $\mathrm{Lang}(\cdot)$ is a language detector for given text. Meanwhile, a generation error rate, e.g. gibberish, will be:
\[E_\mathrm{gib}(\mathrm{en, x})=\sum_{\mathclap{(r,a)\in W_{\mathrm{en,x}}}}\frac{\Scale[0.9]{\mathbb I(\mathrm{Type}(r)=\mathrm{Gibberish})}}{|W_{\mathrm{en,x}}|}\]

where $\mathrm{Type}(\cdot)$ is an LLM-based error type classifier for generation errors.

\paragraph{Oracle performance estimation}
\label{methods-upper-bound}
The xMRC score can be affected by errors in the generation process, under-estimating LLMs' cross-lingual context retrieval ability. To ablate this effect, we estimate the oracle retrieval performance of LLMs by perturbation-based attribution\footnote{We use Captum to perform the attribution (\url{https://github.com/pytorch/captum})} on contextual sentences or spans. 
If the sentence/span with the correct answer receives the largest attribution score, we consider the model's oracle retrieval on this testing sample correct.

\subsection{Mechanism analysis methods}

\subsubsection{Layer-wise attribution to reflect forward process}
\label{sec:layerwise-attribution}
To better understand the forward process of LLMs performing xMRC, we need layer-wise attribution to observe information flow from context parts into the residual stream \cite{voita-etal-2024-neurons} at each layer.
Here we adopt AttentionLRP \cite{achtibatAttnLRPAttentionAwareLayerWise2024}, an attribution method based on Taylor Decomposition on attention, FFN and normalization modules of LLMs, that can calculate the relevance of token representations in each layer to the output. 

Based on the attribution results, we define the \textbf{Major Relevance Depth (MRD)} to estimate the maximum depth to which a token representation $x$ needs to be encoded, by calculating the layer number corresponding to the 95th percentile of its attributed relevance to model $m$'s output:
\begin{align*}
    \text{MRD}(m, x)&=\min_{1\leqslant n\leqslant N} n\\
    \text{s.t.} \sum_{i=1}^nr_{\text{out}}(x,i)&\geqslant 0.95\sum_{i=1}^{N}r_{\text{out}}(x,i)
\end{align*}
where $r_{\text{out}}(x,i)$ refers to the normalized relevance score of token representation $x$ in layer $i$ to the final output given by AttentionLRP.
A token-MRD of $\hat n$ indicates that the token information participates in the context retrieval process only in the first $\hat n$-th layers.
Then, for parts of the input, i.e., task description, demonstrations, question and context, we take the maximum token-MRD of each to represent them, and calculate the mean part-MRD.

\subsubsection{Hidden state similarity to measure cross-lingual alignment}
To observe the cross-lingual alignment of internal xMRC process, we collect the hidden states in all model layers and calculate their cross-lingual similarity, and define a cross-lingual similarity ratio $S(\text{en,x})$ between English and language x:
\begin{align*}
    S(\text{en,x})&=\overline{\text{Sim}}(E,X)/\overline{\text{Sim}}(X, X) \\
    &=\frac{(K-1)\sum_{e_k\in E, x_k\in X}{\text{Sim}(e_k, x_k)}}{2\sum_{x_i, x_j\in X}{\text{Sim}(x_i,x_j)}}
\end{align*}
where $\overline{\text{Sim}}$ denotes the mean cosine similarity, $\text{Sim}(x,y)=x\cdot y/|x||y|$. $E$ and $X$ denotes en-en and en-x hidden states. $K$ is the total number of samples, $e_k$ and $x_k$ are the hidden states from the $k$-th parallel sample pair between English and language x, $x_i,x_j$ are the hidden states from every two different samples in language x.



\begin{table*}[ht]
    \scriptsize
    \centering
    \setlength{\tabcolsep}{3pt} 
\begin{tabular}{ccccccccc}
\toprule
\multirow{2}{*}{\textbf{}} & \multicolumn{5}{c}{\textbf{F1 scores}} & \multicolumn{3}{c}{\textbf{Error rates}} \\ \cline{2-9} 
 & \textbf{en-en} & \textbf{mean en-x} & \textbf{mean x-x} & \textbf{en-x / en-en} & \textbf{x-x / en-en} & \textbf{mean language} & \textbf{mean generation} & \textbf{en-en generation} \\ \hline
\textbf{LLaMA-3.1-8B} & 75.97 & 49.01 & 70.28 & 0.64 & 0.93 & 0.32 & 8.88 & 5.60 \\
\textbf{LLaMA-3.1-70B} & 82.39 & 58.68 & 74.73 & 0.71 & 0.91 & 60.21 & 2.63 & 1.20 \\
\textbf{Mistral-V0.3-7B} & 79.57 & 58.74 & 64.92 & 0.74 & 0.82 & 21.24 & 14.25 & 0.49 \\
\textbf{Qwen-2.5-7B} & 62.42 & 57.51 & 66.11 & 0.92 & 1.06 & 0.96 & 3.29 & 1.51 \\
\textbf{Qwen-2.5-72B} & 86.03 & 78.92 & 81.16 & 0.92 & 0.94 & 10.90 & 2.53 & 0.00 \\
\textbf{Gemma-2-9B} & 80.42 & 66.82 & 72.90 & 0.83 & 0.91 & 1.91 & 4.11 & 1.02 \\
\textbf{DeepSeek-V2-Lite-16B} & 73.81 & 44.65 & 57.66 & 0.61 & 0.78 & 12.97 & 8.45 & 1.87 \\ \midrule
\textbf{LLaMA-3.1-Instruct-8B} & 77.89 & 72.13 & 65.02 & 0.93 & 0.83 & 0.89 & 2.53 & 0.85 \\
\textbf{LLaMA-3.1-Tuned-8B} & 78.80 & 70.80 & 66.84 & 0.90 & 0.85 & 0.80 & 3.28 & 0.87 \\
\textbf{LLaMA-3.1-Instruct-70B} & 83.29 & 73.07 & 74.13 & 0.88 & 0.89 & 0.23 & 1.87 & 1.85 \\
\textbf{Mistral-V0.3-Instruct-7B} & 62.01 & 56.63 & 49.39 & 0.91 & 0.80 & 2.77 & 3.30 & 1.77 \\
\textbf{Qwen-2.5-Instruct-7B} & 81.83 & 76.43 & 71.61 & 0.93 & 0.88 & 0.67 & 3.21 & 2.75 \\
\textbf{Qwen-2.5-Instruct-72B} & 77.12 & 66.04 & 70.29 & 0.86 & 0.91 & 4.58 & 1.62 & 0.38 \\
\textbf{Gemma-2-IT-9B} & 83.69 & 78.72 & 75.53 & 0.94 & 0.90 & 0.17 & 2.47 & 1.95 \\
\textbf{DeepSeek-V2-Chat-Lite-16B} & 70.30 & 54.03 & 49.95 & 0.77 & 0.71 & 2.36 & 5.92 & 0.58 \\
\textbf{DeepSeek-V3} & 82.21 & 78.55 & 76.80 & 0.96 & 0.93 & 0.18 & 1.60 & 0.00 \\
\textbf{GPT-3.5-Turbo-0125} & 81.74 & 68.75 & 72.04 & 0.84 & 0.88 & 0.16 & 2.80 & 0.00 \\
\textbf{GPT-4o} & 83.29 & 78.76 & 75.68 & 0.95 & 0.91 & 0.10 & 1.40 & 0.00 \\ \bottomrule
\end{tabular}
    \caption{2-shot F1 scores on en-x and x-x tasks, and 2-shot language error and generation error rates (\%) on en-x tasks.}
    \label{combined_table}
\end{table*}

\section{Experiment Settings}
\subsection{Dataset}
We use the XQuAD dataset~\cite{artetxe-etal-2020-cross} to measure the xMRC performance of LLMs, because its testing samples are parallel in all the 12 included languages\footnote{en, de, es, vi, zh, hi, ar, el, ro, ru, th, tr} and thus suitable for cross-lingual transforming. The dataset has 1190 parallel samples for each language and an average context length of 702.50 words. It also covers diverse language families, scripts, and resource levels.

\subsection{Models and tools}
We adopt a variety of SOTA open and business LLMs, including LLaMA-3.1~\cite{grattafiori2024llama3herdmodels}, Mistral~\cite{jiang2023mistral7b}, Qwen-2.5~\cite{qwen2025qwen25technicalreport}, Gemma-2~\cite{gemmateam2024gemma2improvingopen}, DeepSeek V2\&3~\cite{deepseekai2024deepseekv2strongeconomicalefficient,deepseekai2024deepseekv3technicalreport}, GPT-3.5, and GPT-4o, in smaller and larger sizes.
Table \ref{all_models_appendix} in Appendix \ref{appendix:model_lists} shows a full list of all the tested models.
We also tune the LLaMA-3.1-8B model with the TULU-v3 dataset \cite{lambert2025tulu3pushingfrontiers} into a model called LLaMA-3.1-Tuned-8B (Appendix \ref{appendix:llama-tuned-params} for details) to verify the effect of post-training.

For language error detection, we use Lingua\footnote{\url{https://github.com/pemistahl/lingua}} with its high-accuracy mode, the accuracy of which is satisfactory in our tested languages (see Appendix \ref{appendix:lingual_acc}). For generation errors detection, we use Qwen-2.5-72B-Instruct (prompt shown in Appendix \ref{appendix:error-type-prompt}) to identify generation errors. To rule out the potential bias induced by Qwen judging itself, we adopt Gemini-2.5-Flash-Lite \footnote{\url{https://gemini.google.com}} as another judge for cross-validation (see Appendix \ref{appendix:detailed_generate_failure_error}).

\subsection{Prompts}
\label{prompt_templates}
In our xMRC evaluation, we try two different prompt templates and use the one with higher performance for each model (see Appendix \ref{appendix:error-type-prompt}).
Our main evaluation and analysis uses 2-shot for higher performance, and 0-shot results can be found in Appendix \ref{appendix:0-shot_results}.

\section{Results}
\subsection{Evaluation results}
Table \ref{combined_table} summarizes the en-x and x-x MRC performances of the main-list models (see more results in Appendix \ref{appendix:0-shot_results}).

\subsubsection{Cross-lingual performance}
Generally, the English MRC performance of most models are high (over 70 out of 100), but the en-x scores ranges (from 45 to 78), showing the \textbf{performance gap in context retrieval with English and non-English queries}. Down into individual models, while GPT-4o shows the highest cross-lingual performance and smallest language gap, several post-trained open LLMs, such as Gemma-2-IT, Qwen-2.5-Instruct and LLaMA-3.1-Instruct, show performance levels and small language gaps comparable to the commercial models.

An interesting observation is that,
for LLaMA-3.1 and Gemma-2, the en-en performances remains close after post-training, but en-x greatly improve. This phenomenon is \textbf{more prominent in smaller (7-9B) than larger models}, bring the former a smaller performance gap between English and non-English,
which is also observed on Qwen2.5 with growing parameter sizes (See Appendix \ref{appendix:increase_size}). 

\subsubsection{Comparison with Monolingual performance}
In general, the performance gaps between en-en and x-x are much smaller for most models than en-x, and the Qwen models even show higher non-English performance than English. This suggests that \textbf{non-English language fluency is not the main challenge of xMRC}.

Also, for base models and post-trained larger models ($\sim$70B), \textbf{the x-x performances are always higher than en-x}. A possible explanation to this may be the cross-lingual task is less frequent in training, and more difficult because it requires cross-lingual understanding.

However, for post-trained, smaller models (7-9B), the pattern flips, where x-x performances become consistently lower than en-x, suggesting that \textbf{these models use their English context processing ability to assist non-English retrieval}, overcoming the difficulty and low-frequency of the cross-lingual task. Also, since larger LLMs tend to be better at instruction following and understanding, this further highlights that post-training better elicits the cross-lingual context retrieval ability on smaller LLMs.

\subsection{Performance bottleneck analysis}
\subsubsection{Error type ablation}
The right part of Table \ref{combined_table} shows error rates of different types (see details in Appendix \ref{appendix:detailed_generate_failure_error}).

\paragraph{Language and generation errors.} The language error rates of post-trained models (lower part of the table) are significantly lower than base models (upper part of the table), since the former can better follow the cross-lingual task format. Meanwhile, the error rate is low for all the post-trained models, so it cannot be viewed as a bottleneck of xMRC. The generation error rates are minor for most models, regardless of size and post-training, marking it not the bottleneck of xMRC either.


\subsubsection{Estimated oracle performance}
Table \ref{tab:estimated-oracle-combined} shows the estimated oracle performances of the LLaMA-3.1 models. ``Step'' means the attribution is done when generating the first answer token, and ``Sequence'' means the attribution is done on whole-sequence generation until EOS.

First, \textbf{the estimated oracle performances of post-trained models are significantly higher than the base models}, both for en-en and en-x, suggesting the importance of post-training to improving xMRC. However, 70B models show no substantial advantage over 8B, indicating extended pretraining contributes less to xMRC.

Second, the estimated oracle performance for en-x is close to en-en for all the LLaMA-3.1 models, and the oracle for post-trained models are basically over 90\%. This is much higher than the actual performance, suggesting the \textbf{models have the potential to locate correct answers with high accuracy, but the ability needs to be further elicited}.

\begin{table}[t]
\centering
\scriptsize
\begin{tabular}{ccccc}
\toprule
\textbf{Model}                  & \multicolumn{2}{c}{\textbf{Step}} & \multicolumn{2}{c}{\textbf{Sequence}}\\ \midrule
\textbf{Language}               & en-en               & en-x     & en-en               & en-x       \\
\textbf{LLaMA-3.1-8B}      & 66.55            & 67.59      & 35.14           & 36.39                 \\
\textbf{LLaMA-3.1-70B}     & 47.47            & 54.92      & 47.89           & 56.97                \\ \hline
\textbf{LLaMA-3.1-Instruct-8B}  & 89.86            & 83.42   & 93.75           & 86.66                     \\
\textbf{LLaMA-3.1-Tuned-8B}     & 86.49            & 81.04   & 89.53           & 83.07                   \\
\textbf{LLaMA-3.1-Instruct-70B} & 92.82            & 90.67   & 95.44           & 93.54                      \\ \bottomrule
\end{tabular}
\caption{Oracle performance estimated for LLaMA models in en-en and en-x (average) scenarios. The estimation is performed with one generation step (left) and with the whole generated sequence, respectively.}
\label{tab:estimated-oracle-combined}
\end{table}

\begin{figure}[t!]
    \centering
    \includegraphics[width=0.95\linewidth]{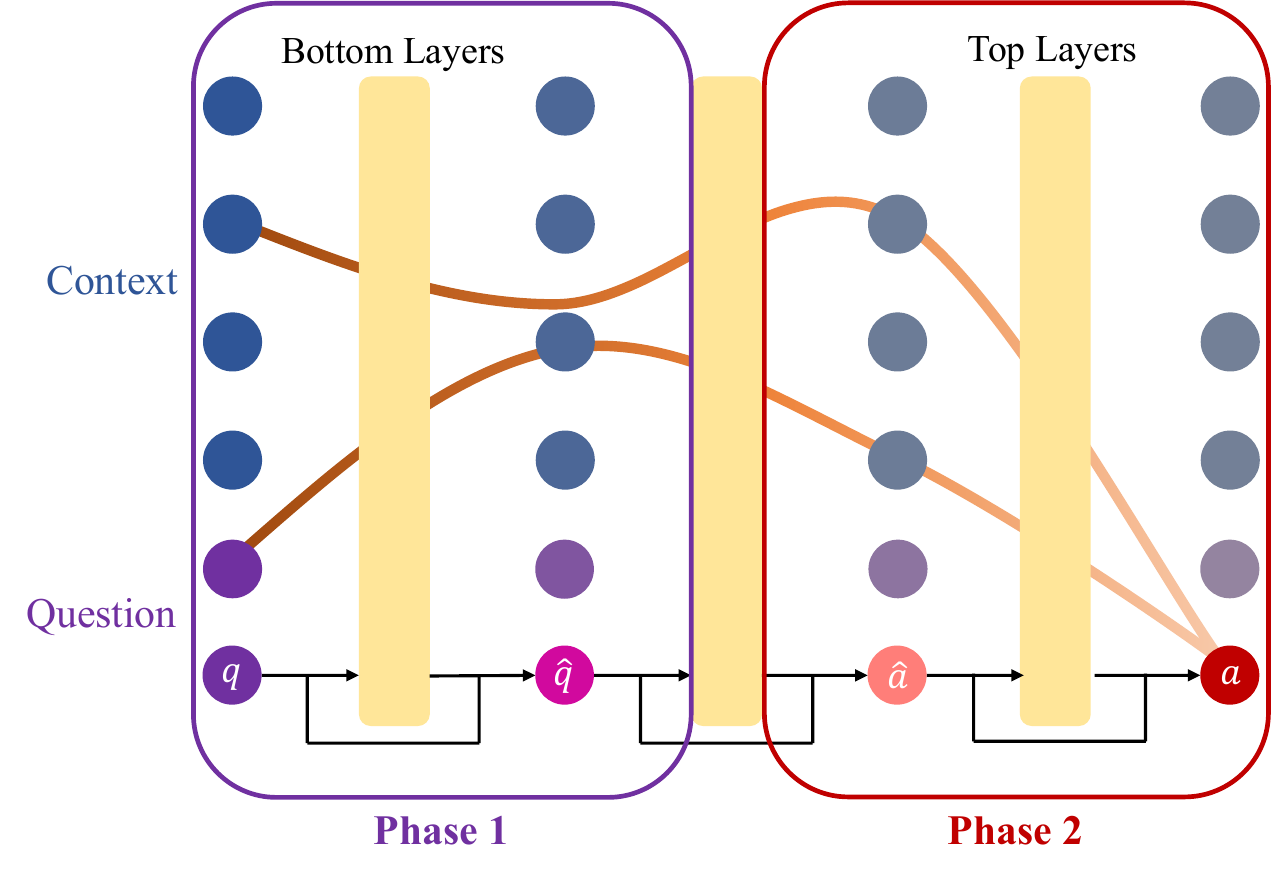}
    \caption{Illustration of the hypothesized two-phased xMRC process. Through layers, the last question token will be transferred to the first answer token in two phases, between which is a cross-lingual question representation.}
    \label{fig:phase-illustrate}
\end{figure}

\begin{figure}[ht]
    \centering
    \begin{subfigure}{0.45\textwidth}
        \includegraphics[width=\textwidth]{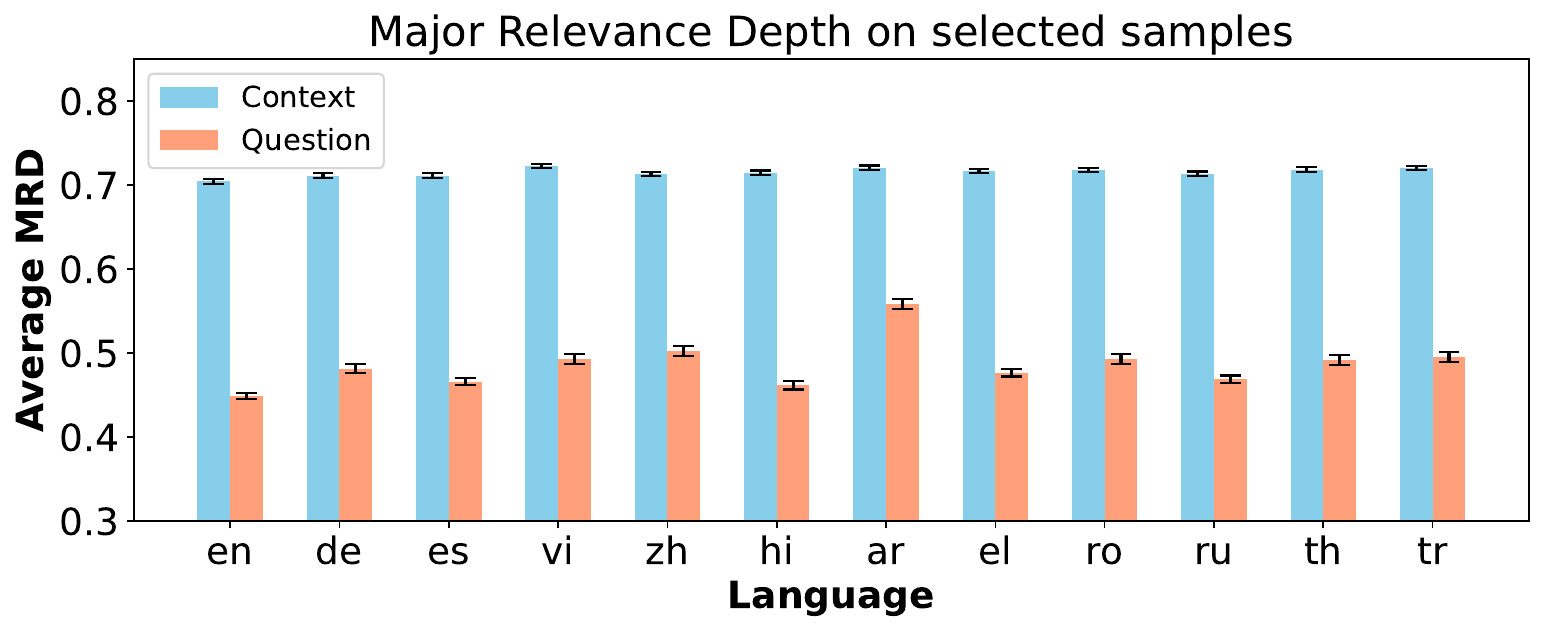}
        \caption{balanced samples}
        \label{llama-3.1-instruct-8b-balanced-cq}
    \end{subfigure}
    \hfill
    \begin{subfigure}{0.45\textwidth}
        \includegraphics[width=\textwidth]{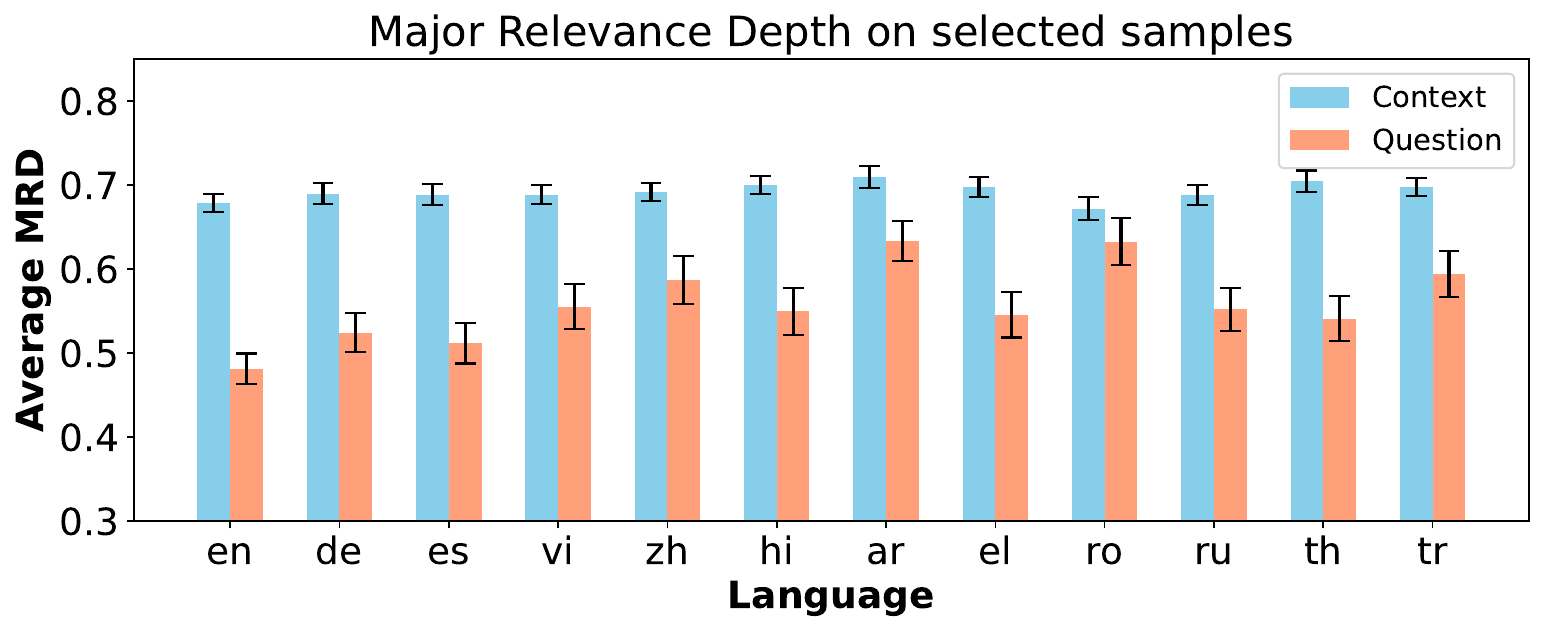}
        \caption{en-superior samples}
        \label{llama-3.1-instruct-8b-en-sup-cq}
    \end{subfigure}
    \caption{Mean MRD of the context and question parts for LLaMA-3.1-Instruct-8B.}
    \label{llama-3.1-instruct-8b-cq}
\end{figure}

\section{Two-phased mechanism of xMRC}
Besides evaluation of xMRC performance and bottlenecks, the mechanism of how LLMs do xMRC also worth exploring. Considering the model forward process from the input prompt to the output answer, we come up with a two-phase hypothesis of the xMRC process (taking en-x as an example):
\begin{enumerate} [nosep,itemsep=2pt,leftmargin=0.2cm]
    \item \textbf{Question encoding.} The non-English queries will be encoded into a shared semantic space, where queries in different languages are aligned and understood in a language-neutral way;
    \item \textbf{Answer retrieval.} The encoded queries will be used to match the answer in the English context according to the task description and format, then the answer is generated by copying from the original context.
\end{enumerate}

Figure \ref{fig:phase-illustrate} shows an illustration of the hypothesized process, which aligns with previous studies \cite{tang-etal-2024-language,wendler-etal-2024-llamas,zhaoHowLargeLanguage2024c}. If the hypothesis holds, then we will be able to extract cross-lingual representations in the middle layers and steer them to control the model retrieval behaviors. We test our hypothesis from attribution and hidden-state views with LLaMA-3.1-Instruct-8B, which shows high performance alignment across languages and is widely used. To further ensure the effect of post-training, we also conduct finetuning and compare the model behavior before and after it.

\subsection{Evidence from attribution view}

Figure \ref{llama-3.1-instruct-8b-cq} shows the mean MRD (\S\ref{sec:layerwise-attribution}) of the contexts and the questions for the LLaMA-3.1-Instruct-8B on testing samples that are identified as either ``balanced'' or ``en-superior'' in all tested languages. We identify a sample as ``balanced'' if the model F1 score on it is above 0.5 in all directions; and a sample as ``en-superior'' if the F1 score in English is higher than the average of other languages with a margin greater than 0.5. 

We find that the mean question MRD is significantly and substantially lower than the mean context MRD in all tested languages and across the LLaMA models, especially for the ``balanced'' samples (Figure \ref{llama-3.1-instruct-8b-cq}), revealing a clear phased behavior.

\begin{figure}[t!]
    \centering
    \begin{subfigure}{0.45\textwidth}
        \includegraphics[width=\textwidth]{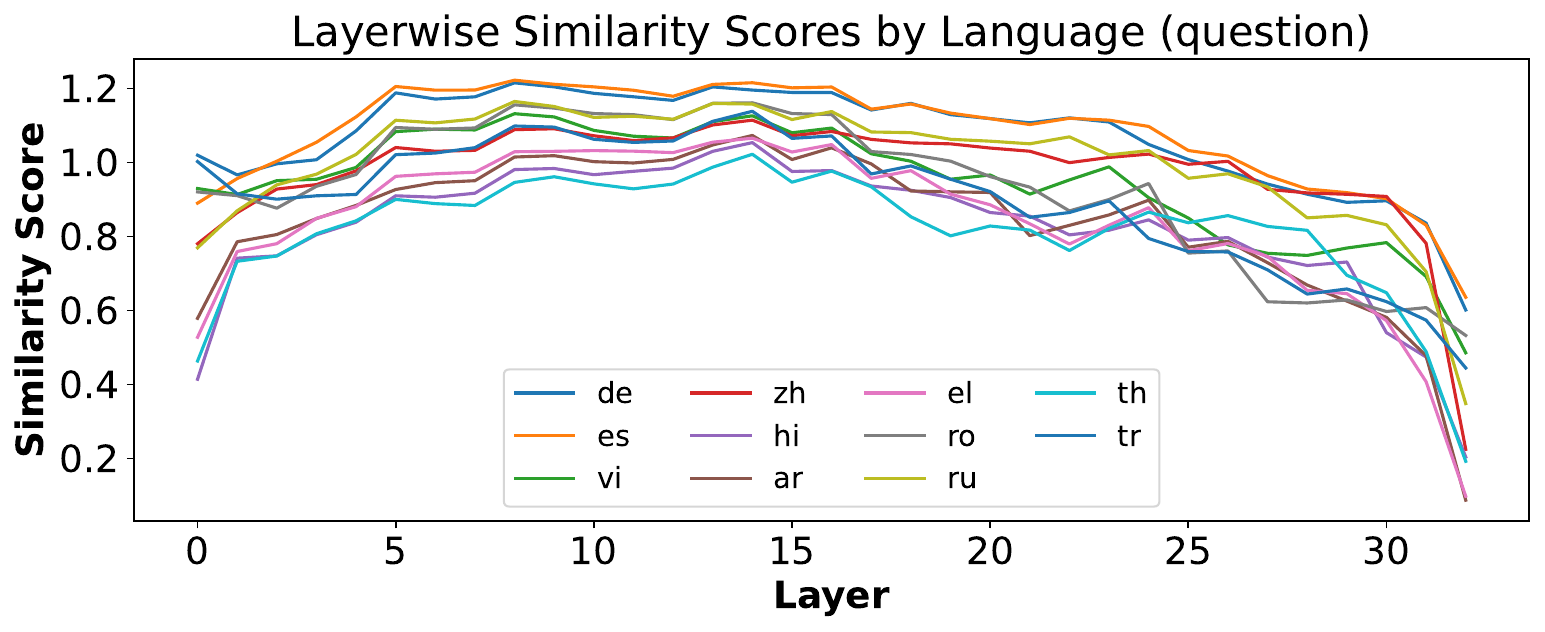}
        \caption{Question}
        \label{llama-3.1-instruct-8b-2-shot-balanced-q}
    \end{subfigure}
    \hfill
    \begin{subfigure}{0.45\textwidth}
        \includegraphics[width=\textwidth]{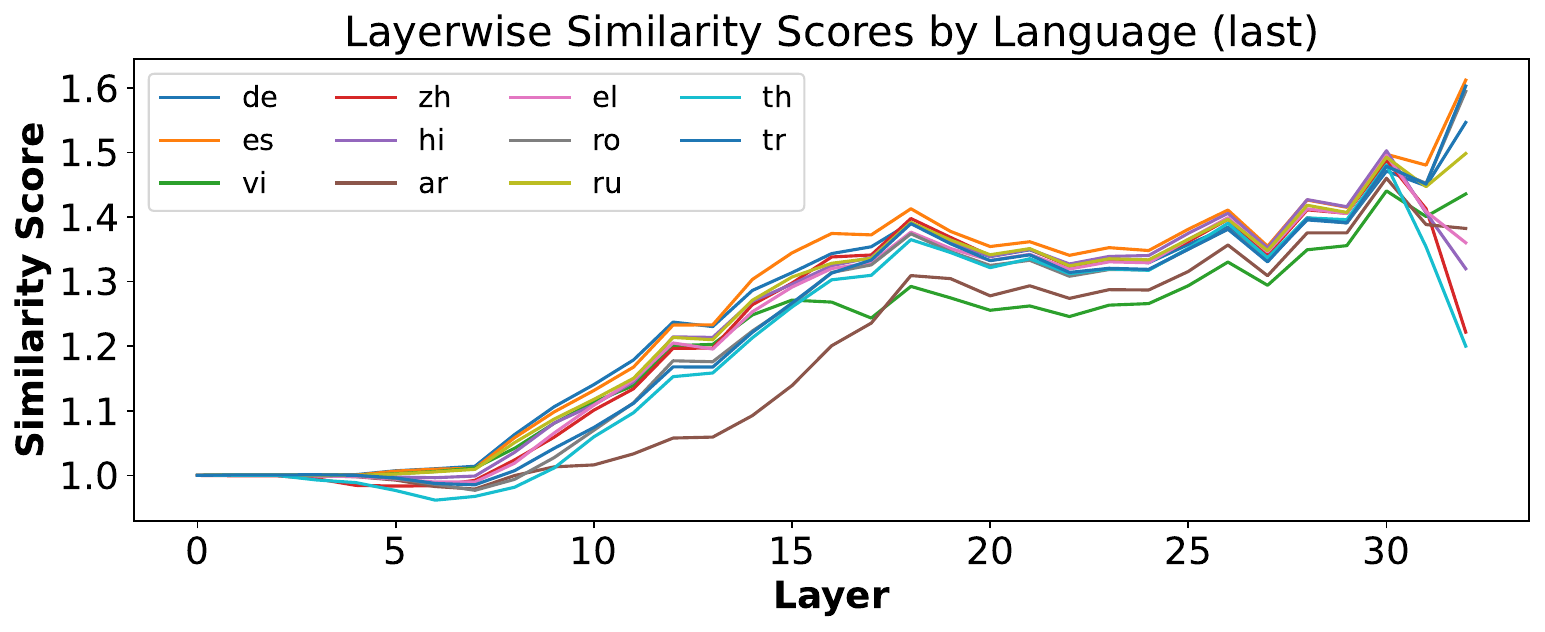}
        \caption{Last token}
        \label{llama-3.1-instruct-8b-2-shot-balanced-last}
    \end{subfigure}
    \hfill
    \begin{subfigure}{0.45\textwidth}
        \includegraphics[width=\textwidth]{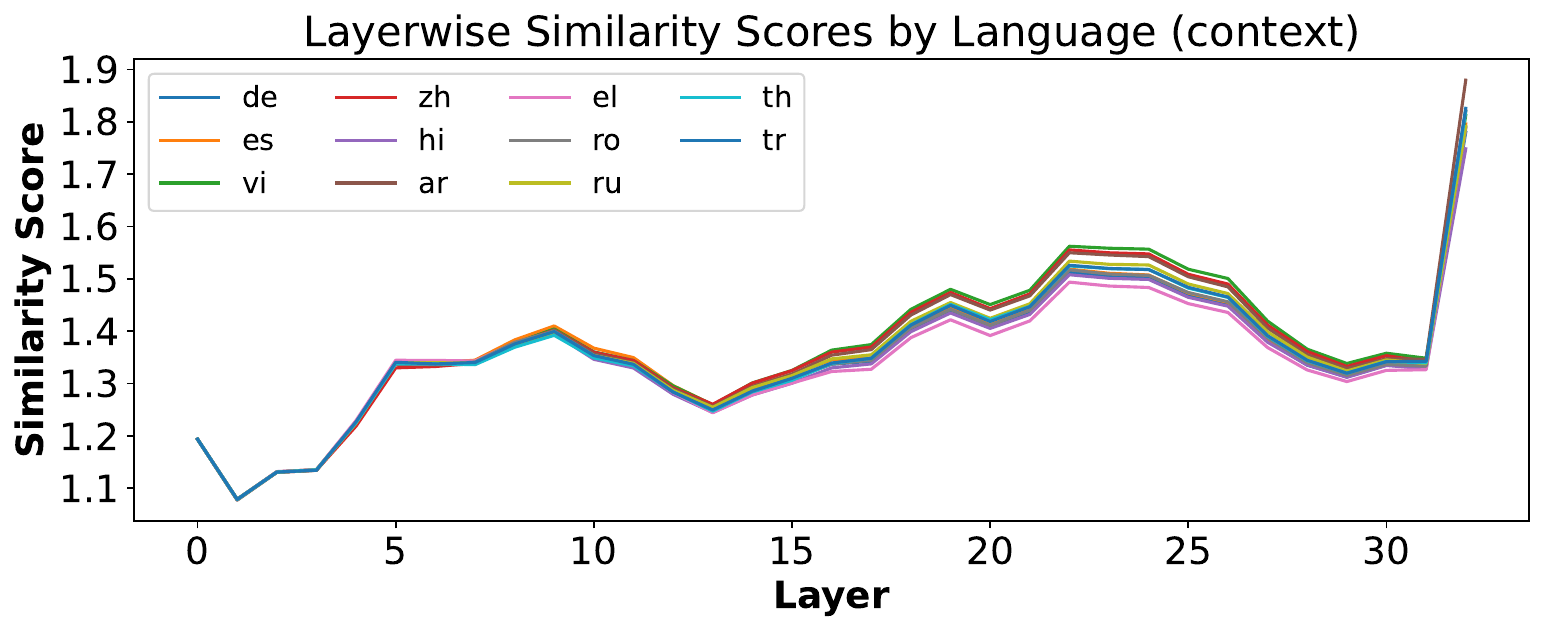}
        \caption{Context}
        \label{llama-3.1-instruct-8b-2-shot-balanced-c}
    \end{subfigure}
    \caption{Question, last token and context hidden state similarity between English and other languages in each layer of the LLaMA-3.1-Instruct-8B model on the ``balanced'' samples.}
    \label{llama-3.1-instruct-8b-2-shot-balanced}
\end{figure}

Also, the MRDs of ``balanced'' samples (Figures \ref{llama-3.1-instruct-8b-balanced-cq}) are more stable than those of ``en-superior'' samples (Figure \ref{llama-3.1-instruct-8b-en-sup-cq}), suggesting correlation between the higher xMRC performance and clearer phasing.

These patterns are consistent for different LLaMA models regardless of prompt formats, except much weaker for LLaMA-2-Chat-7B, which has smaller pre-training capacity and weaker multilingual ability (see Appendix \ref{appendix:further_analysis_MRD} for mode details).

In summary, \textbf{the attribution results supports the two-phase hypothesis}, and indicates that the phased behavior is already formed after pre-training, regardless of model size and prompt format. The phasing strength correlates with the pre-training capacity and the xMRC task performance.



\subsection{Evidence from hidden states view}
The hidden state similarity results also support our hypothesis. Figure \ref{llama-3.1-instruct-8b-2-shot-balanced} shows the en-x hidden state similarity of the question, last input token (predicting the start of the answer) and context parts for the LLaMA-3.1-Instruct-8B model (more results are in Appendix \ref{appendix:other_hidden_sim}). The observed trends are consistent:

\begin{itemize} [nosep,itemsep=2pt,leftmargin=0.2cm]
    \item For question representations, they all show a shared arc-shaped trend, where the highest similarity to English appears at the relative depth of around 1/3;
    \item For context representations, a consistent double-peak trend can be observed, with a ``turning point'' around the relative depth of 0.4 (matching the question MRD) and the second, higher peak at around 0.7 (matching the context MRD);
    \item For last input token representations, one can see a consistent ``plateau'' of similarity starting at around a relative depth of 0.5, which also matches the mean question MRD.
\end{itemize}

It is worth noticing that, though the context parts are all English, there are 2-shot demonstrations with non-English questions, making the representations not identical. The context similarity curves can represent the degree of semantic encoding compared with formal encoding.

Again, for the less powerful model LLaMA-2-Chat-7B (Figure \ref{llama-2-chat-7b-2-shot-hidden-all} in Appendix \ref{appendix:other_hidden_sim}), the trends are weaker: its question similarity to English varies much across languages, and the ``plateau'' of answer similarity to English starts later than other models, which is after the mean question and context MRDs.

These results suggest that \textbf{the hidden states similarity through the xMRC process also undergo two main phases with evident distinction}. This phased behavior already exists in pre-trained LLMs, and is preserved in post-training. Also, the phasing strength correlates with the model capability built during pre-training.




\begin{figure}[ht]
    \centering
    \begin{subfigure}{0.45\textwidth}
        \includegraphics[width=0.95\linewidth]{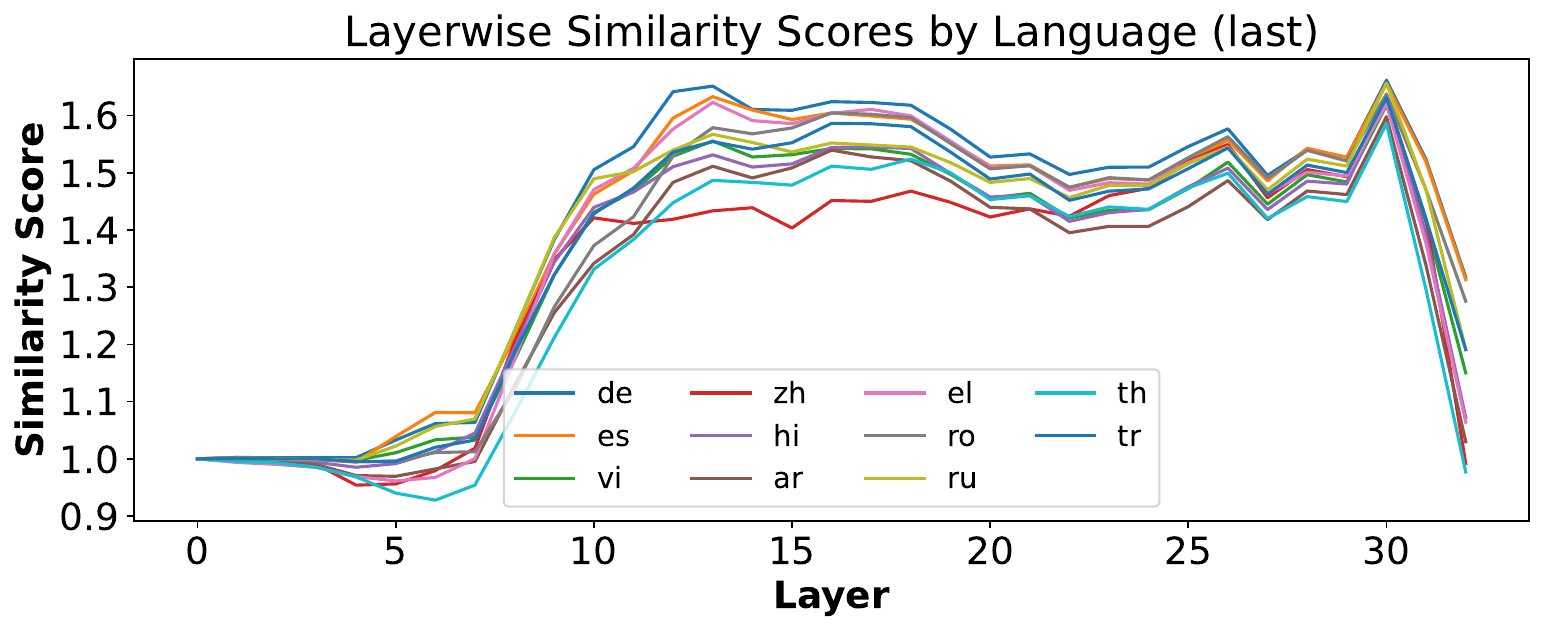}
        \caption{LLaMA-3.1-8B}
        \label{fig:llama-3.1-tuned-8b-2-shot-balanced-compare-a}
    \end{subfigure}
    \hfill
    \begin{subfigure}{0.45\textwidth}
        \includegraphics[width=0.95\linewidth]{./assets/llama-3.1-instruct-8b-2-shot-balanced-l.pdf}
        \caption{LLaMA-3.1-Instruct-8B}
        \label{fig:llama-3.1-tuned-8b-2-shot-balanced-compare-b}
    \end{subfigure}
    \hfill
    \begin{subfigure}{0.45\textwidth}
        \includegraphics[width=0.95\linewidth]{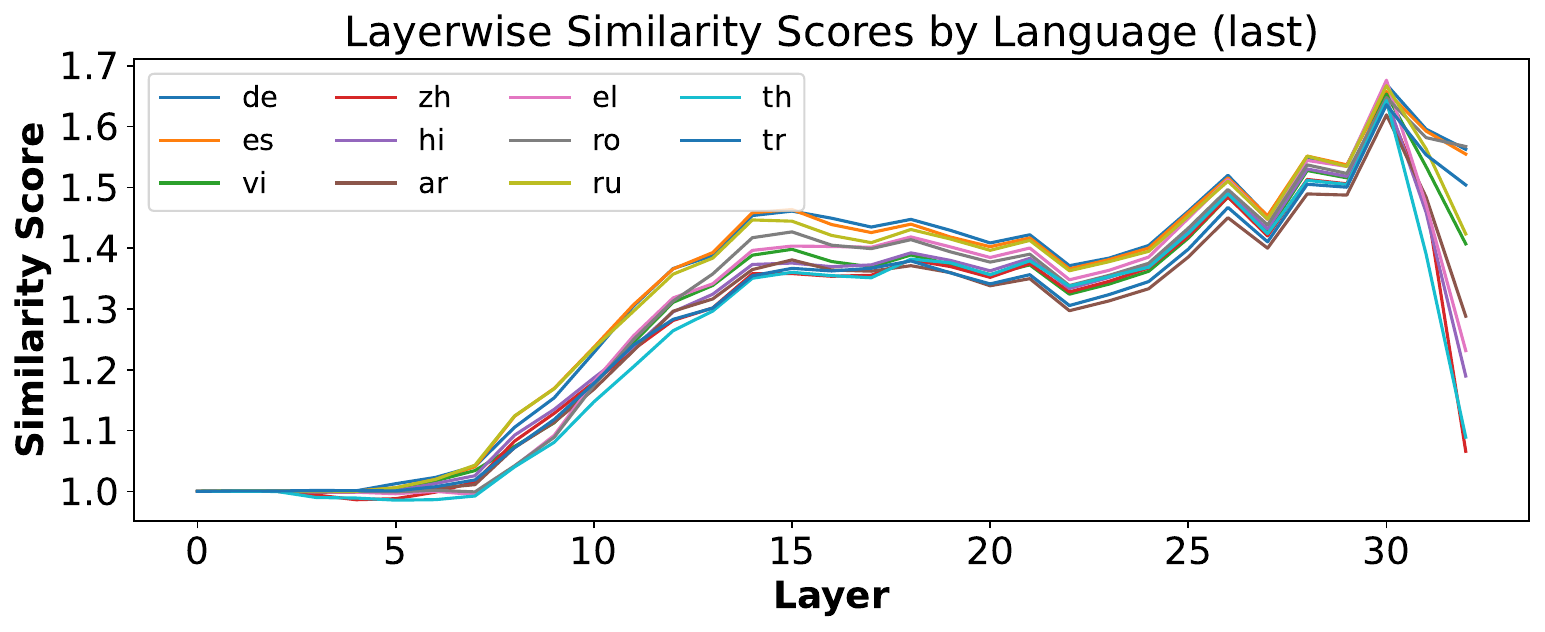}
        \caption{LLaMA-3.1-Tuned-8B}
        \label{fig:llama-3.1-tuned-8b-2-shot-balanced-compare-c}
    \end{subfigure}
    \caption{Change in last-input-token hidden state similarity between English and other languages in each layer of LLaMA-3.1-8B, LLaMA-3.1-Instruct-8B and LLaMA-3.1-Tuned-8B on the ``balanced'' samples.}
    \label{fig:llama-3.1-tuned-8b-2-shot-balanced-compare}
\end{figure}

\begin{figure}[ht]
    \centering
    \begin{subfigure}{0.45\textwidth}
        \includegraphics[width=0.95\linewidth]{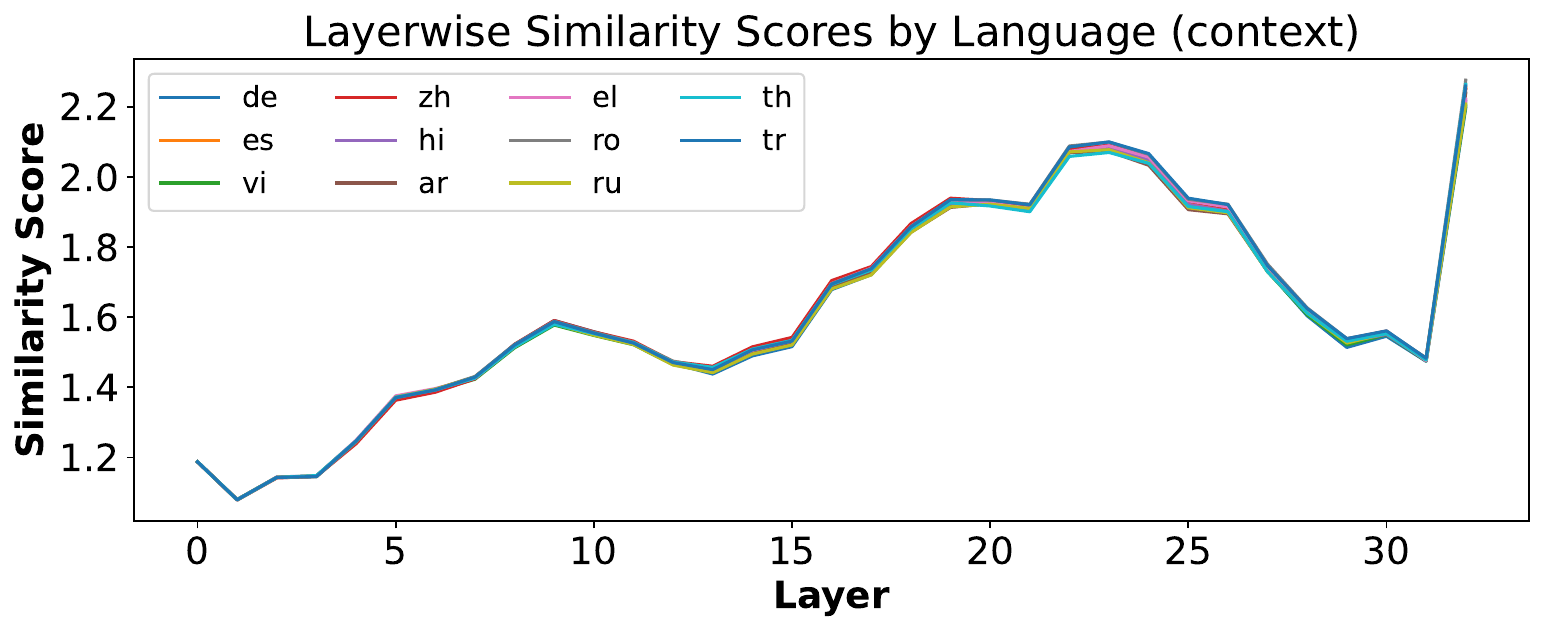}
        \caption{LLaMA-3.1-8B}
    \end{subfigure}
    \hfill
    \begin{subfigure}{0.45\textwidth}
        \includegraphics[width=0.95\linewidth]{./assets/llama-3.1-instruct-8b-2-shot-balanced-c.pdf}
        \caption{LLaMA-3.1-Instruct-8B}
    \end{subfigure}
    \hfill
    \begin{subfigure}{0.45\textwidth}
        \includegraphics[width=0.95\linewidth]{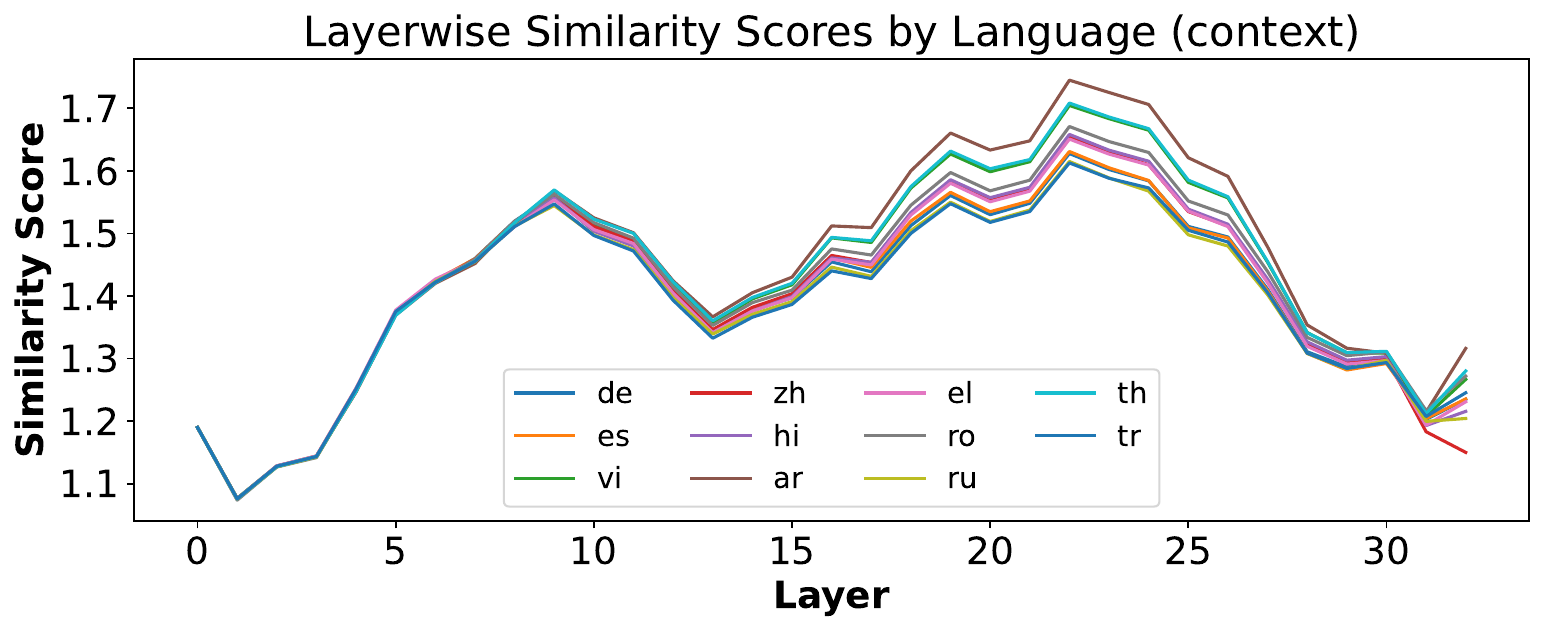}
        \caption{LLaMA-3.1-Tuned-8B}
    \end{subfigure}
    \caption{Change in context hidden state similarity between English and other languages in each layer of LLaMA-3.1-8B, LLaMA-3.1-Instruct-8B and LLaMA-3.1-Tuned-8B on the ``balanced'' samples.}
    \label{fig:llama-3.1-tuned-8b-2-shot-balanced-compare-context}
\end{figure}

\subsection{Importance of post-training}
\label{pre-training_vs_post-training}

Our evaluation results show that post-training is crucial to enhancing xMRC performance. Here, we show the significance of post-training to xMRC from the hidden-state view.

\begin{figure}[ht]
    \centering
    \begin{subfigure}{0.45\textwidth}
        \includegraphics[width=0.95\linewidth]{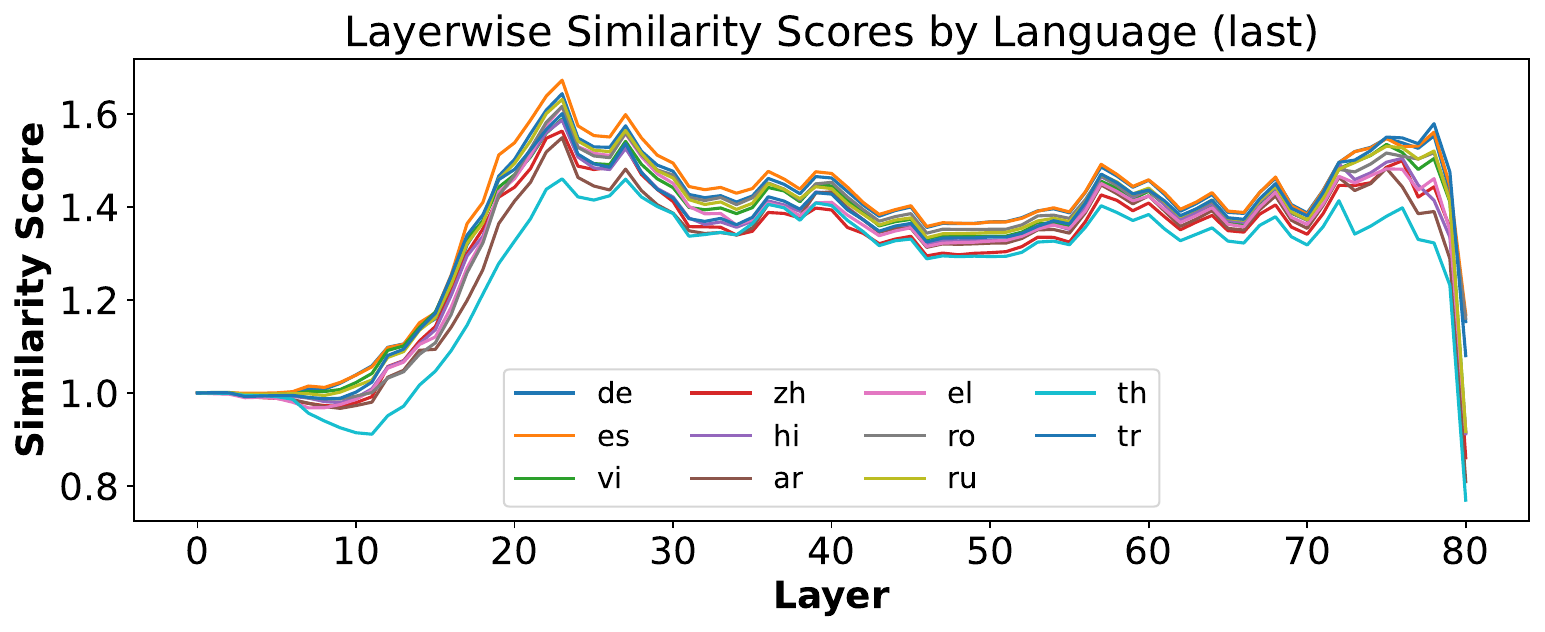}
        \caption{LLaMA-3.1-70B}
        \label{fig:llama-3.1-70b-2-shot-balanced-compare-last-a}
    \end{subfigure}
    \hfill
    \begin{subfigure}{0.45\textwidth}
        \includegraphics[width=0.95\linewidth]{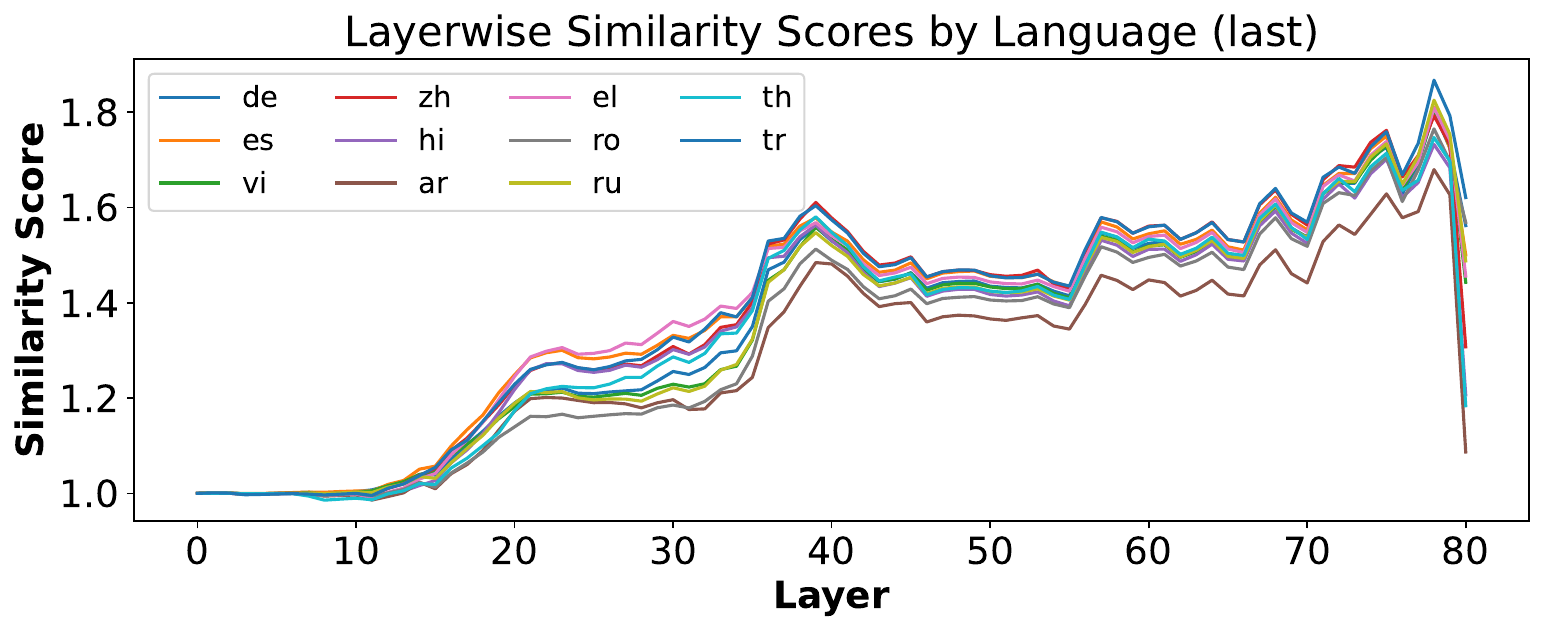}
        \caption{LLaMA-3.1-Instruct-70B}
        \label{fig:llama-3.1-70b-2-shot-balanced-compare-last-b}
    \end{subfigure}
    \caption{Change in last-input-toke hidden state similarity between English and other languages in each layer of LLaMA-3.1-70B and LLaMA-3.1-Instruct-70B on the ``balanced'' samples.}
    \label{fig:llama-3.1-70b-2-shot-balanced-compare-last}
\end{figure}

The first evidence comes from the last-input-token similarity results. One can observe from Figure \ref{fig:llama-3.1-tuned-8b-2-shot-balanced-compare-a} and \ref{fig:llama-3.1-70b-2-shot-balanced-compare-last-a} that, \textbf{the last-input-token similarity of base models experience severe and consistent decline in the last few layers}, across all non-English languages. Since we expect the same English output for all tested languages, this drop in cross-lingual similarity can directly affect the performance and its cross-lingual alignment. However, \textbf{after post-training on 8B models} (Figure \ref{fig:llama-3.1-tuned-8b-2-shot-balanced-compare-b}, \ref{fig:llama-3.1-tuned-8b-2-shot-balanced-compare-c} and \ref{fig:llama-3.1-70b-2-shot-balanced-compare-last-b}), \textbf{the decline significantly narrows}, and even turns into increase for languages with Latino alphabets, especially on the 8B models (Figure \ref{fig:llama-3.1-tuned-8b-2-shot-balanced-compare}). Since this enhancement in similarity can directly turn into the narrowing of language performance gap, this indicates that post-training is especially essential to enhancing the cross-lingual alignment xMRC ability, by taking effect in the last few layers and the final calculation steps. However, \textbf{for the 70B model} (Figure \ref{fig:llama-3.1-70b-2-shot-balanced-compare-last}), \textbf{the decline is still severe} after post-training, partly explaining why they show larger xMRC gap between English and non-English.

\begin{figure}[!ht]
    \centering
    \begin{subfigure}{0.45\textwidth}
        \includegraphics[width=0.95\linewidth]{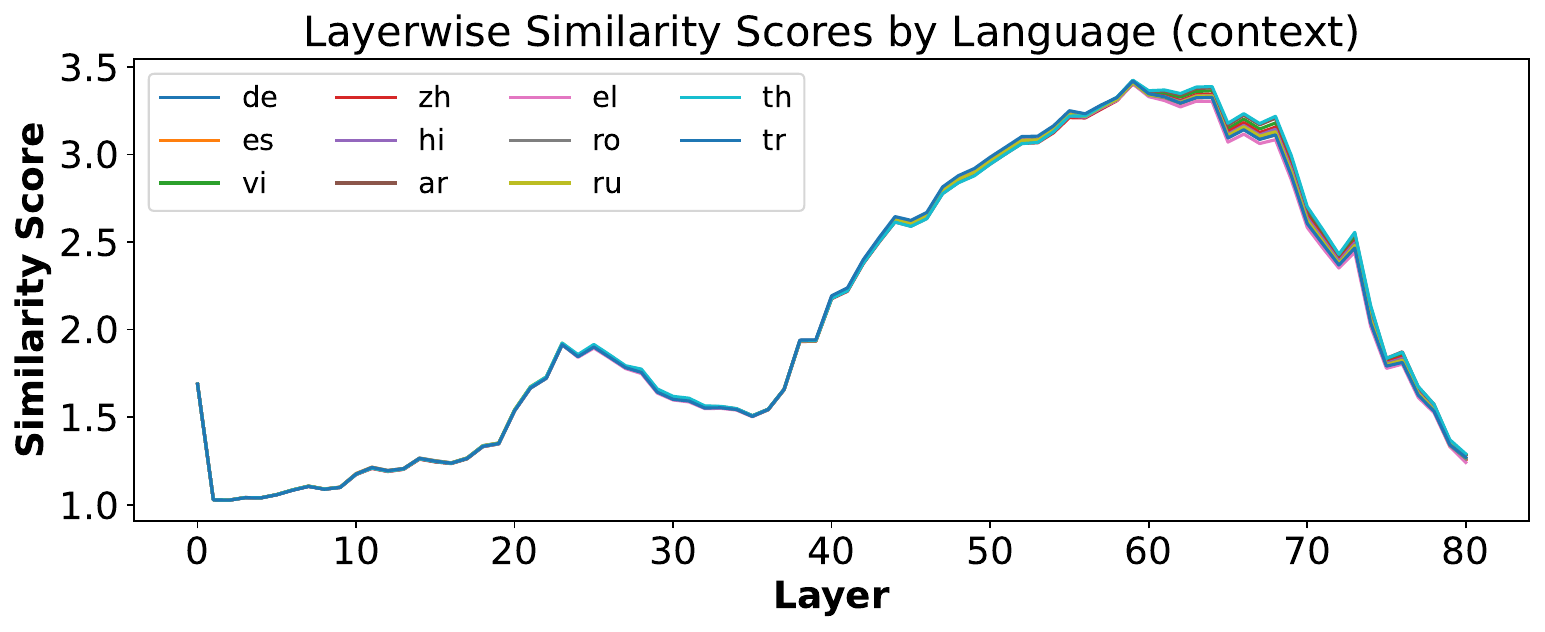}
        \caption{LLaMA-3.1-70B}
    \end{subfigure}
    \hfill
    \begin{subfigure}{0.45\textwidth}
        \includegraphics[width=0.95\linewidth]{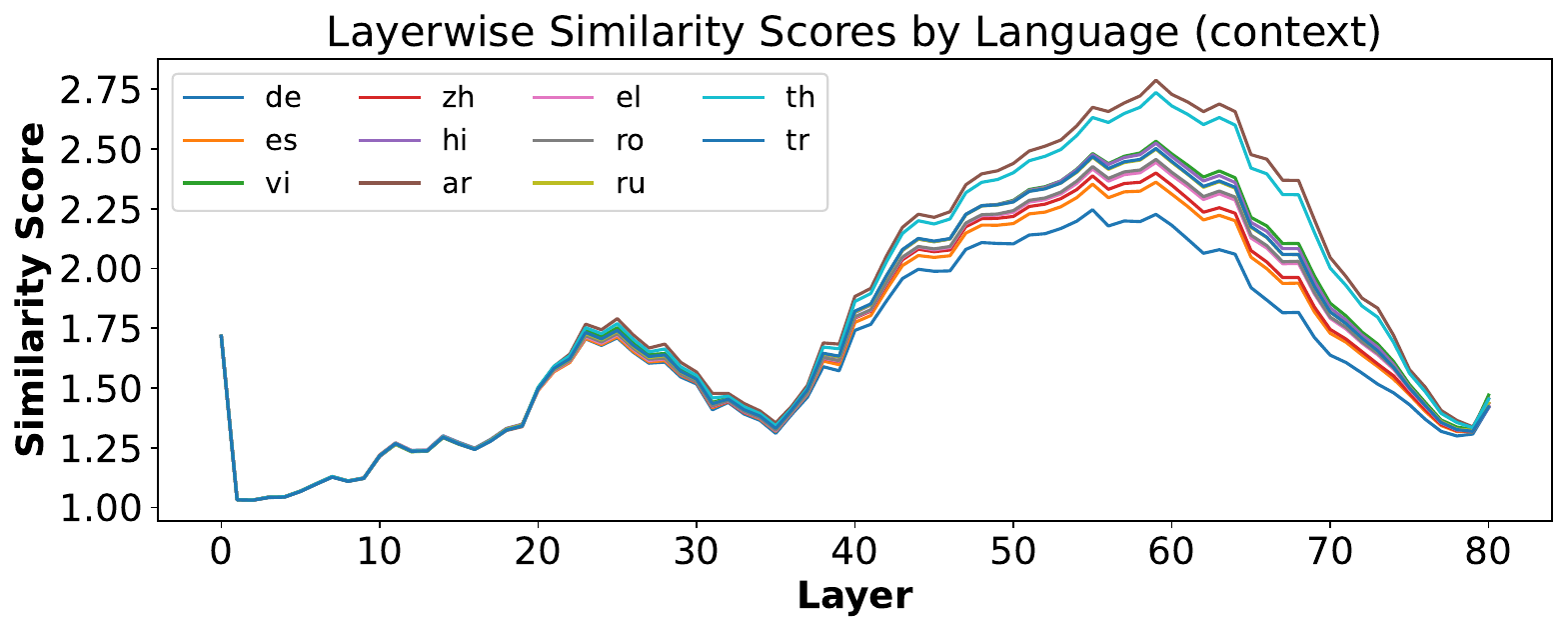}
        \caption{LLaMA-3.1-Instruct-70B}
    \end{subfigure}
    \caption{Change in context hidden state similarity between English and other languages in each layer of LLaMA-3.1-70B and LLaMA-3.1-Instruct-70B on the ``balanced'' samples.}
    \label{fig:llama-3.1-70b-2-shot-balanced-compare-context}
\end{figure}

Another evidence comes from the context similarity results. One can see from Figure \ref{fig:llama-3.1-tuned-8b-2-shot-balanced-compare-context} and \ref{fig:llama-3.1-70b-2-shot-balanced-compare-context} that, the context similarity value significantly decreases after post-training, and the peak value diverges among languages. 
Based on our understanding of the score, \textbf{lower context similarity means its contextual representations become more customized for the demonstrations} with non-English questions, which is potentially beneficial for xMRC. This also partly explains the enhancement of xMRC after post-training.
However, \textbf{the divergence among languages means that this driving effect of demonstrations varies in different languages}, contributing to the performance gap between languages. It is especially evident for the 70B model (Figure \ref{fig:llama-3.1-70b-2-shot-balanced-compare-context}) that, better-xMRC-performing languages (e.g. de, es) tend to show lower peak values, while worse-xMRC-performing languages (e.g. ar, th) tend to show higher. This corresponds with the reasoning that lower context similarity correlates with higher xMRC performance, and adds to the explanation why 70B models have larger performance gaps between languages.

Based on these findings, we propose a possible reason for the larger language gap for 70B post-trained models that \textbf{their post-training is insufficient to reshape the answer retrieving phase, and is biased to certain languages}, which causes the near-base behavior in last-token similarity and the larger divergence in context similarity. With more sufficient and language-balanced post-training, we expect the Instruct-70B model to reveal similar patterns in last-input-token and context hidden states similarity as Instruct-8B.


\section{Related Work}
\subsection{Cross-lingual alignment of LLMs}
Previous studies have shown a misalignment of LLMs with English and other languages. With respect to performance, \citet{laiChatGPTEnglishComprehensive2023}, \citet{ahuja-etal-2024-megaverse} and \citet{wang-etal-2024-seaeval} demonstrated that SOTA LLMs performed better in English, and showed inconsistency when dealing with non-English queries. \citet{etxaniz-etal-2024-multilingual} found that LLMs performed worse with non-English prompts than with self-translated English prompts. Beyond performance, \citet{qiCrossLingualConsistencyFactual2023} demonstrated the low cross-lingual consistency of factual knowledge does of LLMs, and \citet{gao-etal-2024-multilingual} showed that multilingual pre-training and instruction tuning could only enhance superficial levels of cross-lingual alignment. \citet{zhaoHowLargeLanguage2024c} demonstrated that multilingual LLMs employ shared and language-specific circuits to process different languages when needed.
These papers mostly focus on same-language queries, and our work focuses on cross-lingual queries. 

There have also been many techniques to enhance LLMs' cross-lingual alignment. For example, adding parallel data in the pre-training stage \cite{lampleCrosslingualLanguageModel2019a,jiangXLMKImprovingCrossLingual2022,weiPolyLMOpenSource2023,lu-etal-2024-llamax}; and the post-training stage, including instruction tuning \cite{liM3ITLargeScaleDataset2023,zhang-etal-2025-extrapolating,liBactrianXMultilingualReplicable2023,cahyawijaya-etal-2023-instructalign,chaiXCoTCrosslingualInstruction2024a,kuulmets-etal-2024-teaching,shaham-etal-2024-multilingual,kew-etal-2024-turning} and preference tuning \cite{lai-etal-2023-okapi,she-etal-2024-mapo}. Especially, extra translation training is commonly used \cite{zhangBayLingBridgingCrosslingual2023a,yangBigTranslateAugmentingLarge2023a,li-etal-2024-eliciting,ranaldi-etal-2024-empowering,zhu-etal-2024-question,lu-etal-2024-llamax}. In this paper, we examine the effect of some of these techniques by comparing various SOTA models.

\subsection{Cross-lingual machine reading comprehension}
xMRC is a relatively new task of natural language understanding. \citet{cui-etal-2019-cross} proposed the task, in order to improve non-English MRC performance by introducing English resources. There are some representative datasets in this area, such as XQA \cite{liu-etal-2019-xqa}, BiPaR \cite{jing2019biparbilingualparalleldataset}, MLQA \cite{lewis-etal-2020-mlqa}, and XQuAD \cite{artetxe-etal-2020-cross}. \citet{ushioPracticalToolkitMultilingual2023} also proposes a pipeline for multilingual QA generation. However, previous work on the xMRC task mainly focuses on enhancing the performance of task-specific models using techniques such as data augmentation \cite{borneaMultilingualTransferLearning2021,xiangCrossguidanceCrosslingualModel2024}, knowledge injection \cite{duanBridgingLanguageGap2021}, constractive learning \cite{chenGoodBestTwoStage2022} and knowledge transfer \cite{caoSharingTeachingAligning2023,xuMPMRMultilingualPretrained2023}. In this paper, we study the xMRC of multilingual LLMs under the larger topic of cross-lingual alignment.

\section{Conclusion}
This paper investigates the performance and mechanism of cross-lingual context retrieval of LLMs within the xMRC scenario. For evaluation, we demonstrate the strong xMRC abilities of post-trained models, and study their bottlenecks and oracle performances. For mechanism, we verify a two-phased hypothesis of the xMRC process, identifying the effect of post-training, and finding a possible explanation to the language gap for larger post-trained models. We hope our research will inspire future study to foster the cross-lingual alignment of LLMs in a broader scope.

\section*{Limitations}
While this study provides insights of the cross-lingual context retrieval abilities of LLMs, there are also some limitations.

First, the scope of our empirical evaluation is constrained by available resources and time. This necessarily limits the breadth of our testing, preventing us from exhaustively covering the rapidly expanding landscape of LLMs. Besides, while we test across 12 diverse languages, a more comprehensive analysis would ideally include an even wider range of languages, as well as grouping them according to their levels of resource, in order to improve the generalizability of our findings across linguistic diversity. 

Second, although we identify a two-phased feature of xMRC and confirm its correlation with pre-training and post-training, the precise factors within these training processes that drive this outcome remain unclear. Future work could delve deeper into the data and strategies of these training stages to locate factors contributing to the emergence and strength of the phasing.

Finally, within the two major phases we discover, we observe hints of more fine-grained changes in model behavior, particularly in the hidden state similarity curves. These preliminary observations suggest the potential for a more nuanced understanding of the xMRC process. Future studies could further investigate these finer-grained dynamics within each phase to gain a more detailed and complete picture of how LLMs achieve cross-lingual context retrieval.

Beyond these limitations, it is also important to consider potential risks associated with this work. While our research is foundational and not directly tied to specific applications, advancements in cross-lingual context retrieval, like any technology, could be misused. For example, improved cross-lingual capabilities might inadvertently contribute to the spread of misinformation if models are used to retrieve and amplify biased or inaccurate information across languages. Furthermore, if deployed without careful consideration, these technologies could exacerbate existing inequalities by favoring languages and knowledge systems already dominant in LLM training data, potentially marginalizing less-represented languages and perspectives. Future work should consider these dual-use aspects and explore mitigation strategies to ensure responsible development and deployment of cross-lingual NLP technologies, paying special attention to fairness and inclusivity across diverse linguistic communities.

\section*{Ethics Statements}
This research adheres to ethical principles in its use of language models and data. All language models evaluated and finetuned in this study are accessed and utilized in compliance with their respective licenses and terms of service. Furthermore, the XQuAD dataset employed for evaluation, and the TULU-v3 dataset used for finetuning LLaMA-3.1-8B, are both publicly available datasets intended for research purposes. Based on our review and the documented nature of these datasets, we have determined that they are not designed to collect or contain personally identifiable information or offensive content. To the best of our knowledge, and as indicated in their public documentation, neither dataset includes data that names or uniquely identifies individual people, nor do they present offensive content.

Our use of these existing artifacts, including both language models and datasets, is aligned with their intended use within research contexts. Specifically, derivatives of data accessed for research purposes, such as model outputs and analysis results, are used solely within the bounds of academic inquiry and are not disseminated or utilized outside of these research contexts, in accordance with responsible data handling practices.

\section*{Acknowledgments}
We would like to thank the anonymous reviewers for their insightful comments. Shujian Huang is the corresponding author. This work is supported by National Science Foundation of China (No. 62376116, 62176120), research project of Nanjing University-China Mobile Joint Institute (NJ20250038), the Fundamental Research Funds for the Central Universities (No. 2024300507, 2025300390).


\clearpage
\newpage
\normalem
\appendix

\section{Additional experiment information}

\subsection{Example of unnecessary source text in outputs}
\label{appendix:example_unnecessary}
We do not use exact matching (EM) as the main performance metric because it can be easily affected by unnecessary source text in the model outputs. Here is an example:\\

\noindent\fbox{%
    \parbox{0.45\textwidth}{\small
        Context: ``The Panthers defense gave up just 308 points, ranking sixth in the league, while also leading the NFL in interceptions with 24 and boasting four Pro Bowl selections. Pro Bowl defensive tackle Kawann Short led the team in sacks with 11, while also forcing three fumbles and recovering two. Fellow lineman Mario Addison added 6½ sacks. The Panthers line also featured veteran defensive end Jared Allen, a 5-time pro bowler who was the NFL's active career sack leader with 136, along with defensive end Kony Ealy, who had 5 sacks in just 9 starts. Behind them, two of the Panthers three starting linebackers were also selected to play in the Pro Bowl: Thomas Davis and Luke Kuechly. Davis compiled 5½ sacks, four forced fumbles, and four interceptions, while Kuechly led the team in tackles (118) forced two fumbles, and intercepted four passes of his own. Carolina's secondary featured Pro Bowl safety Kurt Coleman, who led the team with a career high seven interceptions, while also racking up 88 tackles and Pro Bowl cornerback Josh Norman, who developed into a shutdown corner during the season and had four interceptions, two of which were returned for touchdowns.''\\

        Question: How many Panthers defense players were selected for the Pro Bowl?\\

        Reference: four\\

        Model Answer: four Pro Bowl selections.
    }%
}%
\par\par The reference is ``four'', which is consistent with the model output, but the ``Pro Bowl selections.'' in the model output is unnecessary and will cause the EM metric to give a 0 result. 

\subsection{Accuracy of language detection tool}
\label{appendix:lingual_acc}
We use Lingua with its high accuracy mode in this work to detect the language of model outputs. Table \ref{tab:lingua-acc} shows its reported accuracies on the tested languages, which is satisfactory to serve in our experiments.

\begin{table}[t]
\centering
\footnotesize
\begin{tabular}{@{}ccc@{}}
\toprule
\textbf{Code} & \textbf{Language} & \textbf{\begin{tabular}[c]{@{}c@{}}Average Acc\\ (high accuracy mode)\end{tabular}} \\ \midrule
en & English & 81 \\
de & German & 89 \\
es & Spanish & 70 \\
vi & Vietnamese & 91 \\
zh & Chinese & 100 \\
hi & Hindi & 73 \\
ar & Arabic & 98 \\
el & Greek & 100 \\
ro & Romanian & 87 \\
ru & Russian & 90 \\
th & Thai & 99 \\
tr & Turkish & 94 \\ \bottomrule
\end{tabular}
\caption{Detection accuracies of the tested language with Lingua in its high-accuracy mode, adapted from their GitHub page (\url{https://github.com/pemistahl/lingua}).}
\label{tab:lingua-acc}
\end{table}

\subsection{Prompt used in evaluation and error type detection}
We use the default system prompt and chat templates assigned in the \texttt{tokenizer.config} files of the model repositories. Then, we apply these configurations to two prompt formats which we call v1 and v2.

The v1 prompt format is:\\

\noindent\fbox{%
    \parbox{0.45\textwidth}{\small
        \{system prompt\}\\

        Below is a reading comprehension task. There will be paragraphs of context, each followed by a question related to its content. You should only present your answer to the last question by strictly copying the corresponding part of the context. Please provide a direct answer in English without extra output. Your answer should be in the form of ``Answer: \{Your Answer\}''\\

        Context: \{demo context 1\}\\

        Question: \{demo question 1\}\\

        Answer: \{demo answer 1\}\\

        Context: \{demo context 2\}\\

        Question: \{demo question 2\}\\

        Answer: \{demo answer 2\}\\

        Your task starts here:\\
        
        Context: \{text context\}\\

        Question: \{text question\}
    }%
}%

\par\par The v2 prompt format is:\\

\noindent\fbox{%
    \parbox{0.45\textwidth}{\small
        \{system prompt\}\\

        Context: \{demo context 1\}\\

        Question: \{demo question 1\}\\

        Answer: \{demo answer 1\}\\

        Context: \{demo context 2\}\\

        Question: \{demo question 2\}\\

        Answer: \{demo answer 2\}\\

        Your task starts here:\\
        
        Context: \{text context\}\\

        Question: \{text question\}
        
        You should only present your answer to the last question by strictly copying the corresponding part of the context. Please provide a direct answer in English without extra output. Your answer should be in the form of ``Answer: \{Your Answer\}''
    }%
}%

\par\par The prompt for error type detection is:\\

\label{appendix:error-type-prompt}
\noindent\fbox{%
    \parbox{0.45\textwidth}{\small
        \texttt{<|im\_start|>}system\\
You are Qwen, created by Alibaba Cloud. You are a helpful assistant.\texttt{<|im\_end|>}\\
\texttt{<|im\_start|>}user\\
You are tasked with identifying the type of a given raw answer. You will be provided with a question and a raw answer. Your job is to determine whether the raw answer falls into one of the following categories based on the given question:\\

0. Reasonable Answer: The answer seems like some attempt to answer the question, regardless of whether it is correct or not.\\

1. Blank Answer: No response is provided.\\

2. Gibberish: Incoherent text with no clear meaning or cannot be seen as some kind of answer to the question, e.g. ``\{Your Answer\}''.\\

3. Denial of Answer: A statement indicating inability to answer, such as ``I apologize, but I cannot answer this question because...''.\\

You must provide your response as a SINGLE number representing the category (0, 1, 2, or 3) without extra output.
    }%
}%

\section{More evaluation results}
\label{sec:appendix}

\subsection{Full list of models evaluated}
\label{appendix:model_lists}

A comprehensive list of all models evaluated during this study, including older versions and alternative sizes, is available in Table \ref{all_models_appendix} to supplement the main list.

\begin{table}[!hbt]
  \centering
  \small
  \setlength{\tabcolsep}{4pt}
  \renewcommand{\arraystretch}{1.1}
  \begin{tabularx}{0.48\textwidth}{@{}>{\centering\arraybackslash}p{2.8cm}>{\centering\arraybackslash}p{1.8cm}>{\centering\arraybackslash}p{2.5cm}@{}}
    \hline
    \textbf{Name} & \textbf{Modes} & \textbf{Sizes} \\
    \hline
    LLaMA 2 & Base / Chat & 7B / 13B / 70B \\
    LLaMA 3 & Base / Instruct & 8B / 70B \\
    LLaMA 3.1 & Base / Instruct & 8B / 70B \\
    LLaMAX-2-Alpaca & - & 7B \\
    LLaMAX-3-Alpaca & - & 8B \\
    Mistral V0.1 & Base / Instruct & 7B \\
    Mistral V0.3 & Base / Instruct & 7B \\
    Qwen 1.5 & Base / Chat & 7B / 14B / 72B \\
    Qwen 2 & Base / Instruct & 7B / 72B \\
    Qwen 2.5 & Base / Instruct & 7B / 72B \\
    DeepSeek V2 & Base / Chat & Lite (16B) \\
    Gemma 2 & Base / IT & 9B \\
    GPT-3.5-Turbo-0125 & - & - \\
    GPT-4o & - & - \\\hline
  \end{tabularx}
  \caption{Full list of models evaluated. This table presents a complete list of all models tested in this study, encompassing older versions and alternative sizes.}
  \label{all_models_appendix}
\end{table}

\subsection{Detailed evaluation results}
\label{appendix:0-shot_results}
Table \ref{0-shot_all_results} presents the complete evaluation results from our 0-shot experiments, encompassing both the English-to-NonEnglish (en-x) and non-English monolingual (x-x) tasks in all models and languages tested. Table \ref{2-shot_F1_detailed} further presents detailed F1 scores for en-x and x-x tasks in the 2-shot setting.

\subsection{Performance vs. parameter size}
\label{appendix:increase_size}

To observe the effect of increasing model size on xMRC performance, we evaluate the Qwen-2.5 models from 1.5B to 72B, showing results in Table \ref{tab:qwen-sizes}. The trend is consistent with our findings in the main body, and an advantage of the 7B model stands out. 

\begin{table*}[hb]
\centering
\scriptsize
\begin{tabular}{@{}c|cccccc@{}}
\toprule
\textbf{Version} & \textbf{Size} & \textbf{en-en} & \textbf{mean en-x} & \textbf{mean x-x} & \textbf{en-x / en-en} & \textbf{x-x / en-en} \\ \midrule
\multirow{6}{*}{Base} & 1.5B & 61.26 & 47.10 & 54.55 & 0.77 & 0.89 \\
 & 3B & 71.09 & 58.75 & 60.00 & 0.83 & 0.84 \\
 & 7B & 62.42 & 57.51 & 66.11 & \textbf{0.92} & \textbf{1.06} \\
 & 14B & 76.71 & 69.04 & 74.60 & 0.90 & 0.97 \\
 & 32B & 80.37 & 73.23 & 74.02 & 0.91 & 0.92 \\
 & 72B & \textbf{86.03} & \textbf{78.92} & \textbf{81.16} & \textbf{0.92} & 0.94 \\ \midrule
\multirow{6}{*}{Instruct} & 1.5B & 70.63 & 55.21 & 58.11 & 0.78 & 0.82 \\
 & 3B & 71.65 & 64.41 & 60.39 & 0.90 & 0.84 \\
 & 7B & \textbf{81.83} & \textbf{76.43} & 71.61 & \textbf{0.93} & 0.88 \\
 & 14B & 80.91 & 67.64 & \textbf{72.95} & 0.84 & 0.90 \\
 & 32B & 80.97 & 73.77 & 71.90 & 0.91 & 0.89 \\
 & 72B & 77.12 & 66.04 & 70.29 & 0.86 & \textbf{0.91} \\ \bottomrule
\end{tabular}
\caption{xMRC performances of Qwen-2.5 models with increasing parameter sizes.}
\label{tab:qwen-sizes}
\end{table*}

\begin{table*}[ht]
    \scriptsize
    \centering
    \begin{center}
    \begin{subtable}[t]{\textwidth}
    \begin{tabular}{@{}ccccccccccccc@{}}
    \hline
    \toprule
    \textbf{}                          & \textbf{en-en} & \textbf{en-de} & \textbf{en-es} & \textbf{en-vi} & \textbf{en-zh} & \textbf{en-hi} & \textbf{en-ar} & \textbf{en-el} & \textbf{en-ro} & \textbf{en-ru} & \textbf{en-th} & \textbf{en-tr} \\ \midrule
    \textbf{LLaMA-3.1-8B}              & 28.49          & 26.39          & 23.70          & 16.97          & 21.76          & 21.68          & 19.66          & 24.96          & 22.27          & 22.69          & 27.73          & 15.71          \\
    \textbf{LLaMA-3.1-70B}             & 49.92          & 38.32          & 36.22          & 41.34          & 45.13          & 24.62          & 27.40          & 37.73          & 38.49          & 34.87          & 47.90          & 33.95          \\
    \textbf{Mistral-V0.3-7B}           & 20.50          & 21.43          & 19.50          & 10.84          & 8.66           & 11.60          & 14.87          & 14.20          & 12.44          & 18.32          & 16.13          & 10.67          \\
    \textbf{Qwen-2.5-7B}               & 38.66          & 36.05          & 37.31          & 48.32          & 42.02          & 38.91          & 42.44          & 34.87          & 35.55          & 38.57          & 45.63          & 41.60          \\
    \textbf{Qwen-2.5-72B}              & 63.95          & 56.55          & 56.72          & 53.19          & 52.86          & 55.97          & 56.30          & 52.27          & 54.03          & 55.55          & 55.29          & 51.68          \\
    \textbf{DeepSeek-V2-Lite-16B}      & 12.35          & 11.26          & 4.12           & 4.20           & 9.58           & 9.58           & 4.79           & 7.73           & 7.06           & 10.84          & 5.80           & 4.96           \\
\bottomrule
    \end{tabular}
    \caption{0-shot Exact Match (EM) scores (\%) on en-x tasks.} 
    \end{subtable}
    \end{center}

    \centering
    \begin{center}
    \begin{subtable}[t]{\textwidth}
    \begin{tabular}{@{}ccccccccccccc@{}}
    \hline
    \toprule
    \textbf{}                          & \textbf{en-en} & \textbf{en-de} & \textbf{en-es} & \textbf{en-vi} & \textbf{en-zh} & \textbf{en-hi} & \textbf{en-ar} & \textbf{en-el} & \textbf{en-ro} & \textbf{en-ru} & \textbf{en-th} & \textbf{en-tr} \\ \midrule
    \textbf{LLaMA-3.1-8B}              & 37.60          & 34.11          & 33.87          & 21.86          & 35.12          & 36.97          & 30.11          & 36.66          & 30.82          & 30.37          & 39.43          & 21.79          \\
    \textbf{LLaMA-3.1-70B}             & 68.03          & 58.29          & 56.22          & 60.33          & 55.20          & 50.63          & 41.23          & 56.41          & 57.13          & 55.87          & 62.87          & 54.32          \\
    \textbf{Mistral-V0.3-7B}           & 40.07          & 27.17          & 31.26          & 19.26          & 13.19          & 17.94          & 23.98          & 19.16          & 18.23          & 28.26          & 21.86          & 18.63          \\
    \textbf{Qwen-2.5-7B}               & 59.02          & 55.58          & 55.61          & 64.37          & 60.26          & 56.32          & 60.61          & 53.14          & 51.21          & 56.72          & 63.23          & 55.52          \\
    \textbf{Qwen-2.5-72B}              & 80.50          & 73.73          & 74.48          & 72.48          & 71.37          & 73.60          & 73.23          & 70.59          & 71.44          & 73.36          & 72.78          & 68.18          \\
    \textbf{DeepSeek-V2-Lite-16B}      & 24.88          & 25.14          & 11.07          & 13.25          & 18.55          & 14.01          & 13.69          & 15.32          & 14.72          & 23.10          & 9.28           & 13.73          \\
\bottomrule
    \end{tabular}
    \caption{0-shot F1 Scores on en-x tasks.}
    \end{subtable}
    \end{center}

    \centering
    \begin{center}
    \begin{subtable}[t]{\textwidth}
    \begin{tabular}{@{}ccccccccccccc@{}}
    \hline
    \toprule
    \textbf{}                          & \textbf{de-de} & \textbf{es-es} & \textbf{vi-vi} & \textbf{zh-zh} & \textbf{hi-hi} & \textbf{ar-ar} & \textbf{el-el} & \textbf{ro-ro} & \textbf{ru-ru} & \textbf{th-th} & \textbf{tr-tr} \\ \midrule
    \textbf{LLaMA-3.1-8B}              & 24.87          & 19.16          & 18.66          & 33.95          & 16.05          & 17.31          & 10.00          & 18.15          & 17.90          & 33.87          & 14.37          \\
    \textbf{LLaMA-3.1-70B}             & 42.94          & 37.23          & 42.86          & 50.08          & 30.00          & 40.17          & 36.30          & 40.76          & 35.13          & 57.06          & 37.40          \\
    \textbf{Mistral-V0.3-7B}           & 27.31          & 25.29          & 20.17          & 34.62          & 7.98           & 21.09          & 17.73          & 22.69          & 11.93          & 23.45          & 12.44          \\
    \textbf{Qwen-2.5-7B}               & 29.50          & 35.04          & 38.57          & 57.23          & 23.11          & 42.94          & 21.09          & 31.85          & 32.18          & 53.95          & 31.51          \\
    \textbf{Qwen-2.5-72B}              & 46.89          & 46.39          & 51.93          & 78.32          & 41.18          & 50.67          & 36.64          & 50.67          & 43.03          & 63.78          & 41.34          \\
    \textbf{DeepSeek-V2-Lite-16B}      & 4.87           & 2.44           & 1.01           & 7.56           & 2.02           & 1.43           & 4.03           & 3.19           & 2.02           & 5.13           & 2.94           \\
    \textbf{Gemma-2-9B}                & 2.02           & 14.03          & 8.91           & 12.10          & 8.66           & 1.43           & 1.43           & 4.62           & 10.92          & 9.66           & 13.28          \\
\bottomrule
    \end{tabular}
    \caption{0-shot Exact Match (EM) scores (\%) on x-x tasks}
    \end{subtable}
    \end{center}
\end{table*}
    
\begin{table*}[ht]
    \scriptsize
    \ContinuedFloat
    \begin{center}
    \begin{subtable}[t]{\textwidth}
    \begin{tabular}{@{}ccccccccccccc@{}}
    \hline
    \toprule
    \textbf{}                          & \textbf{de-de} & \textbf{es-es} & \textbf{vi-vi} & \textbf{zh-zh} & \textbf{hi-hi} & \textbf{ar-ar} & \textbf{el-el} & \textbf{ro-ro} & \textbf{ru-ru} & \textbf{th-th} & \textbf{tr-tr} \\ \hline
    \textbf{LLaMA-3.1-8B}              & 37.62          & 34.52          & 36.04          & 43.66          & 41.18          & 34.54          & 32.58          & 34.61          & 35.68          & 47.38          & 28.02          \\
    \textbf{LLaMA-3.1-70B}             & 62.66          & 59.02          & 61.54          & 55.26          & 56.81          & 63.43          & 56.52          & 59.81          & 55.99          & 70.20          & 57.39          \\
    \textbf{Mistral-V0.3-7B}           & 41.31          & 47.37          & 39.40          & 41.73          & 24.04          & 42.28          & 35.43          & 37.91          & 29.51          & 37.70          & 28.32          \\
    \textbf{Qwen-2.5-7B}               & 49.22          & 58.04          & 62.82          & 68.32          & 46.19          & 63.63          & 44.78          & 52.10          & 52.96          & 68.59          & 54.30          \\
    \textbf{Qwen-2.5-72B}              & 70.77          & 72.81          & 74.95          & 84.33          & 68.09          & 72.94          & 64.72          & 73.17          & 68.56          & 78.50          & 67.79          \\
    \textbf{DeepSeek-V2-Lite-16B}      & 12.56          & 8.24           & 7.64           & 9.73           & 8.35           & 6.98           & 9.59           & 9.39           & 7.79           & 9.03           & 8.89           \\
    \textbf{Gemma-2-9B}                & 4.16           & 20.57          & 13.66          & 14.74          & 15.58          & 2.53           & 3.07           & 6.96           & 18.84          & 14.14          & 23.02          \\
\bottomrule
    \end{tabular}
    \caption{0-shot F1 Scores on x-x tasks.}
    \end{subtable}
    \end{center}
    \caption{0-shot evaluation results on en-x and x-x tasks.}
    \label{0-shot_all_results}
\end{table*}

\begin{table*}[ht]
    \scriptsize
    \begin{center}
    \begin{subtable}[t]{\textwidth}
    \begin{tabular}{@{}ccccccccccccc@{}}
    \toprule
    \textbf{}                          & \textbf{en-en} & \textbf{en-de} & \textbf{en-es} & \textbf{en-vi} & \textbf{en-zh} & \textbf{en-hi} & \textbf{en-ar} & \textbf{en-el} & \textbf{en-ro} & \textbf{en-ru} & \textbf{en-th} & \textbf{en-tr} \\ \midrule
    \textbf{LLaMA-3.1-8B}              & 75.97          & 45.97          & 43.90          & 50.71          & 47.70          & 48.41          & 50.98          & 50.40          & 47.22          & 49.12          & 59.77          & 44.94          \\
    \textbf{LLaMA-3.1-70B}             & 82.39          & 60.46          & 60.00          & 59.57          & 62.68          & 57.38          & 51.50          & 59.43          & 54.28          & 56.15          & 66.54          & 57.48          \\
    \textbf{Mistral-V0.3-7B}           & 79.57          & 66.94          & 67.57          & 59.82          & 53.66          & 48.36          & 52.56          & 47.83          & 67.30          & 70.62          & 48.56          & 62.94          \\
    \textbf{Qwen-2.5-7B}               & 62.42          & 56.68          & 56.45          & 59.15          & 58.84          & 55.85          & 62.94          & 50.62          & 54.43          & 58.37          & 61.61          & 57.72          \\
    \textbf{Qwen-2.5-72B}              & 86.03          & 77.24          & 79.22          & 80.16          & 80.14          & 79.09          & 78.70          & 76.72          & 80.41          & 80.77          & 78.48          & 77.16          \\
    \textbf{DeepSeek-V2-Lite-16B}      & 73.81          & 46.34          & 47.65          & 52.75          & 41.77          & 34.45          & 44.61          & 46.18          & 50.39          & 51.91          & 34.58          & 40.57          \\
\bottomrule
    \end{tabular}
    \caption{2-shot F1 scores on en-x tasks.} 
    \end{subtable}
    \end{center}
    
    \begin{center}
    \begin{subtable}[t]{\textwidth}
    \begin{tabular}{cccccccccccc}
    \toprule
    \textbf{}                          & \textbf{de-de} & \textbf{es-es} & \textbf{vi-vi} & \textbf{zh-zh} & \textbf{hi-hi} & \textbf{ar-ar} & \textbf{el-el} & \textbf{ro-ro} & \textbf{ru-ru} & \textbf{th-th} & \textbf{tr-tr} \\ \midrule
    \textbf{LLaMA-3.1-8B}              & 71.67          & 74.17          & 73.19          & 64.96          & 71.91          & 68.20          & 69.35          & 74.65          & 67.17          & 70.89          & 66.89          \\
    \textbf{LLaMA-3.1-70B}             & 76.24          & 78.68          & 76.90          & 71.67          & 75.85          & 76.43          & 72.41          & 78.06          & 68.43          & 76.70          & 70.67          \\
    \textbf{Mistral-V0.3-7B}           & 71.02          & 72.91          & 69.91          & 66.65          & 56.73          & 58.25          & 62.81          & 71.73          & 63.57          & 62.08          & 58.44          \\
    \textbf{Qwen-2.5-7B}               & 58.74          & 57.36          & 72.29          & 73.38          & 63.74          & 72.74          & 67.24          & 58.82          & 64.12          & 77.89          & 60.84          \\
    \textbf{Qwen-2.5-72B}              & 81.29          & 81.15          & 82.82          & 89.08          & 79.12          & 79.53          & 77.83          & 82.18          & 77.49          & 85.00          & 77.29          \\
    \textbf{DeepSeek-V2-Lite-16B}      & 65.26          & 67.33          & 64.81          & 63.68          & 48.64          & 46.98          & 51.04          & 65.73          & 56.19          & 49.43          & 55.20          \\
    \textbf{Gemma-2-9B}                & 75.62          & 76.34          & 73.60          & 66.92          & 73.90          & 72.51          & 71.87          & 78.14          & 67.38          & 74.20          & 71.43          \\
\bottomrule
    \end{tabular}
    \caption{2-shot F1 scores on x-x tasks}
    \end{subtable}
    \end{center}
    \caption{Detailed 2-shot F1 scores on en-x and x-x tasks in each language.}
    \label{2-shot_F1_detailed}
\end{table*}

\subsection{Detailed language error and generation error rates}
\label{appendix:detailed_generate_failure_error}

Tables \ref{2-shot_language_error} and \ref{2-shot_generation_failure_error} show detailed language and generation error rates across all tested languages. Note that Table \ref{2-shot_generation_failure_error} contain the results judged by both Qwen-2.5-Instruct-72B and Gemini-2.5-Flash-Lite, which are consistent to each other.
Meanwhile, Table \ref{2-shot_detailed_generation_failure_error} provides a more granular view of the generation errors discussed in the main text. 

While Table \ref{2-shot_generation_failure_error} presents the aggregated rate of these errors, Table \ref{2-shot_detailed_generation_failure_error} is further subdivided into three separate tables: Table \ref{Gibberish}, Table \ref{Refusal}, and Table \ref{Blank}. These tables individually display the error rates for gibberish errors, refusal errors, and blank errors, respectively, across all tested models and languages in the 2-shot en-x xMRC task setting.

\begin{table*}[ht]
    \scriptsize
    \begin{center}
\begin{tabular}{cccccccccccc}
    \toprule
    \textbf{}                          & \textbf{en-de} & \textbf{en-es} & \textbf{en-vi} & \textbf{en-zh} & \textbf{en-hi} & \textbf{en-ar} & \textbf{en-el} & \textbf{en-ro} & \textbf{en-ru} & \textbf{en-th} & \textbf{en-tr} \\ \midrule
    \textbf{LLaMA-3.1-8B}              & 0.68           & 0.00           & 1.28           & 0.00           & 0.10           & 0.10           & 0.00           & 0.00           & 0.19           & 0.10           & 1.10           \\
    \textbf{LLaMA-3.1-70B}             & 60.20          & 54.59          & 63.73          & 48.90          & 69.09          & 66.86          & 56.83          & 67.88          & 56.69          & 56.09          & 61.50          \\
    \textbf{Mistral-V0.3-7B}           & 2.11           & 1.78           & 16.39          & 23.80          & 61.82          & 54.17          & 6.96           & 2.65           & 1.02           & 46.35          & 16.63          \\
    \textbf{Qwen-2.5-7B}               & 2.42           & 1.01           & 0.43           & 0.43           & 0.00           & 0.00           & 0.00           & 1.15           & 0.00           & 0.00           & 5.17           \\
    \textbf{Qwen-2.5-72B}              & 7.39           & 6.35           & 11.22          & 0.98           & 5.74           & 2.82           & 19.78          & 11.36          & 3.59           & 27.01          & 23.64          \\
    \textbf{DeepSeek-V2-Lite-16B}      & 8.90           & 1.87           & 1.69           & 26.75          & 46.80          & 4.76           & 3.68           & 2.27           & 0.47           & 35.17          & 10.31          \\
    \textbf{Gemma-2-9B}                & 0.00           & 0.10           & 9.78           & 2.37           & 0.50           & 0.20           & 0.34           & 3.15           & 0.00           & 2.23           & 2.37           \\
\bottomrule
    \end{tabular}
    \end{center}
    \caption{2-shot language error rates (\%) on en-x xMRC tasks.}
    \label{2-shot_language_error}
\end{table*}

\begin{table*}[ht]
    \scriptsize
    \begin{center}
    
    \begin{subtable}[t]{\textwidth}
    \begin{tabular}{@{}ccccccccccccc@{}}
\toprule
\textbf{} & \textbf{en-en} & \textbf{en-de} & \textbf{en-es} & \textbf{en-vi} & \textbf{en-zh} & \textbf{en-hi} & \textbf{en-ar} & \textbf{en-el} & \textbf{en-ro} & \textbf{en-ru} & \textbf{en-th} & \textbf{en-tr} \\ \midrule
\textbf{LLaMA-3.1-8B} & 5.60 & 6.77 & 8.55 & 10.97 & 5.00 & 9.59 & 8.37 & 10.56 & 9.64 & 7.94 & 9.74 & 10.52 \\
\textbf{LLaMA-3.1-70B} & 1.20 & 1.98 & 2.41 & 2.19 & 2.62 & 4.12 & 2.33 & 2.35 & 1.71 & 1.57 & 2.70 & 4.94 \\
\textbf{Mistral-V0.3-7B} & 0.49 & 4.85 & 4.93 & 17.66 & 10.00 & 32.00 & 13.97 & 18.50 & 4.86 & 5.25 & 26.43 & 18.33 \\
\textbf{Qwen-2.5-7B} & 1.51 & 2.03 & 1.30 & 5.72 & 1.96 & 2.26 & 3.11 & 6.09 & 2.10 & 3.99 & 2.85 & 4.77 \\
\textbf{Qwen-2.5-72B} & 0.00 & 1.19 & 1.35 & 0.97 & 1.82 & 3.69 & 2.62 & 3.16 & 1.91 & 2.94 & 4.93 & 3.29 \\
\textbf{DeepSeek-V2-Lite-16B} & 1.87 & 3.26 & 2.33 & 11.97 & 2.39 & 20.61 & 10.10 & 8.04 & 4.27 & 2.76 & 15.97 & 11.23 \\
\textbf{Gemma-2-9B} & 1.02 & 1.34 & 1.38 & 5.39 & 2.60 & 6.51 & 4.69 & 4.98 & 3.68 & 2.64 & 5.05 & 6.98 \\
\bottomrule
\end{tabular}
    \caption{Qwen-judged results}
    \label{2-shot_generation_failure_error_qwen}
    \end{subtable}

    \begin{subtable}[t]{\textwidth}
    \begin{tabular}{@{}ccccccccccccc@{}}
\toprule
\textbf{} & \textbf{en-en} & \textbf{en-de} & \textbf{en-es} & \textbf{en-vi} & \textbf{en-zh} & \textbf{en-hi} & \textbf{en-ar} & \textbf{en-el} & \textbf{en-ro} & \textbf{en-ru} & \textbf{en-th} & \textbf{en-tr} \\ \midrule
\textbf{LLaMA-3.1-8B} & 4.80 & 4.85 & 7.53 & 7.61 & 4.18 & 6.06 & 5.88 & 7.77 & 7.70 & 6.44 & 5.98 & 8.31 \\
\textbf{LLaMA-3.1-70B} & 0.00 & 0.22 & 0.22 & 0.00 & 0.72 & 1.24 & 0.36 & 0.64 & 0.19 & 0.99 & 1.08 & 0.83 \\
\textbf{Mistral-V0.3-7B} & 0.49 & 1.35 & 0.81 & 4.86 & 4.15 & 9.38 & 6.43 & 5.33 & 1.35 & 0.00 & 9.43 & 7.38 \\
\textbf{Qwen-2.5-7B} & 1.08 & 1.11 & 0.37 & 0.84 & 0.39 & 0.19 & 0.24 & 1.18 & 0.35 & 1.00 & 0.44 & 1.59 \\
\textbf{Qwen-2.5-72B} & 0.00 & 0.00 & 0.00 & 0.00 & 0.45 & 0.46 & 0.00 & 1.19 & 0.48 & 0.49 & 0.90 & 0.82 \\
\textbf{DeepSeek-V2-Lite-16B} & 1.12 & 1.46 & 1.34 & 3.69 & 0.30 & 10.63 & 4.83 & 1.93 & 1.78 & 2.20 & 5.89 & 3.50 \\
\textbf{Gemma-2-9B} & 0.00 & 0.44 & 0.23 & 1.88 & 0.97 & 1.25 & 0.29 & 1.15 & 1.31 & 0.00 & 1.27 & 1.96 \\
\bottomrule
\end{tabular}
    \caption{Gemini-judged results}
    \label{2-shot_generation_failure_error_gemini}
    \end{subtable}
    \end{center}
    
    \caption{2-shot generation error rates (\%) on en-x xMRC tasks judged by Qwen and Gemini.}
    \label{2-shot_generation_failure_error}
\end{table*}

\begin{table*}[ht]
    \scriptsize
    \begin{center}
    \begin{subtable}[t]{\textwidth}
    \begin{tabular}{@{}ccccccccccccc@{}}
    \toprule
    \textbf{}                          & \textbf{en-en} & \textbf{en-de} & \textbf{en-es} & \textbf{en-vi} & \textbf{en-zh} & \textbf{en-hi} & \textbf{en-ar} & \textbf{en-el} & \textbf{en-ro} & \textbf{en-ru} & \textbf{en-th} & \textbf{en-tr} \\ \midrule
    \textbf{LLaMA-3.1-8B}              & 4.80           & 4.03           & 1.83           & 7.08           & 2.83           & 5.38           & 3.56           & 2.99           & 2.45           & 2.70           & 6.42           & 7.81           \\
    \textbf{LLaMA-3.1-70B}             & 0.60           & 1.32           & 2.19           & 1.31           & 2.14           & 3.50           & 2.15           & 1.93           & 1.33           & 1.18           & 2.43           & 4.12           \\
    \textbf{Mistral-V0.3-7B}           & 0.00           & 3.77           & 4.11           & 16.78          & 9.62           & 31.16          & 12.50          & 18.17          & 4.32           & 4.94           & 26.43          & 16.90          \\
    \textbf{Qwen-2.5-7B}               & 1.29           & 1.11           & 1.11           & 4.45           & 1.57           & 1.69           & 2.63           & 5.41           & 1.75           & 3.59           & 2.63           & 4.17           \\
    \textbf{Qwen-2.5-72B}              & 0.00           & 1.19           & 1.35           & 0.97           & 1.82           & 3.69           & 2.62           & 3.16           & 1.91           & 2.45           & 4.93           & 2.88           \\
    \textbf{DeepSeek-V2-Lite-16B}      & 1.87           & 2.77           & 2.00           & 10.31          & 2.24           & 19.72          & 8.24           & 6.27           & 3.20           & 2.02           & 15.84          & 10.79          \\
\bottomrule
    \end{tabular}
    \caption{Gibberish error.} 
    \label{Gibberish}
    \end{subtable}
    \end{center}
    
    \begin{center}
    \begin{subtable}[t]{\textwidth}
    \begin{tabular}{@{}ccccccccccccc@{}}
    \toprule
    \textbf{}                          & \textbf{en-en} & \textbf{en-de} & \textbf{en-es} & \textbf{en-vi} & \textbf{en-zh} & \textbf{en-hi} & \textbf{en-ar} & \textbf{en-el} & \textbf{en-ro} & \textbf{en-ru} & \textbf{en-th} & \textbf{en-tr} \\ \midrule
    \textbf{LLaMA-3.1-8B}              & 0.00           & 0.16           & 0.00           & 0.00           & 0.00           & 0.34           & 0.18           & 0.18           & 0.00           & 0.00           & 0.22           & 0.00           \\
    \textbf{LLaMA-3.1-70B}             & 0.00           & 0.00           & 0.00           & 0.22           & 0.00           & 0.00           & 0.18           & 0.21           & 0.00           & 0.00           & 0.27           & 0.41           \\
    \textbf{Mistral-V0.3-7B}           & 0.00           & 0.00           & 0.00           & 0.00           & 0.19           & 0.17           & 1.10           & 0.00           & 0.00           & 0.31           & 0.00           & 0.00           \\
    \textbf{Qwen-2.5-7B}               & 0.22           & 0.92           & 0.19           & 0.21           & 0.39           & 0.38           & 0.48           & 0.34           & 0.35           & 0.40           & 0.22           & 0.20           \\
    \textbf{Qwen-2.5-72B}              & 0.00           & 0.00           & 0.00           & 0.00           & 0.00           & 0.00           & 0.00           & 0.00           & 0.00           & 0.00           & 0.00           & 0.00           \\
    \textbf{DeepSeek-V2-Lite-16B}      & 0.00           & 0.00           & 0.00           & 0.37           & 0.00           & 0.38           & 0.62           & 0.16           & 0.00           & 0.37           & 0.13           & 0.00           \\
\bottomrule
    \end{tabular}
    \caption{Refusal error.}
    \label{Refusal}
    \end{subtable}
    \end{center}
    
    \begin{center}
    \begin{subtable}[t]{\textwidth}
    \begin{tabular}{@{}ccccccccccccc@{}}
    \toprule
    \textbf{}                          & \textbf{en-en} & \textbf{en-de} & \textbf{en-es} & \textbf{en-vi} & \textbf{en-zh} & \textbf{en-hi} & \textbf{en-ar} & \textbf{en-el} & \textbf{en-ro} & \textbf{en-ru} & \textbf{en-th} & \textbf{en-tr} \\ \midrule
    \textbf{LLaMA-3.1-8B}              & 0.80           & 2.58           & 6.72           & 3.89           & 2.17           & 3.87           & 4.63           & 7.39           & 7.19           & 5.24           & 3.10           & 2.71           \\
    \textbf{LLaMA-3.1-70B}             & 0.60           & 0.66           & 0.22           & 0.66           & 0.48           & 0.62           & 0.00           & 0.21           & 0.38           & 0.39           & 0.00           & 0.41           \\
    \textbf{Mistral-V0.3-7B}           & 0.49           & 1.08           & 0.82           & 0.88           & 0.19           & 0.67           & 0.37           & 0.33           & 0.54           & 0.00           & 0.00           & 1.43           \\
    \textbf{Qwen-2.5-7B}               & 0.00           & 0.00           & 0.00           & 1.06           & 0.00           & 0.19           & 0.00           & 0.34           & 0.00           & 0.00           & 0.00           & 0.40           \\
    \textbf{Qwen-2.5-72B}              & 0.00           & 0.00           & 0.00           & 0.00           & 0.00           & 0.00           & 0.00           & 0.00           & 0.00           & 0.49           & 0.00           & 0.41           \\
    \textbf{DeepSeek-V2-Lite-16B}      & 0.00           & 0.49           & 0.33           & 1.29           & 0.15           & 0.51           & 1.24           & 1.61           & 1.07           & 0.37           & 0.00           & 0.44           \\
\bottomrule
    \end{tabular}
    \caption{Blank error.}
    \label{Blank}
    \end{subtable}
    \end{center}
    \caption{Detailed percentages (\%) of generation error types (gibberish error, refusal error, and blank error) on 2-shot en-x tasks.}
    \label{2-shot_detailed_generation_failure_error}
\end{table*}

\section{Two-phased xMRC Analysis}

\subsection{Further analysis on MRD}
\label{appendix:further_analysis_MRD}

\subsubsection{Example of attribution results}
\begin{figure}[ht]
    \centering
    \includegraphics[width=8cm]{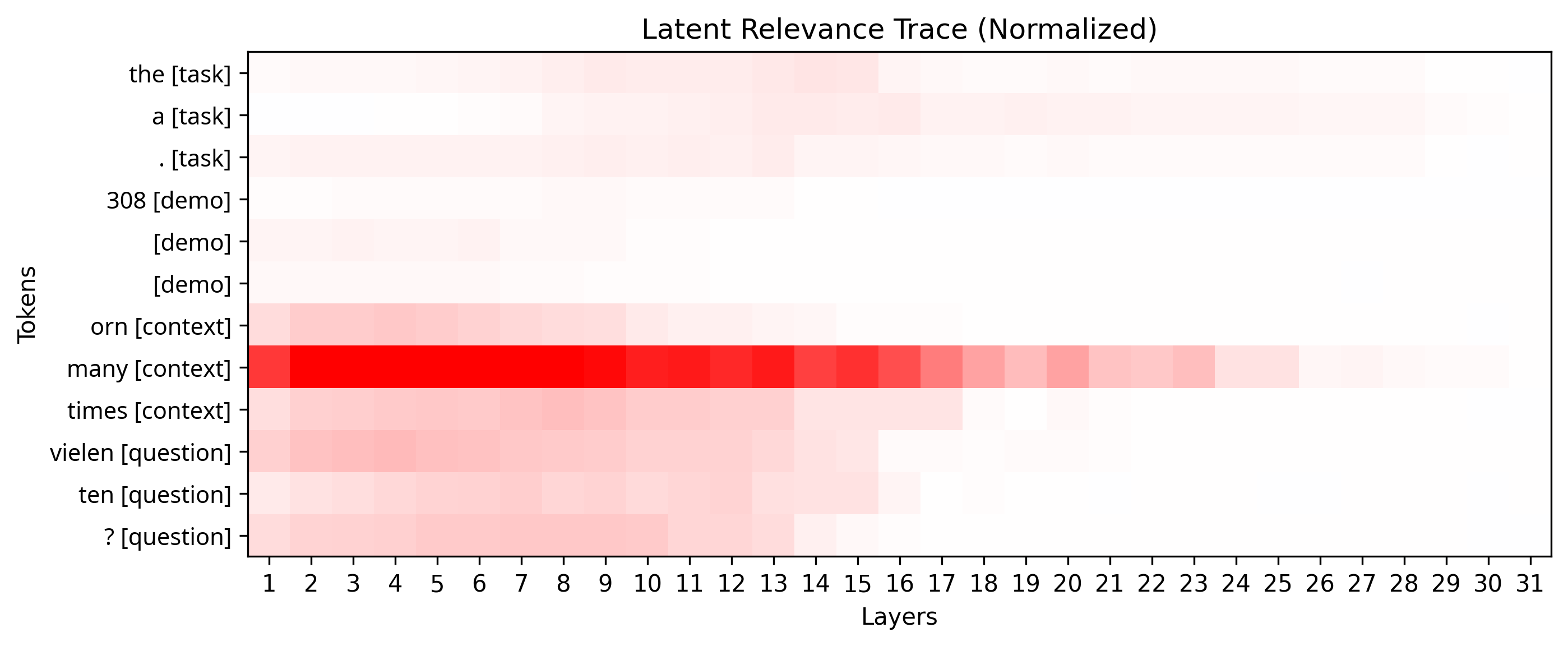}
    \caption{An example output of layer-wise attribution with LLaMA-3.1-Instruct-8B, where only the top 3 tokens from each input part are shown.} 
    \label{layer_attr_sample}
\end{figure}

Figure \ref{layer_attr_sample} shows an example of the attribution outcome for LLaMA-3.1-Instruct-8B. 

\subsubsection{MRD for other LLaMA models}
\label{appendix:more_MRD_cq_examples}

\begin{figure}[ht]
    \centering
    \begin{subfigure}{0.45\textwidth}
        \includegraphics[width=\textwidth]{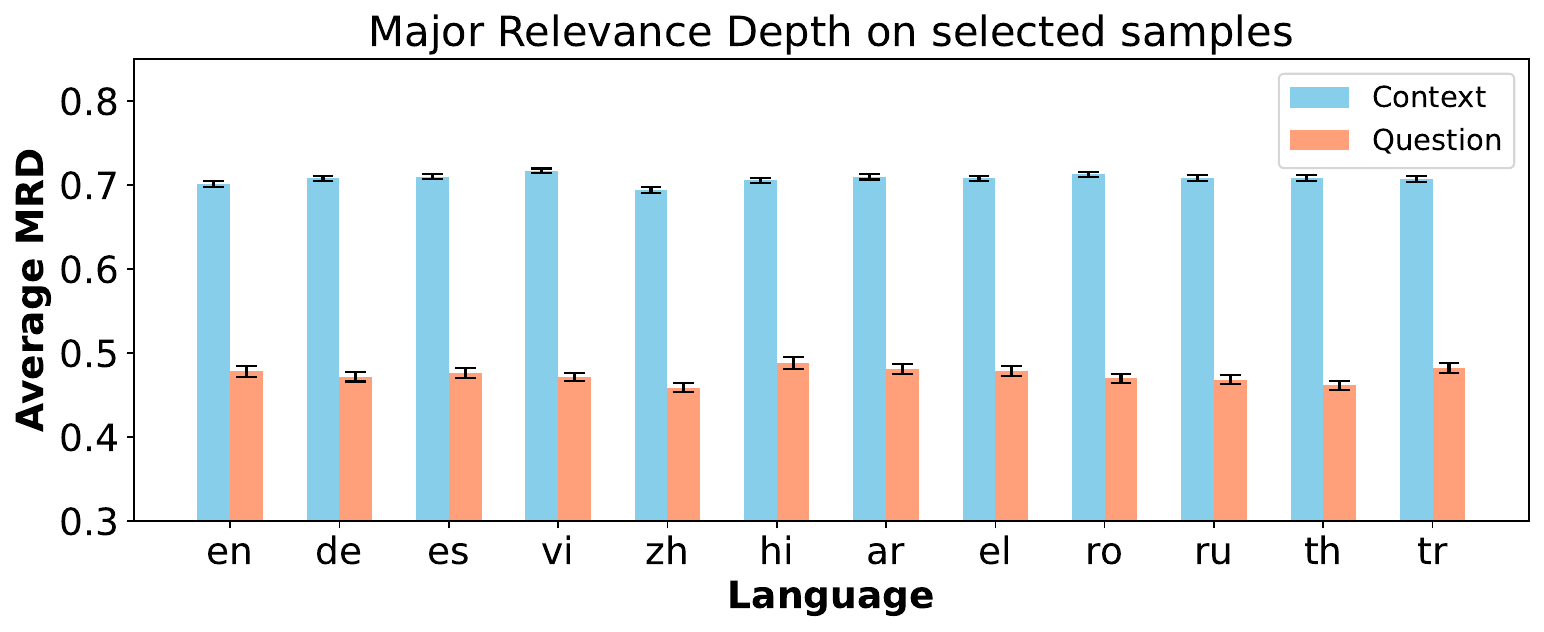}
        \caption{balanced samples}
        \label{llama-3.1-base-8b-balanced-cq}
    \end{subfigure}
    \hfill
    \begin{subfigure}{0.45\textwidth}
        \includegraphics[width=\textwidth]{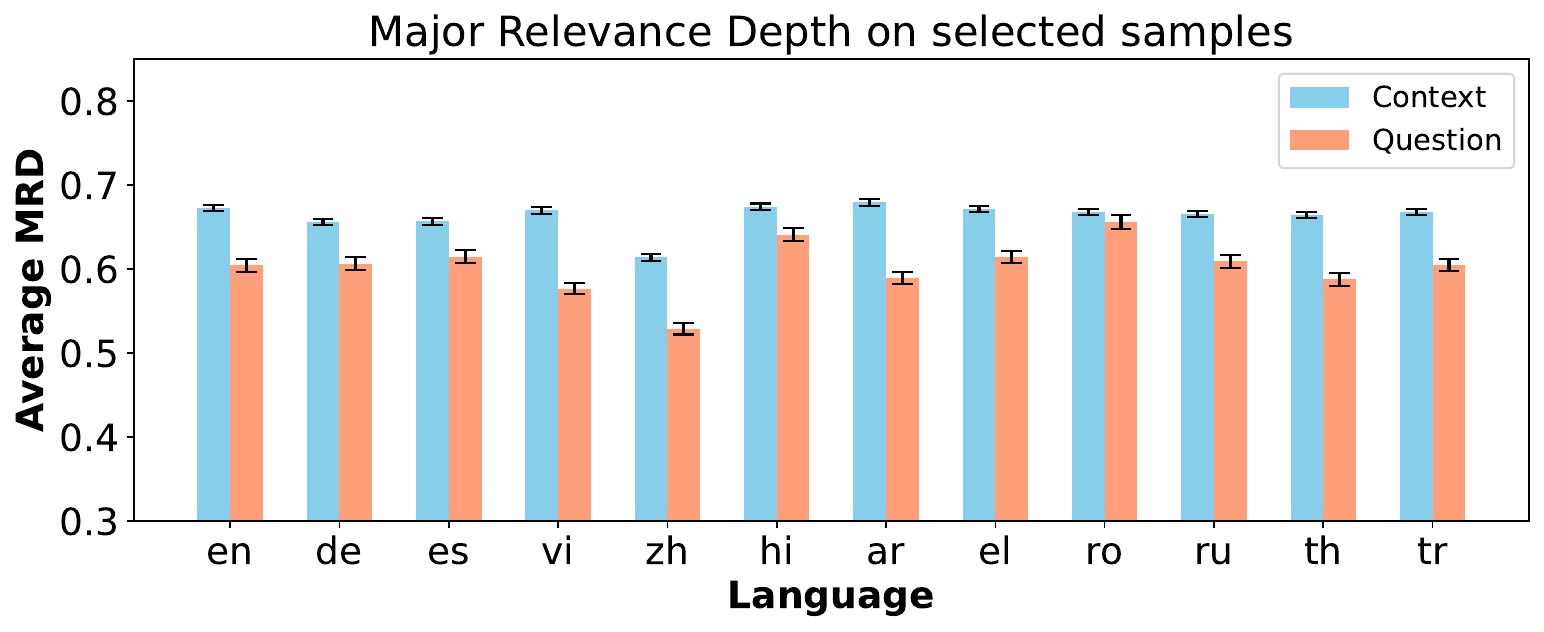}
        \caption{en-superior samples}
        \label{llama-3.1-base-8b-en-sup-cq}
    \end{subfigure}
    \caption{Mean MRD of the context and question parts for LLaMA-3.1-Base-8B.}
    \label{llama-3.1-base-8b-cq}
\end{figure}

\begin{figure}[ht]
    \centering
    \begin{subfigure}{0.45\textwidth}
        \includegraphics[width=\textwidth]{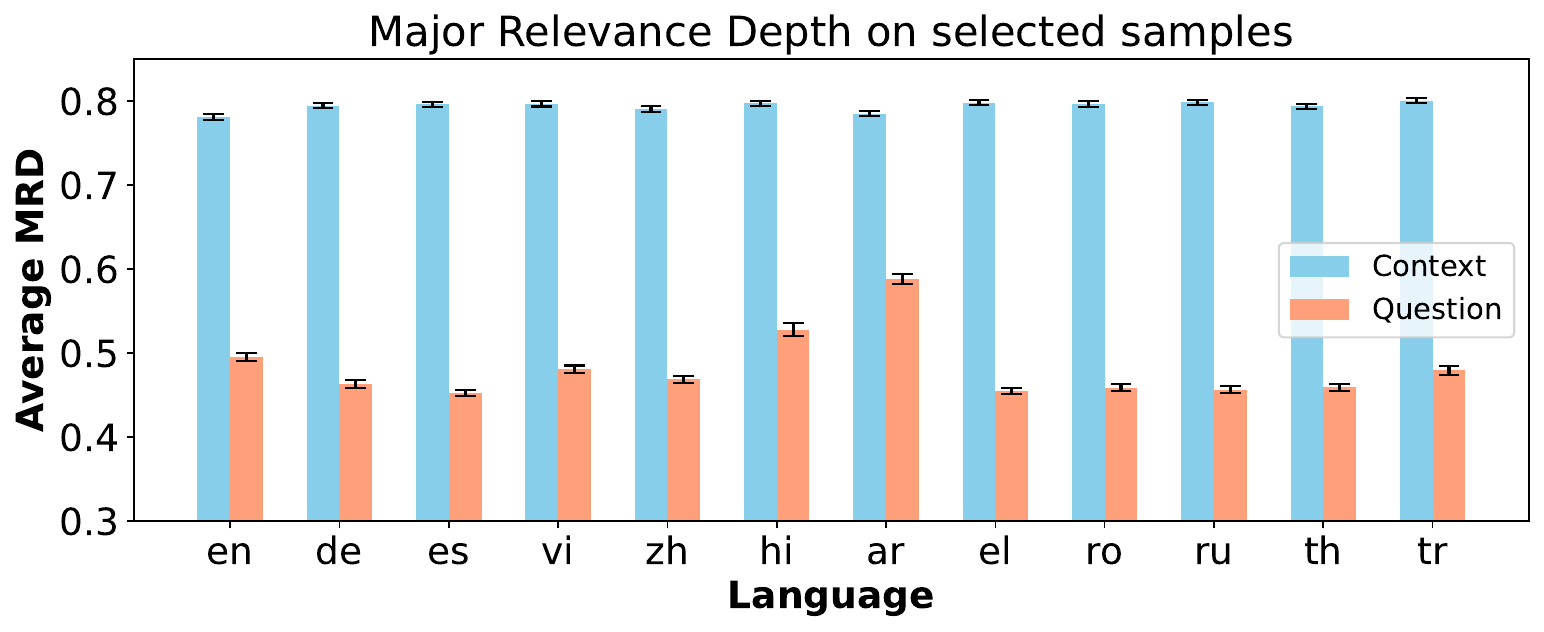}
        \caption{balanced samples}
        \label{llama-3.1-instruct-70b-balanced-cq}
    \end{subfigure}
    \hfill
    \begin{subfigure}{0.45\textwidth}
        \includegraphics[width=\textwidth]{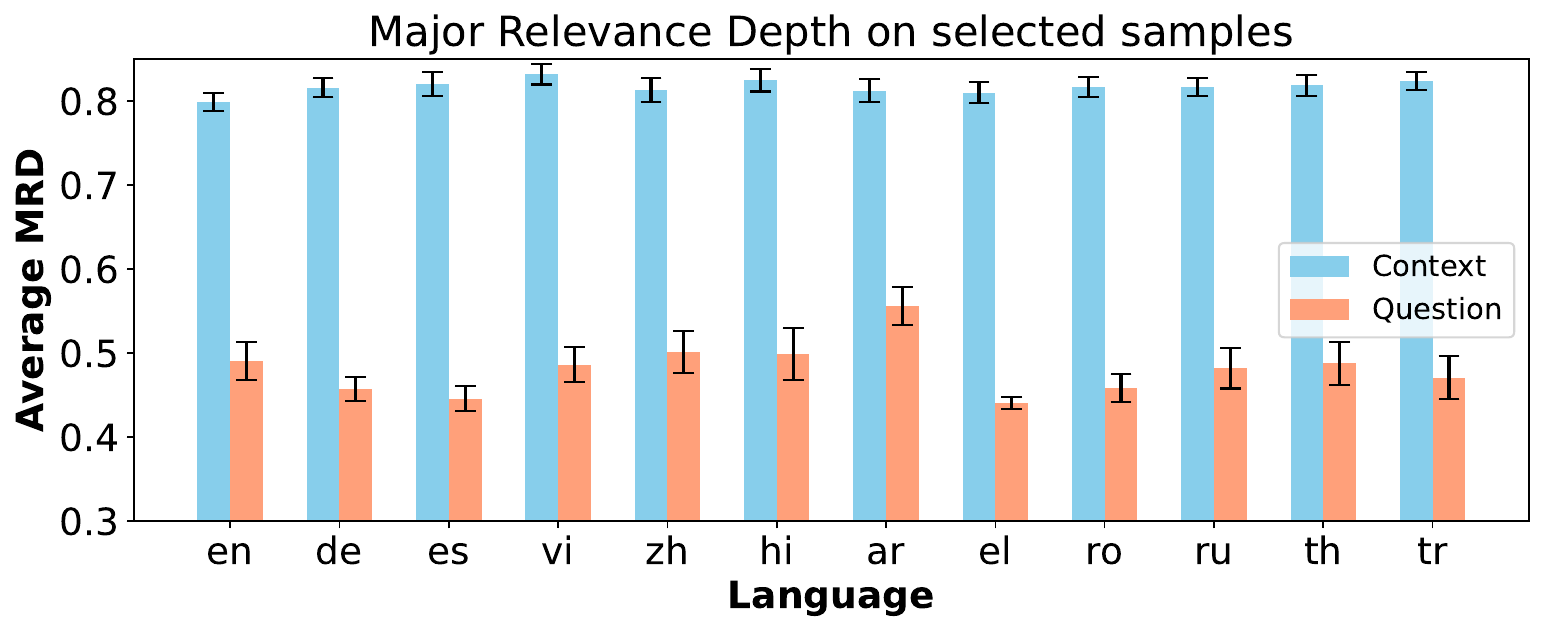}
        \caption{en-superior samples}
        \label{llama-3.1-instruct-70b-en-sup-cq}
    \end{subfigure}
    \caption{Mean MRD of the context and question parts for LLaMA-3.1-Instruct-70B.}
    \label{llama-3.1-instruct-70b-cq}
\end{figure}

\begin{figure}[ht]
    \centering
    \begin{subfigure}{0.45\textwidth}
        \includegraphics[width=\textwidth]{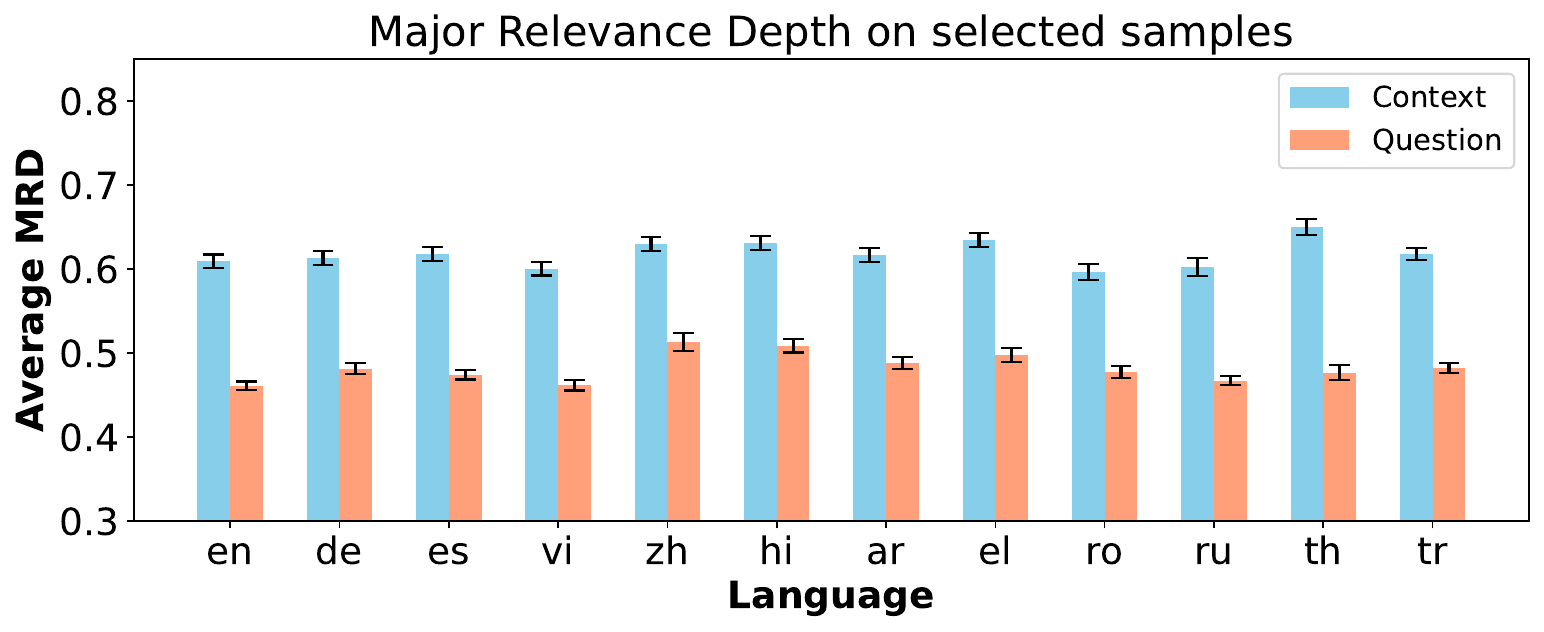}
        \caption{balanced samples}
        \label{llama-2-chat-7b-balanced-cq}
    \end{subfigure}
    \hfill
    \begin{subfigure}{0.45\textwidth}
        \includegraphics[width=\textwidth]{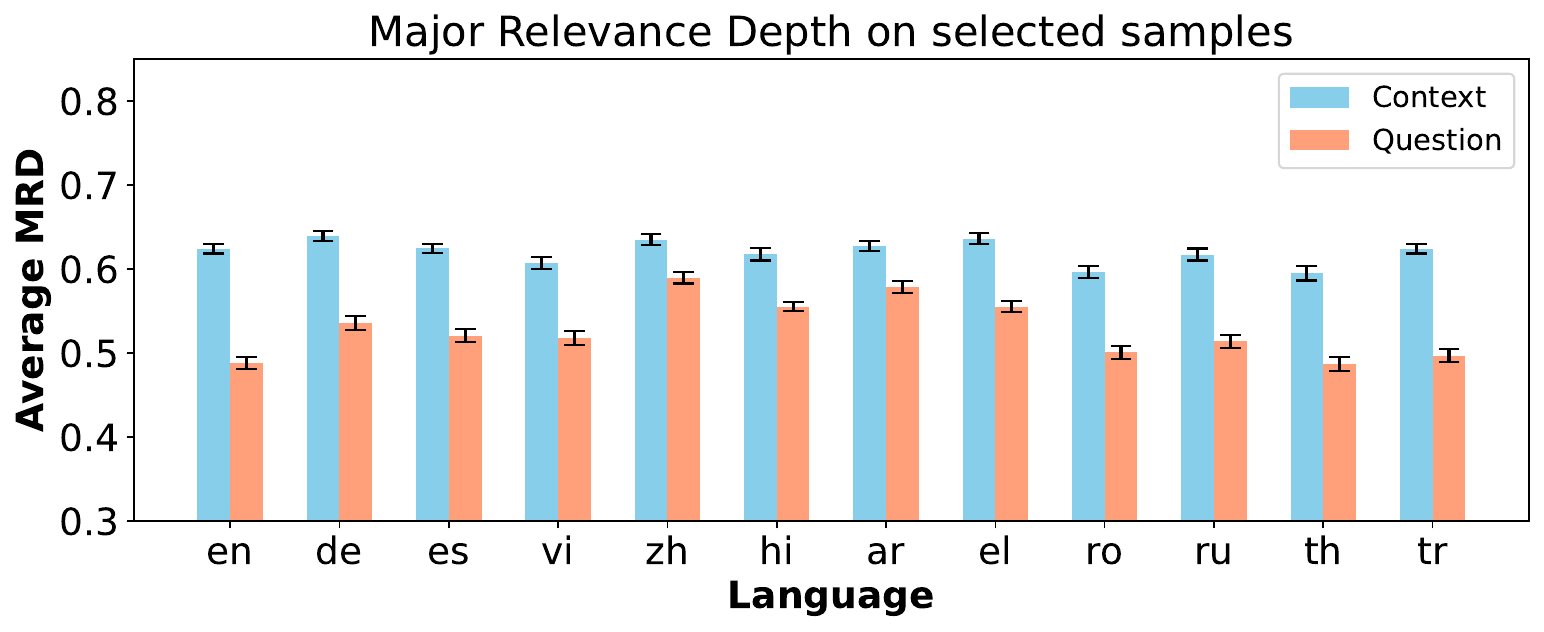}
        \caption{en-superior samples}
        \label{llama-2-chat-7b-en-sup-cq}
    \end{subfigure}
    \caption{Mean MRD of the context and question parts for LLaMA-2-Chat-7B.}
    \label{llama-2-chat-7b-cq}
\end{figure}

Figures \ref{llama-3.1-base-8b-cq}, \ref{llama-3.1-instruct-70b-cq} and \ref{llama-2-chat-7b-cq} provide further illustrative examples of the mean MRD for context and question components, specifically for LLaMA-3.1-8B, LLaMA-3.1-Instruct-70B and LLaMA-2-Chat-7B. These figures complement the MRD analysis presented in the main body of this paper.

\subsubsection{Analysis of task descriptions and demonstrations}
\label{appendix:task_descriptions_demos_discussion}

\begin{figure}[ht]
    \centering
    \begin{subfigure}{0.45\textwidth}
        \includegraphics[width=\textwidth]{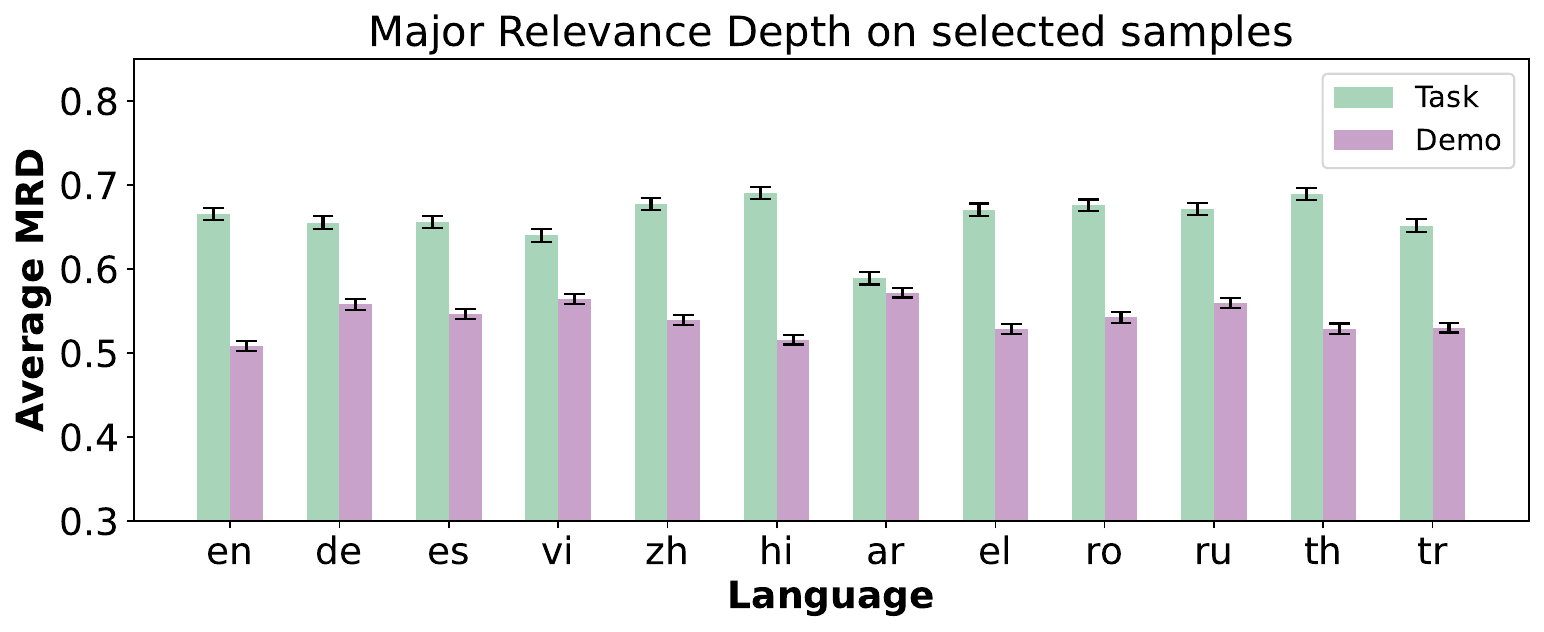}
        \caption{balanced samples}
        \label{llama-3.1-instruct-8b-balanced-td}
    \end{subfigure}
    \hfill
    \begin{subfigure}{0.45\textwidth}
        \includegraphics[width=\textwidth]{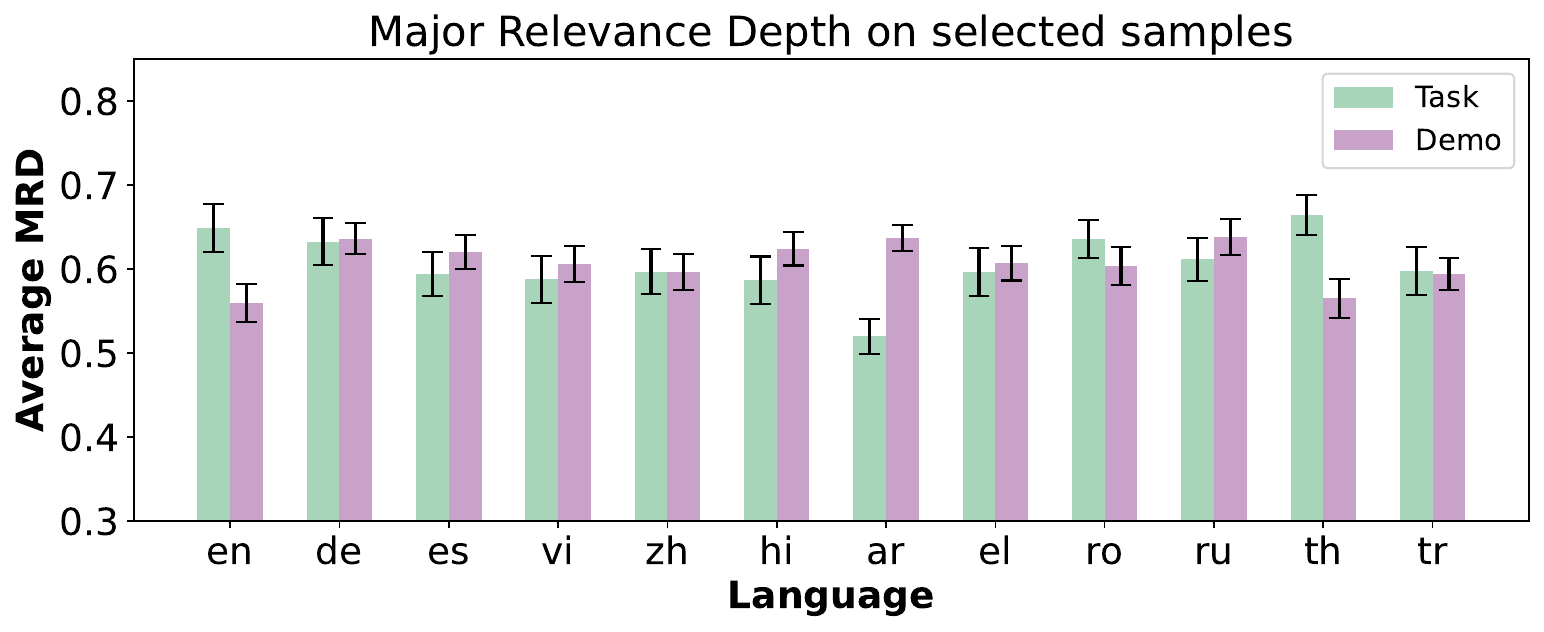}
        \caption{en-superior samples}
        \label{llama-3.1-instruct-8b-en-sup-td}
    \end{subfigure}
    \caption{Mean MRD of the task descriptions and demonstrations parts for LLaMA-3.1-Instruct-8B.}
    \label{llama-3.1-instruct-8b-td}
\end{figure}

\begin{figure}[ht]
    \centering
    \begin{subfigure}{0.45\textwidth}
        \includegraphics[width=\textwidth]{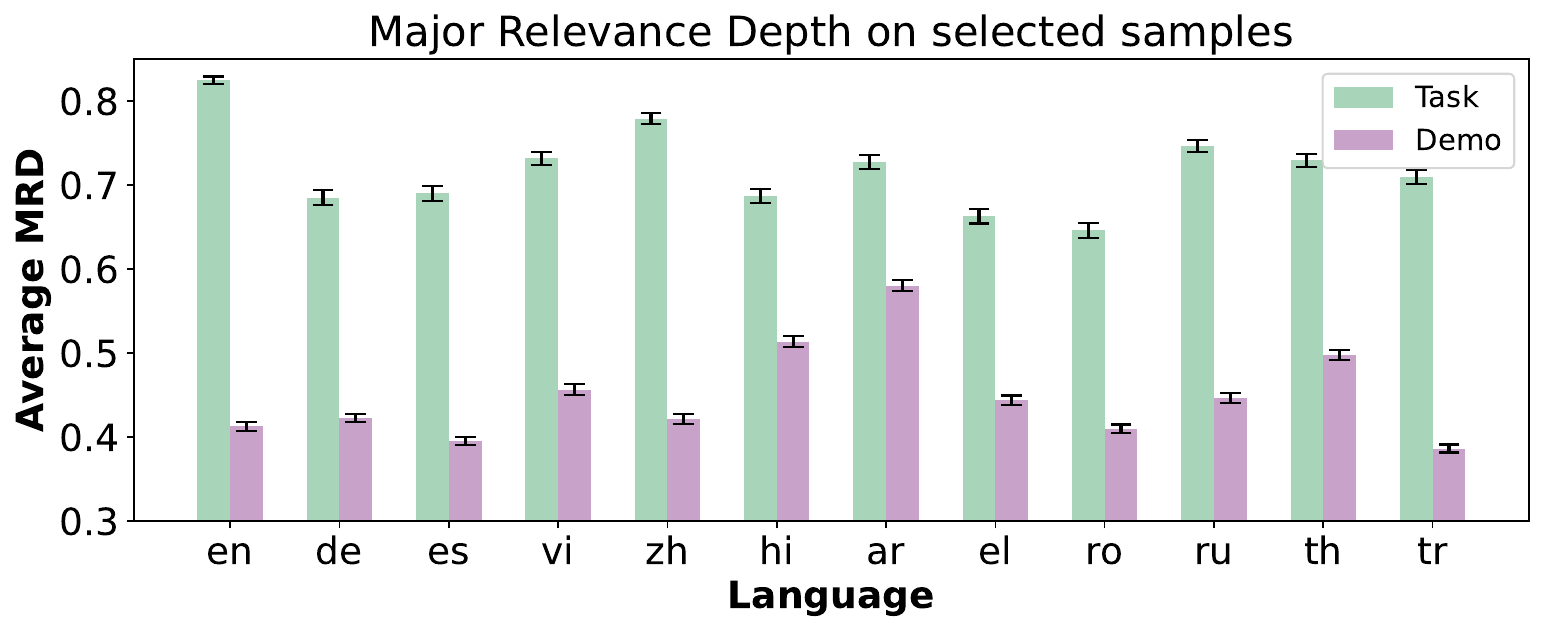}
        \caption{balanced samples}
        \label{llama-3.1-instruct-70b-balanced-td}
    \end{subfigure}
    \hfill
    \begin{subfigure}{0.45\textwidth}
        \includegraphics[width=\textwidth]{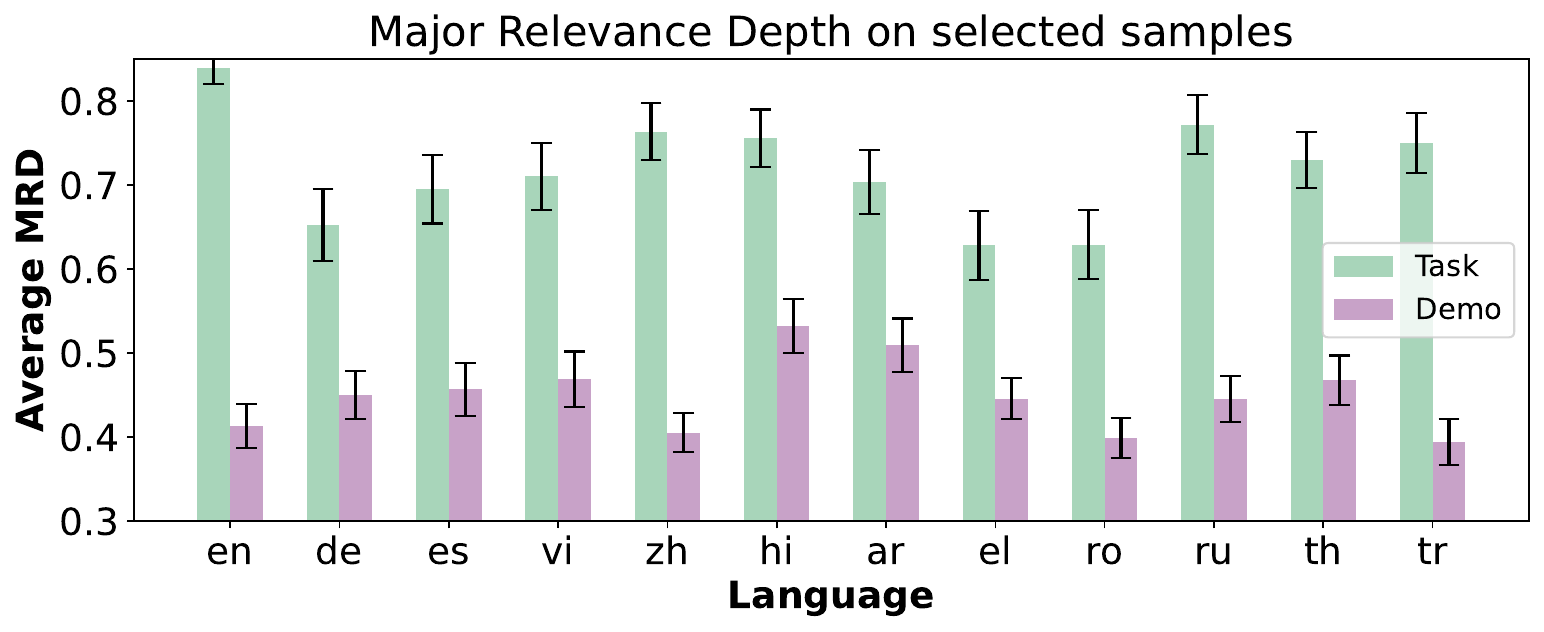}
        \caption{en-superior samples}
        \label{llama-3.1-instruct-70b-en-sup-td}
    \end{subfigure}
    \caption{Mean MRD of the task descriptions and demonstrations parts for LLaMA-3.1-Instruct-70B.}
    \label{llama-3.1-instruct-70b-td}
\end{figure}

\begin{figure}[ht]
    \centering
    \begin{subfigure}{0.45\textwidth}
        \includegraphics[width=\textwidth]{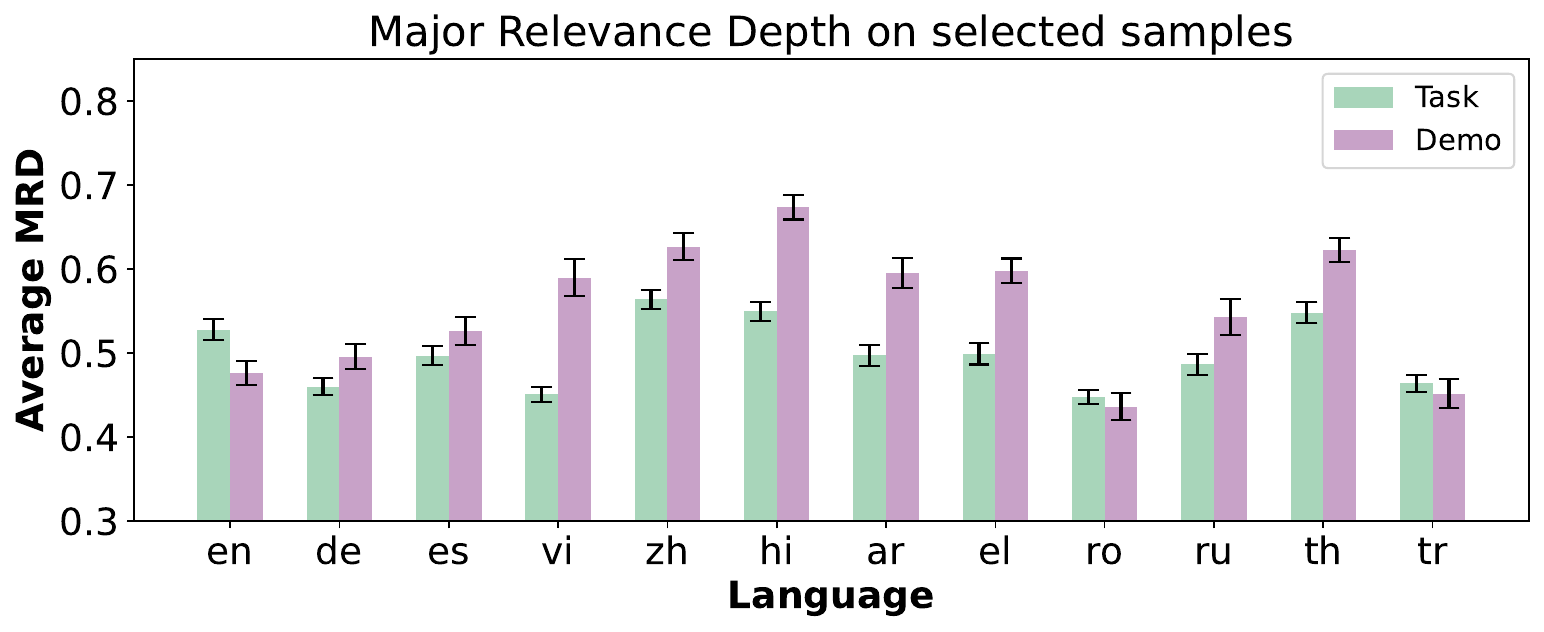}
        \caption{balanced samples}
        \label{llama-2-chat-7b-balanced-td}
    \end{subfigure}
    \hfill
    \begin{subfigure}{0.45\textwidth}
        \includegraphics[width=\textwidth]{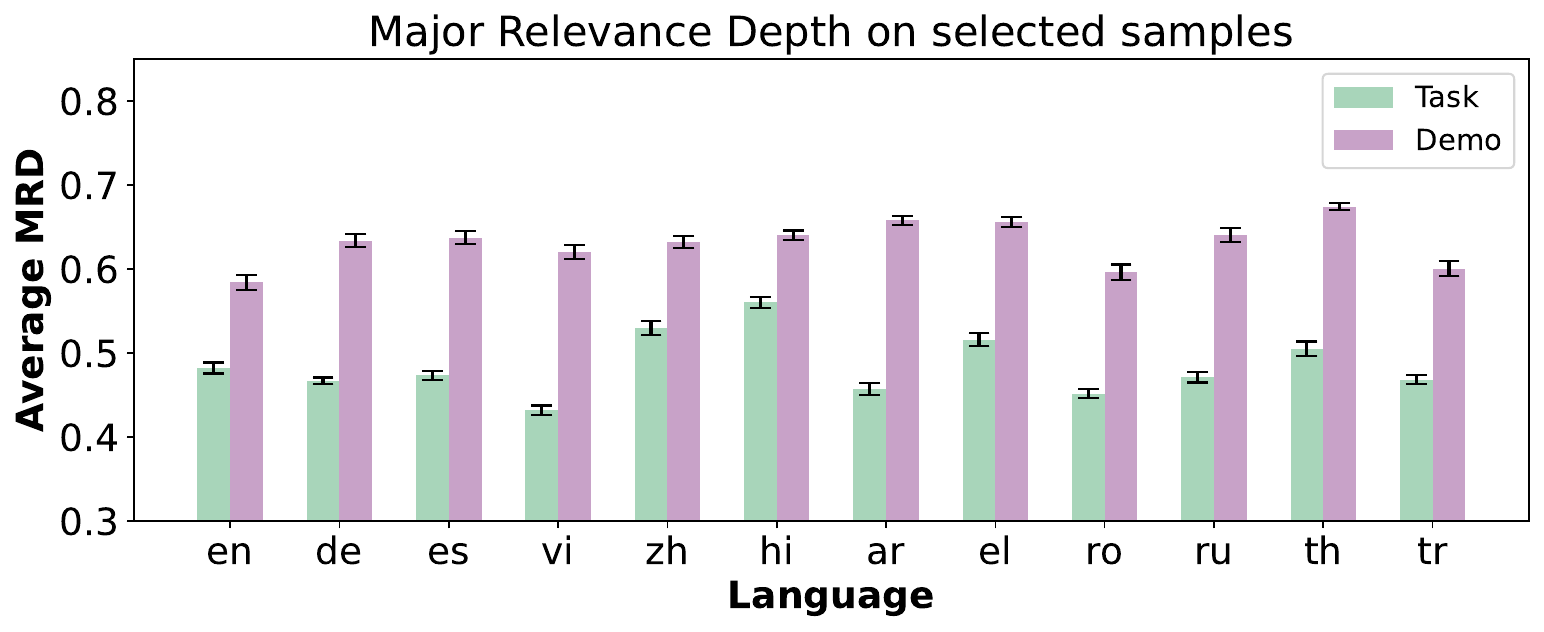}
        \caption{en-superior samples}
        \label{llama-2-chat-7b-en-sup-td}
    \end{subfigure}
    \caption{Mean MRD of the task descriptions and demonstrations parts for LLaMA-2-Chat-7B.}
    \label{llama-2-chat-7b-td}
\end{figure}

Analyzing the MRD of task descriptions and demonstrations in our 2-shot setting (Figures \ref{llama-3.1-instruct-8b-td}-\ref{llama-2-chat-7b-td}) reveals a general trend where demonstrations tend to exhibit a comparable or slightly higher MRD than task descriptions across the LLaMA model family, suggesting demonstrations are at least as important as, if not slightly more impactful than, task descriptions in guiding the models. This could indicate that providing concrete examples is a particularly effective way to communicate the desired behavior for cross-lingual context retrieval to these models.

However, the precise relationship is not uniform and varies across models. For example, while LLaMA-3.1-Instruct-8B shows a relatively balanced MRD between task descriptions and demonstrations, LLaMA-2-Chat-7B consistently demonstrates a higher MRD for demonstrations, which implies that older or smaller models might lean more heavily on the provided in-context examples. In contrast, LLaMA-3.1-Instruct-70B exhibits the most pronounced difference, with a significantly elevated MRD for task descriptions across all languages and sample types, suggesting that larger models can become highly attuned to and reliant on user-specified task commands. 

\subsubsection{Influence of prompt format on MRD pattern}
\label{appendix:prompt_formats_influence_on_MRD}

\begin{figure}[!htb]
    \centering
    \begin{subfigure}{0.45\textwidth}
        \includegraphics[width=\textwidth]{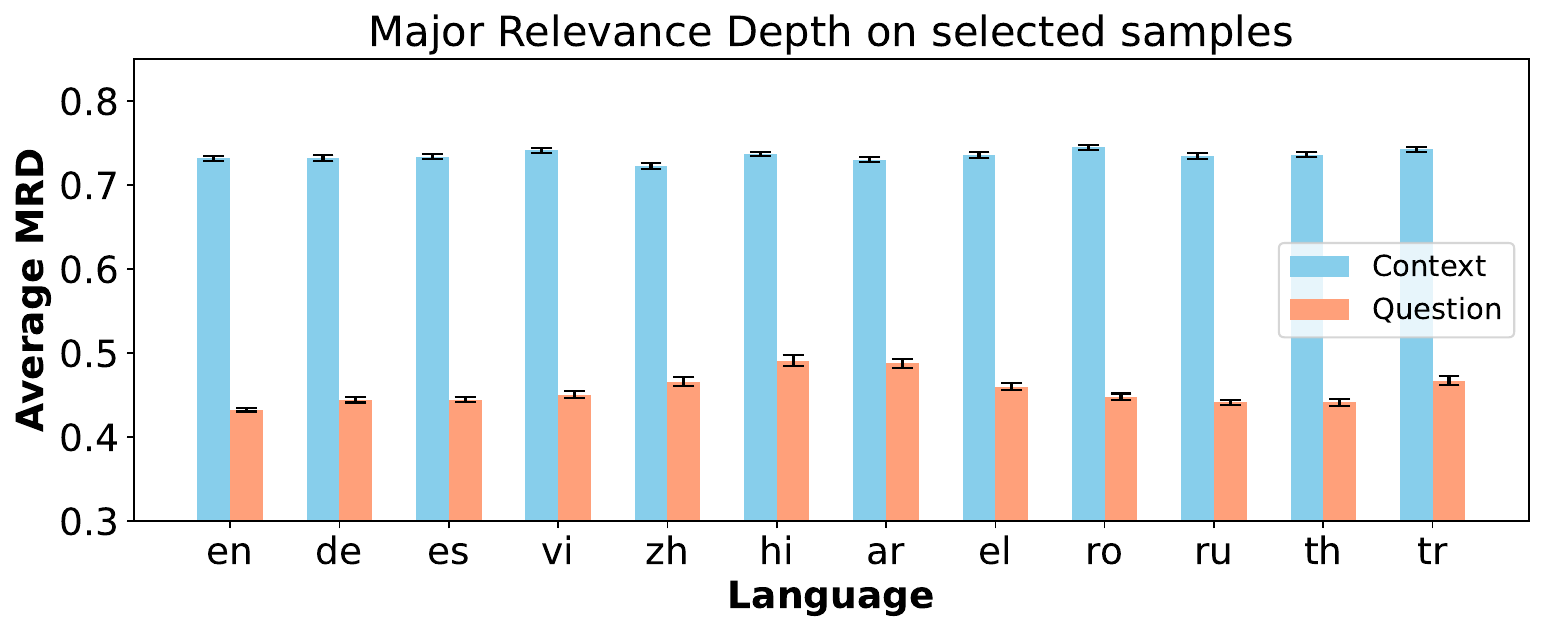}
        \caption{balanced samples}
        \label{llama-3.1-instruct-8b-v1-balanced-cq}
    \end{subfigure}
    \hfill
    \begin{subfigure}{0.45\textwidth}
        \includegraphics[width=\textwidth]{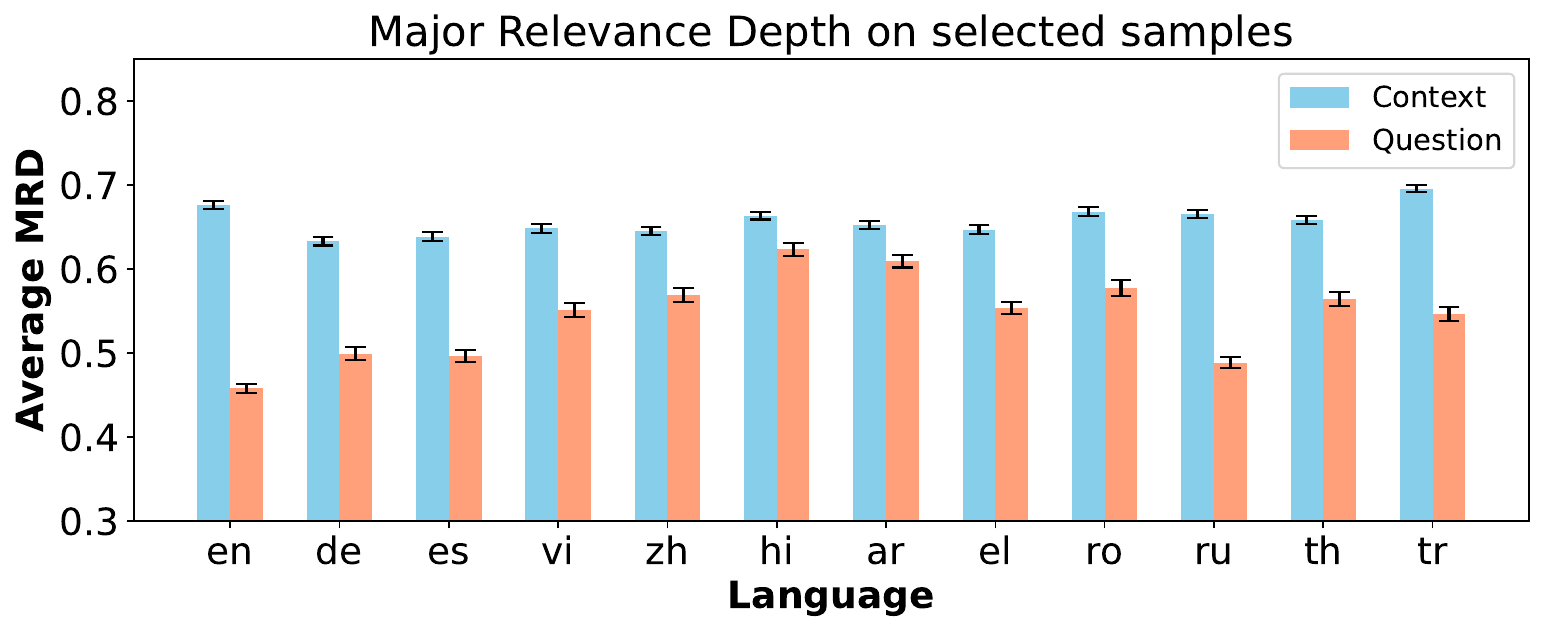}
        \caption{en-superior samples}
        \label{llama-3.1-instruct-8b-v1-en-sup-cq}
    \end{subfigure}
    \caption{Mean MRD for LLaMA-3.1-Instruct-8B on both ``balanced'' and ``en-superior'' samples in v1 prompting format. Only the results of context and question parts of the prompt are displayed.}
    \label{llama-3.1-instruct-8b-v1-cq}
\end{figure}

We test the influence of different prompt formats (v1, v2) on LLaMA-3.1-Instruct-8B, and by comparing the results in Figure \ref{llama-3.1-instruct-8b-v1-cq} and Figure \ref{llama-3.1-instruct-8b-cq}, which present results obtained using the prompt format v1 and v2, respectively, it is clear that the fundamental pattern observed in the mean MRD is consistent across both formats. Therefore, the trend of the mean question MRD being consistently and substantially lower than the mean context MRD is maintained regardless of the prompt format employed.

\subsection{Hidden state similarity results for other LLaMA models}
\label{appendix:other_hidden_sim}

\begin{figure*}[ht]
    \centering
    \begin{subfigure}{0.45\textwidth}
        \includegraphics[width=\textwidth]{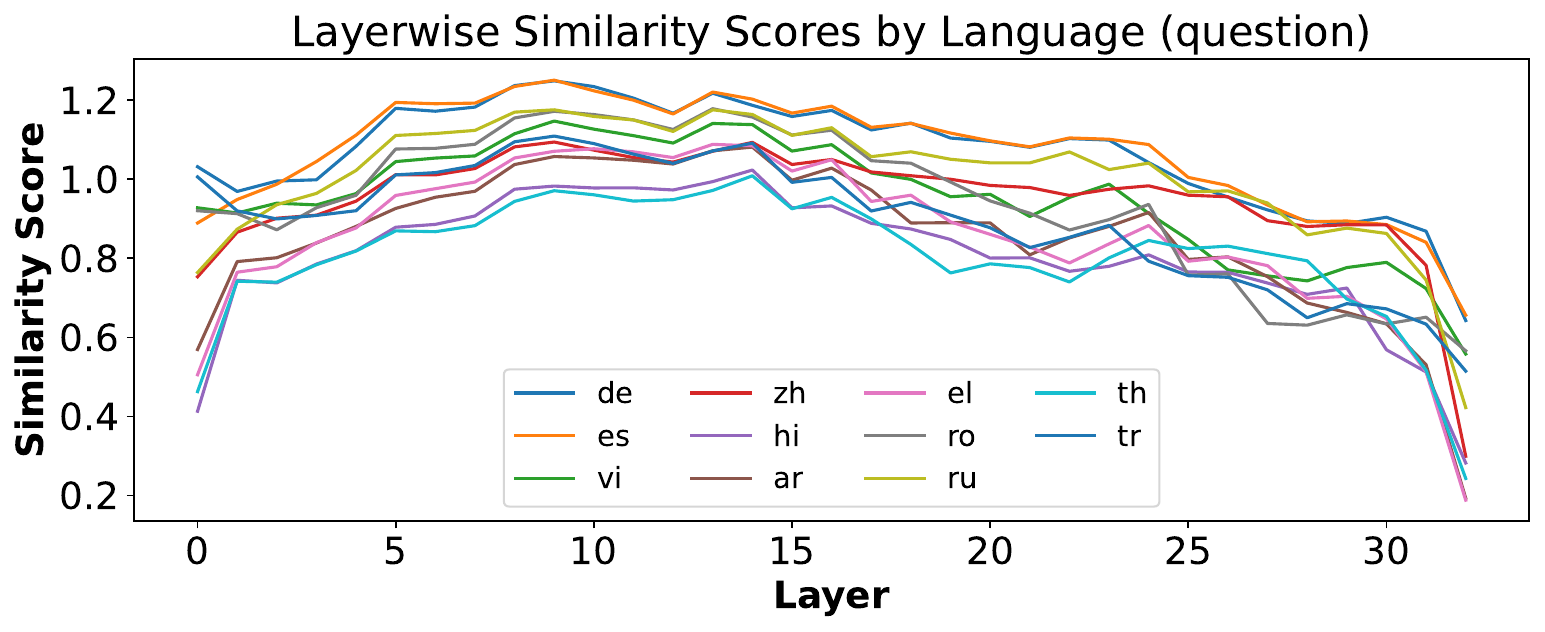}
        \caption{Question hidden state similarity for balanced samples.}
    \end{subfigure}
    \hfill
    \begin{subfigure}{0.45\textwidth}
        \includegraphics[width=\textwidth]{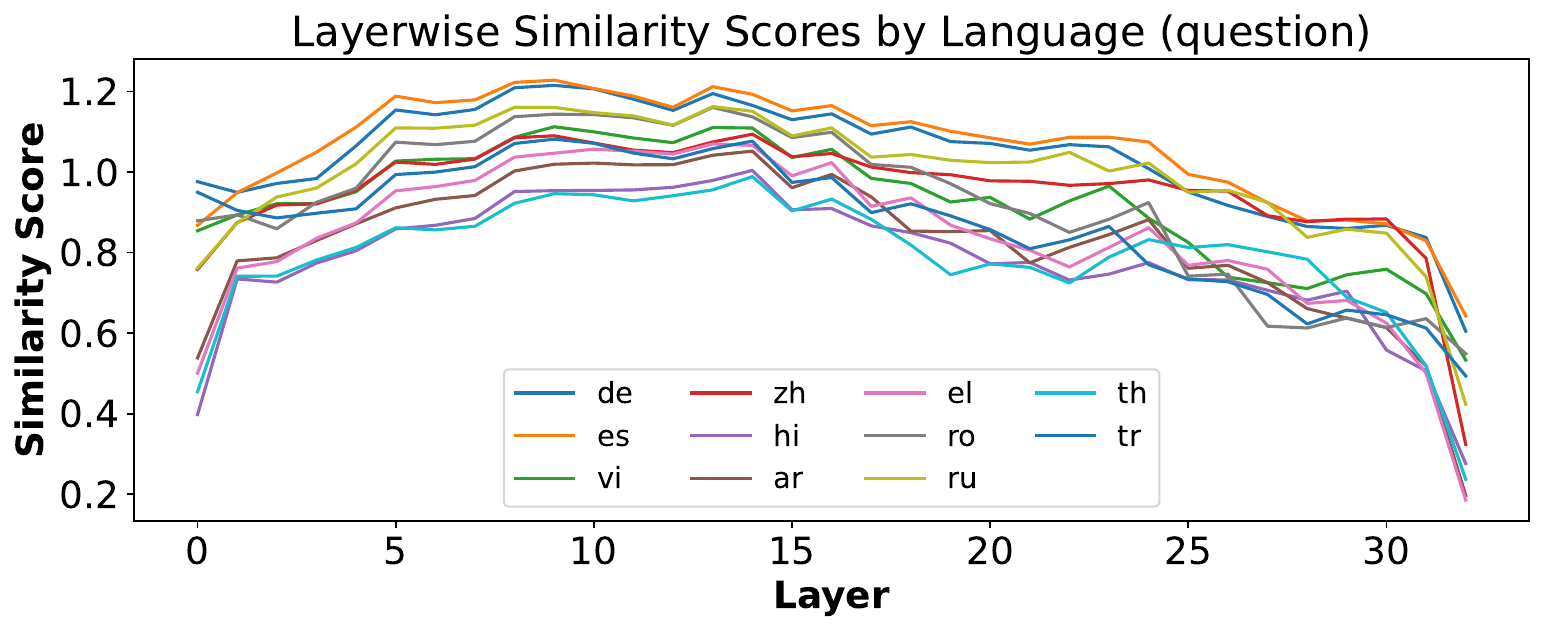}
        \caption{Question hidden state similarity for en-superior samples.}
    \end{subfigure}
    
    \vspace{0.5cm}
    
    \begin{subfigure}{0.45\textwidth}
        \includegraphics[width=\textwidth]{./assets/llama-3.1-base-8b-2-shot-balanced-c.pdf}
        \caption{Context hidden state similarity for balanced samples.}
    \end{subfigure}
    \hfill
    \begin{subfigure}{0.45\textwidth}
        \includegraphics[width=\textwidth]{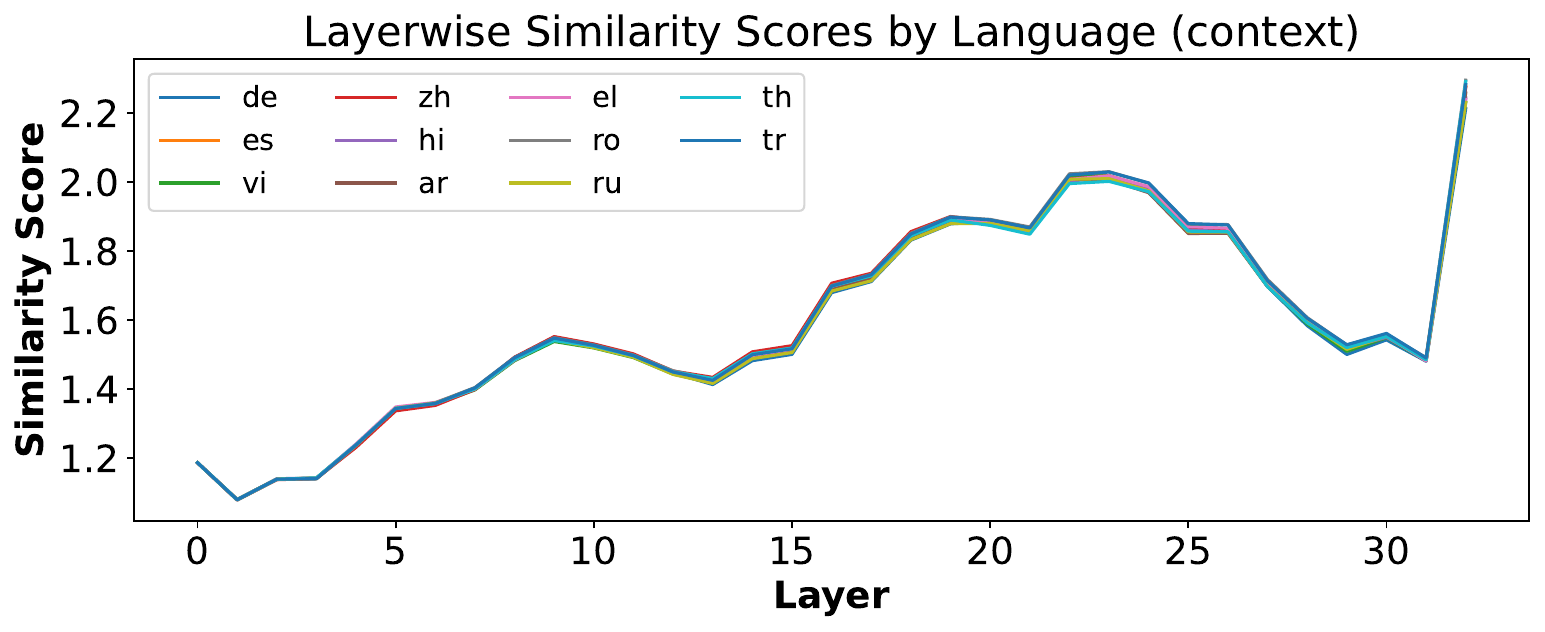}
        \caption{Context hidden state similarity for en-superior samples.}
    \end{subfigure}
    
    \vspace{0.5cm}
    
    \begin{subfigure}{0.45\textwidth}
        \includegraphics[width=\textwidth]{./assets/llama-3.1-base-8b-2-shot-balanced-l.pdf}
        \caption{Last-input-token hidden state similarity for balanced samples.}
    \end{subfigure}
    \hfill
    \begin{subfigure}{0.45\textwidth}
        \includegraphics[width=\textwidth]{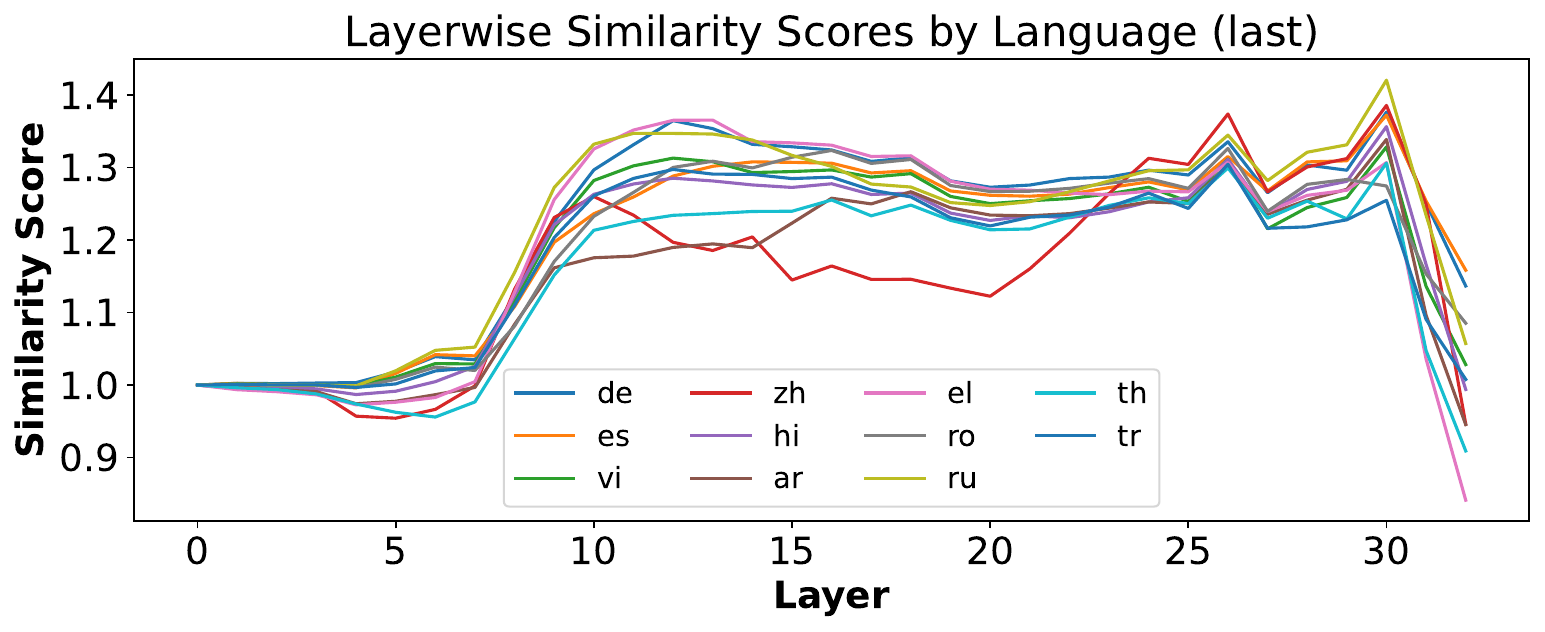}
        \caption{Last-input-token hidden state similarity for en-superior samples.}
    \end{subfigure}
    
    \caption{Hidden state similarity between English and other languages on different parts of the selected samples in each layer of the LLaMA-3.1-8B model.}
    \label{llama-3.1-base-8b-2-shot-hidden-all}
\end{figure*}

\begin{figure*}[ht]
    \centering
    \begin{subfigure}{0.45\textwidth}
        \includegraphics[width=\textwidth]{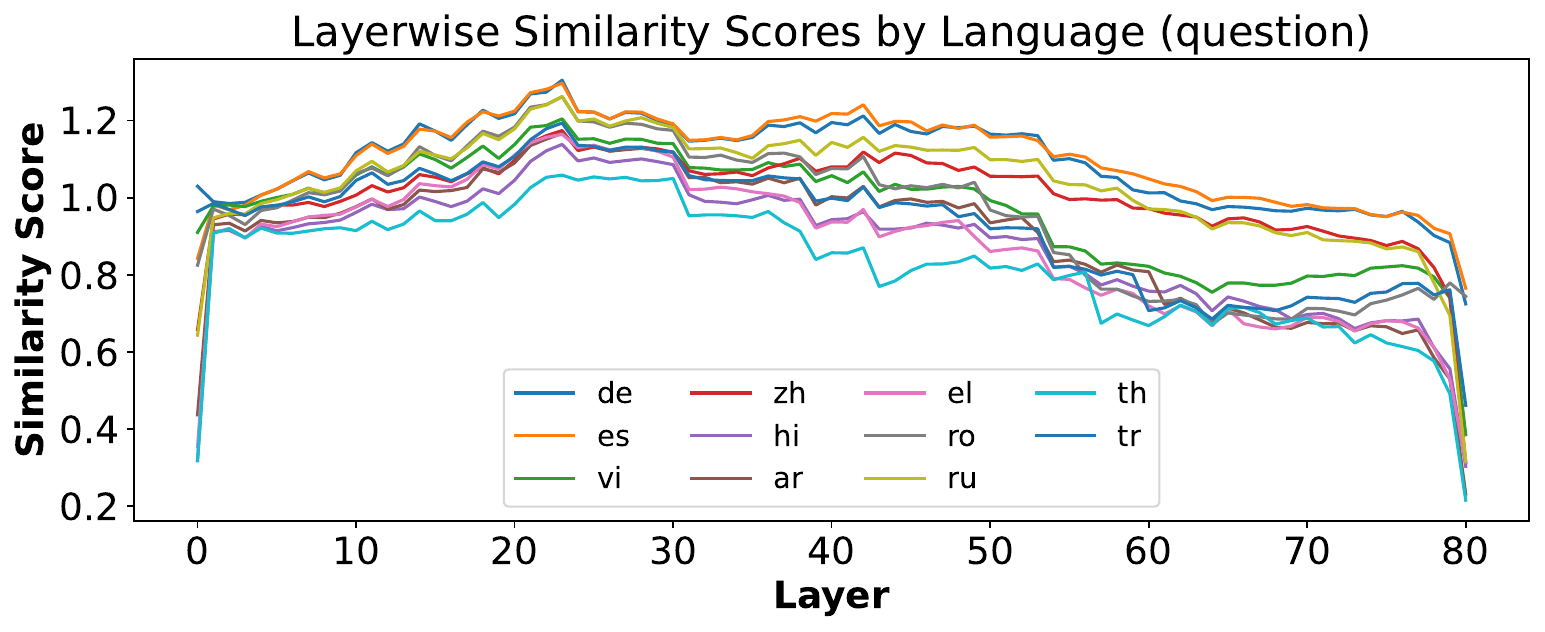}
        \caption{Question hidden state similarity for balanced samples.}
    \end{subfigure}
    \hfill
    \begin{subfigure}{0.45\textwidth}
        \includegraphics[width=\textwidth]{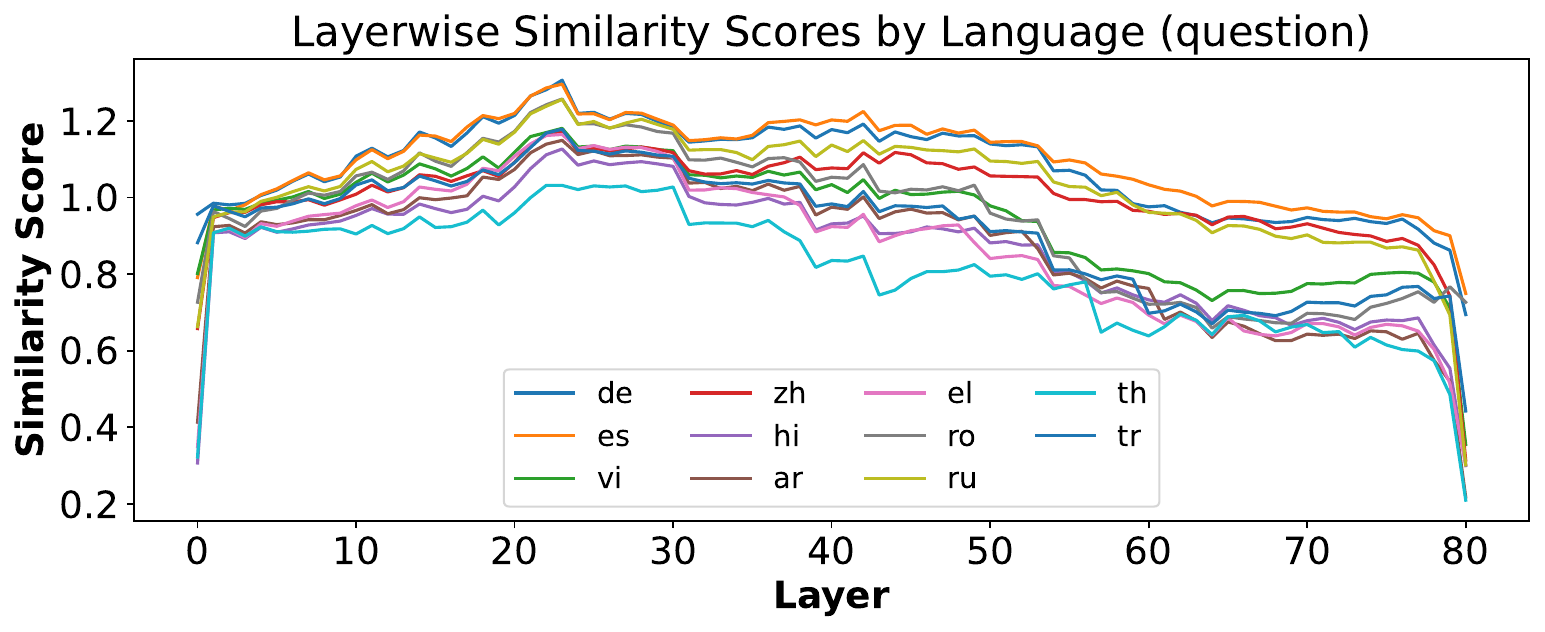}
        \caption{Question hidden state similarity for en-superior samples.}
    \end{subfigure}
    
    \vspace{0.5cm}
    
    \begin{subfigure}{0.45\textwidth}
        \includegraphics[width=\textwidth]{./assets/llama-3.1-base-70b-2-shot-balanced-c.pdf}
        \caption{Context hidden state similarity for balanced samples.}
    \end{subfigure}
    \hfill
    \begin{subfigure}{0.45\textwidth}
        \includegraphics[width=\textwidth]{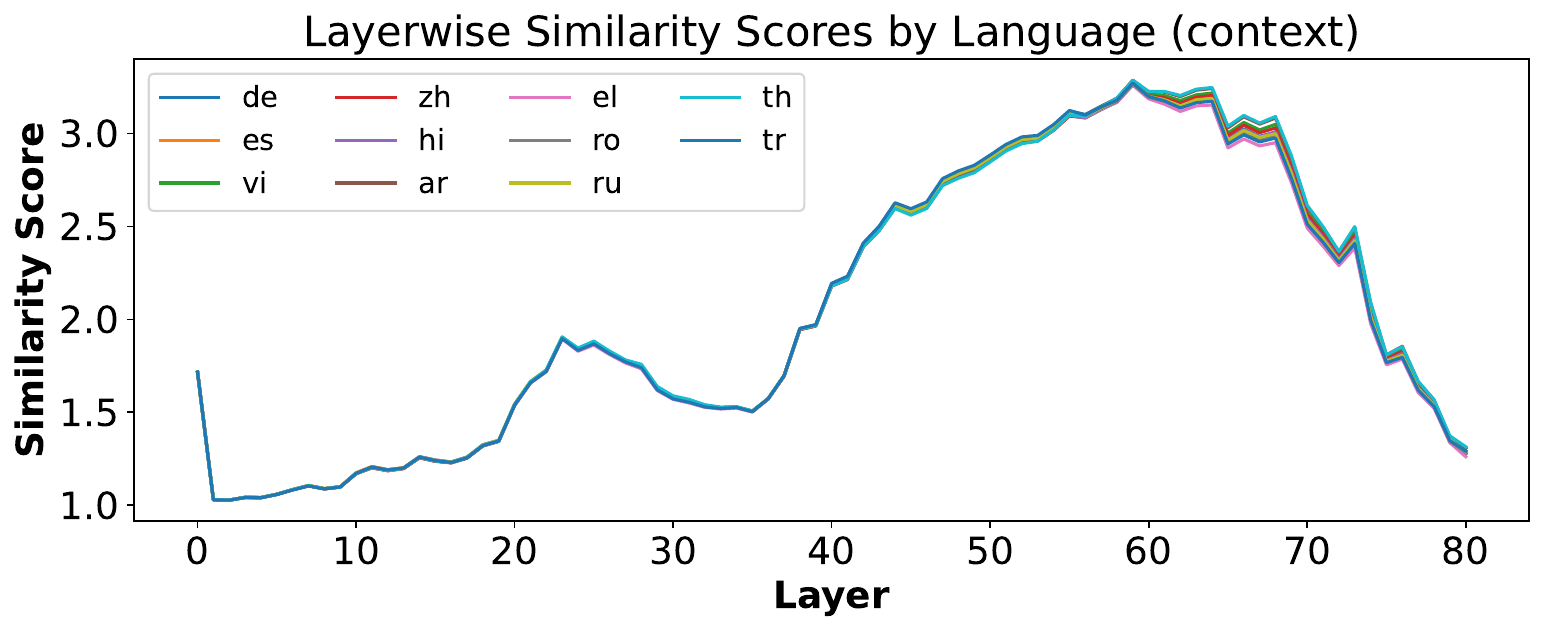}
        \caption{Context hidden state similarity for en-superior samples.}
    \end{subfigure}
    
    \vspace{0.5cm}
    
    \begin{subfigure}{0.45\textwidth}
        \includegraphics[width=\textwidth]{./assets/llama-3.1-base-70b-2-shot-balanced-l.pdf}
        \caption{Last-input-token hidden state similarity for balanced samples.}
    \end{subfigure}
    \hfill
    \begin{subfigure}{0.45\textwidth}
        \includegraphics[width=\textwidth]{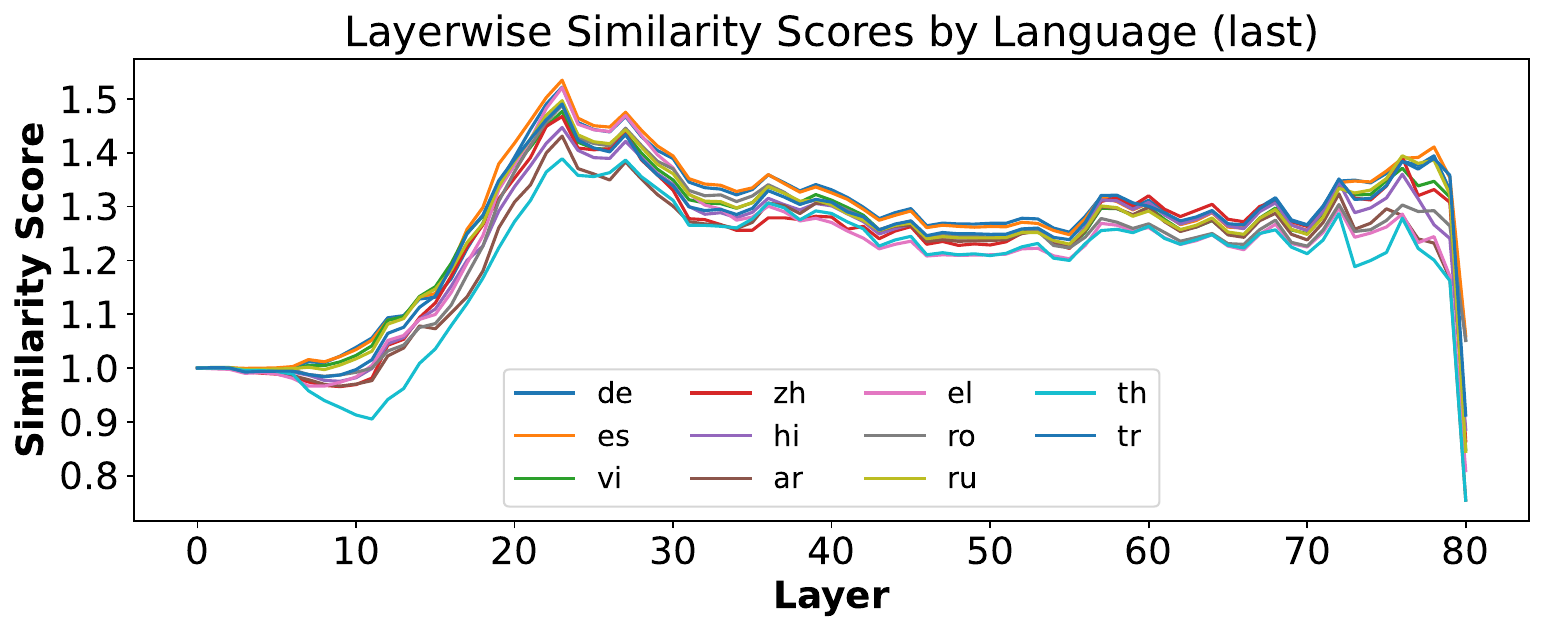}
        \caption{Last-input-token hidden state similarity for en-superior samples.}
    \end{subfigure}
    
    \caption{Hidden state similarity between English and other languages on different parts of the selected samples in each layer of the LLaMA-3.1-70B model.}
    \label{llama-3.1-base-70b-2-shot-hidden-all}
\end{figure*}

\begin{figure*}[ht]
    \centering
    \begin{subfigure}{0.45\textwidth}
        \includegraphics[width=\textwidth]{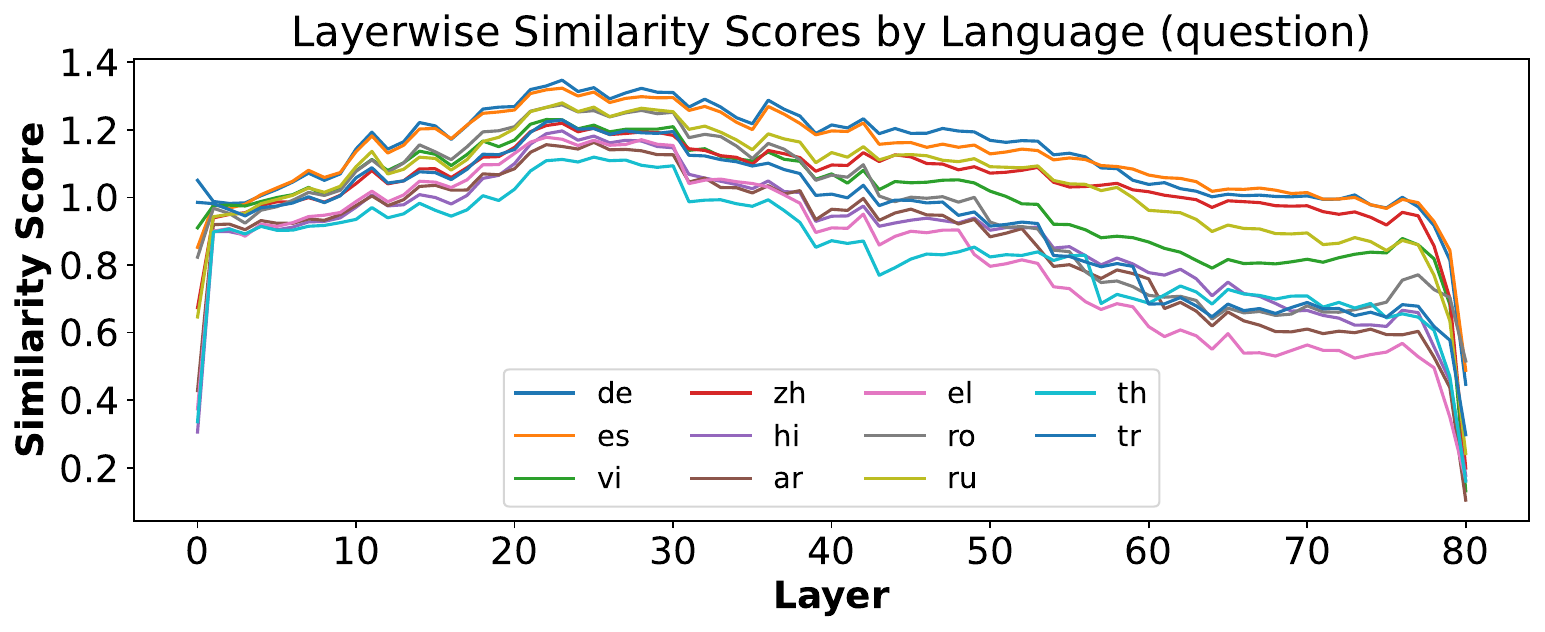}
        \caption{Question hidden state similarity for balanced samples.}
    \end{subfigure}
    \hfill
    \begin{subfigure}{0.45\textwidth}
        \includegraphics[width=\textwidth]{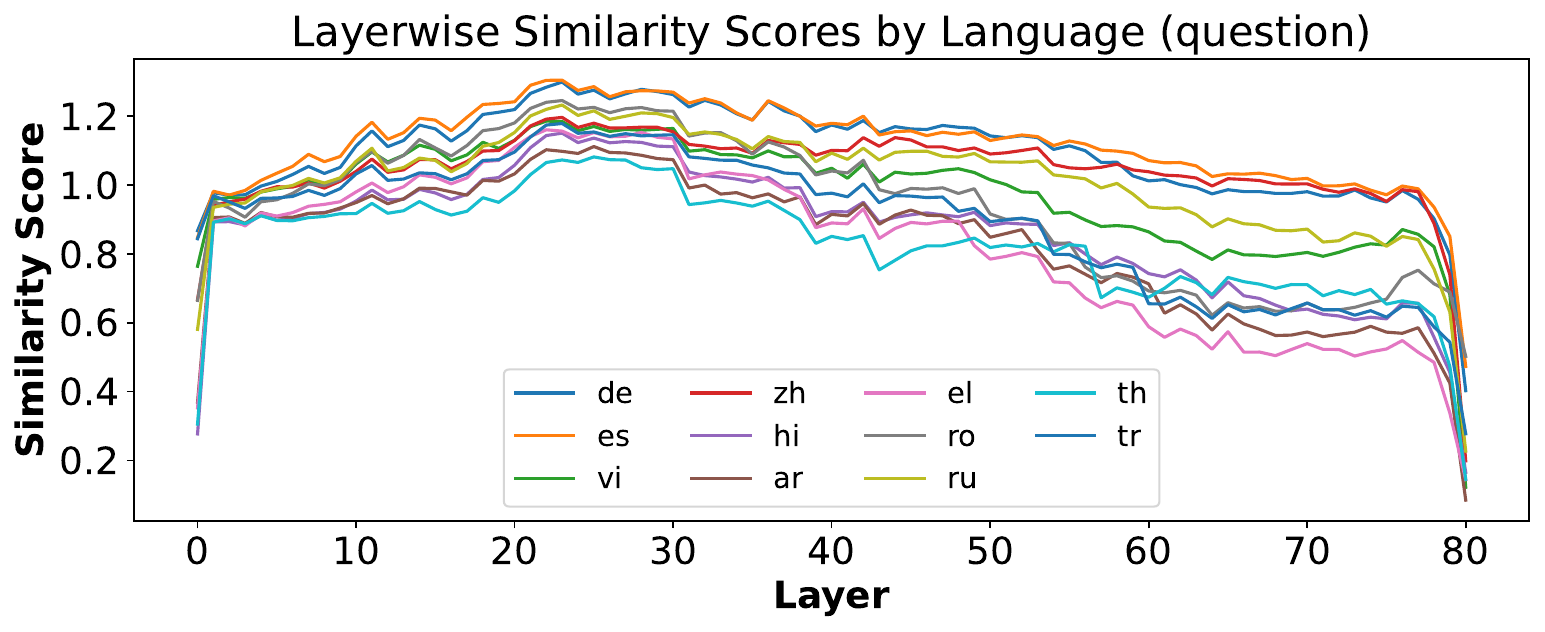}
        \caption{Question hidden state similarity for en-superior samples.}
    \end{subfigure}
    
    \vspace{0.5cm}
    
    \begin{subfigure}{0.45\textwidth}
        \includegraphics[width=\textwidth]{./assets/llama-3.1-instruct-70b-2-shot-balanced-c.pdf}
        \caption{Context hidden state similarity for balanced samples.}
    \end{subfigure}
    \hfill
    \begin{subfigure}{0.45\textwidth}
        \includegraphics[width=\textwidth]{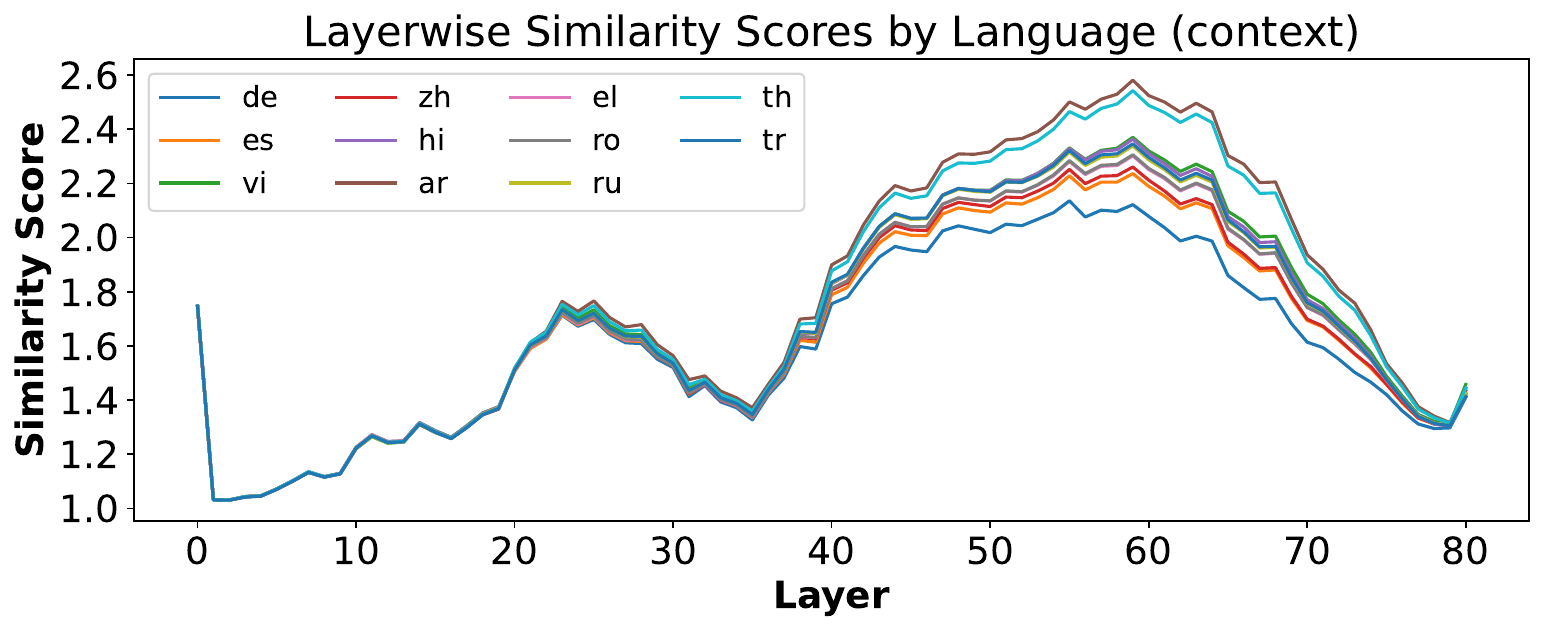}
        \caption{Context hidden state similarity for en-superior samples.}
    \end{subfigure}
    
    \vspace{0.5cm}
    
    \begin{subfigure}{0.45\textwidth}
        \includegraphics[width=\textwidth]{./assets/llama-3.1-instruct-70b-2-shot-balanced-l.pdf}
        \caption{Last-input-token hidden state similarity for balanced samples.}
    \end{subfigure}
    \hfill
    \begin{subfigure}{0.45\textwidth}
        \includegraphics[width=\textwidth]{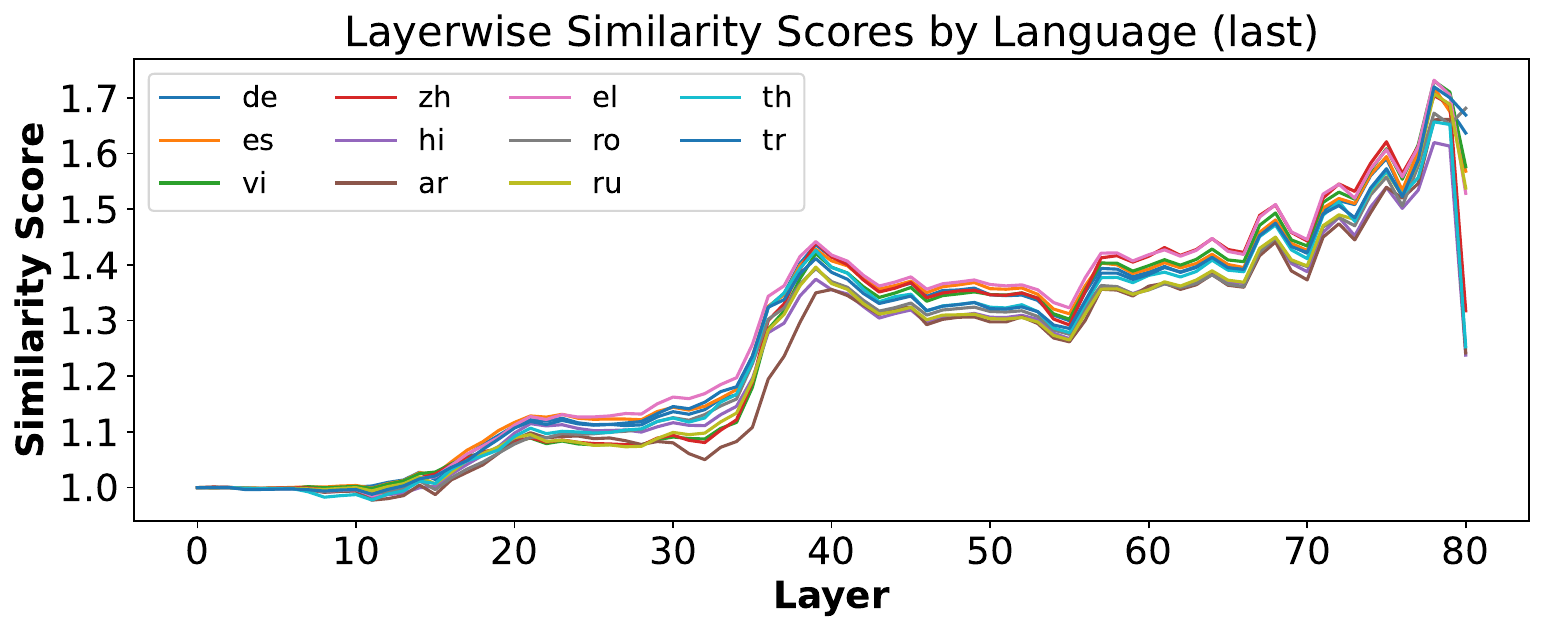}
        \caption{Last-input-token hidden state similarity for en-superior samples.}
    \end{subfigure}
    
    \caption{Hidden state similarity between English and other languages on different parts of the selected samples in each layer of the LLaMA-3.1-Instruct-70B model.}
    \label{llama-3.1-instruct-70b-2-shot-hidden-all}
\end{figure*}

\begin{figure*}[ht]
    \centering
    \begin{subfigure}{0.45\textwidth}
        \includegraphics[width=\textwidth]{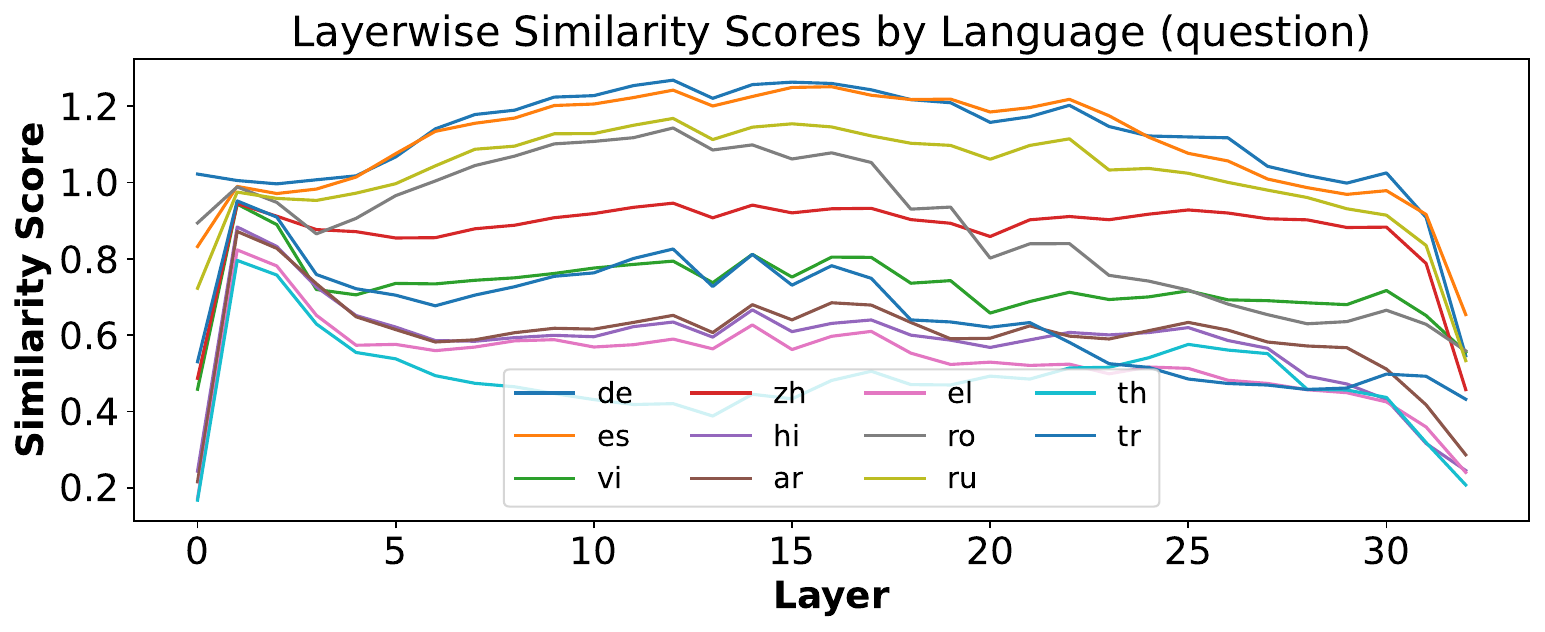}
        \caption{Question hidden state similarity for balanced samples.}
    \end{subfigure}
    \hfill
    \begin{subfigure}{0.45\textwidth}
        \includegraphics[width=\textwidth]{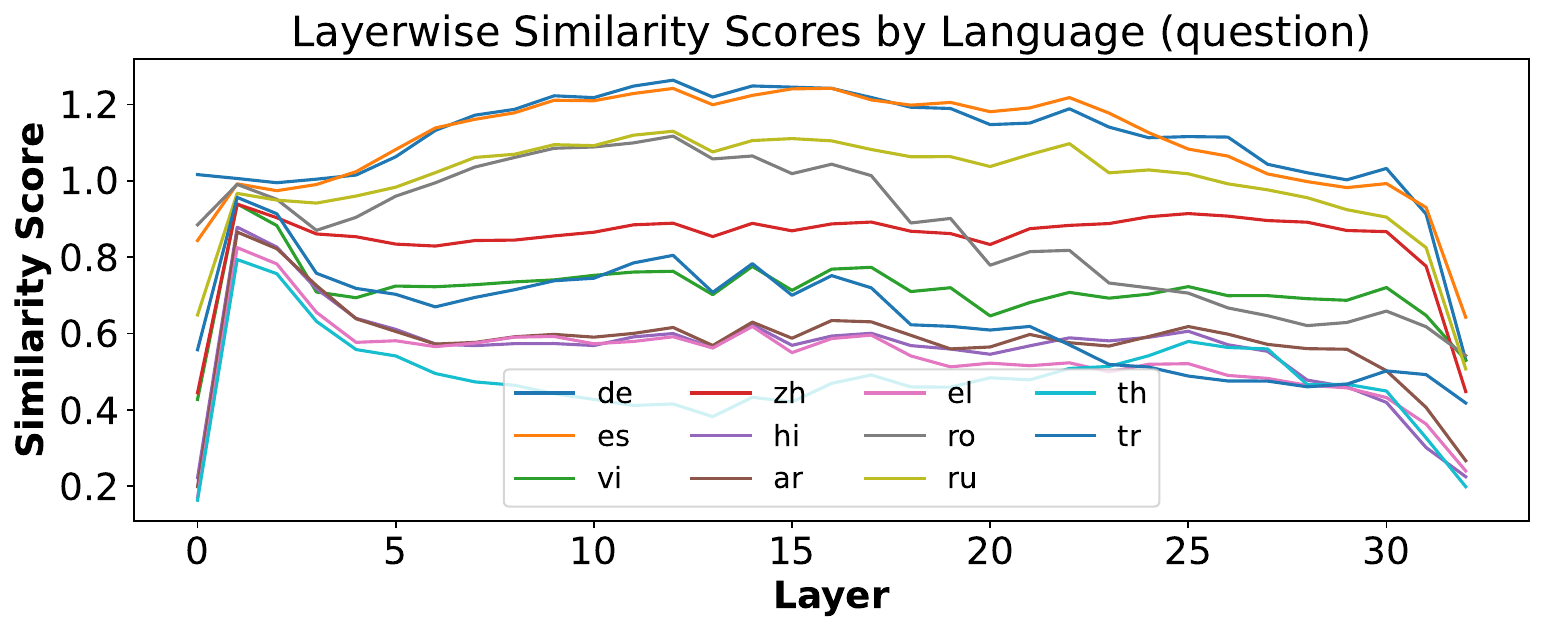}
        \caption{Question hidden state similarity for en-superior samples.}
    \end{subfigure}
    
    \vspace{0.5cm}
    
    \begin{subfigure}{0.45\textwidth}
        \includegraphics[width=\textwidth]{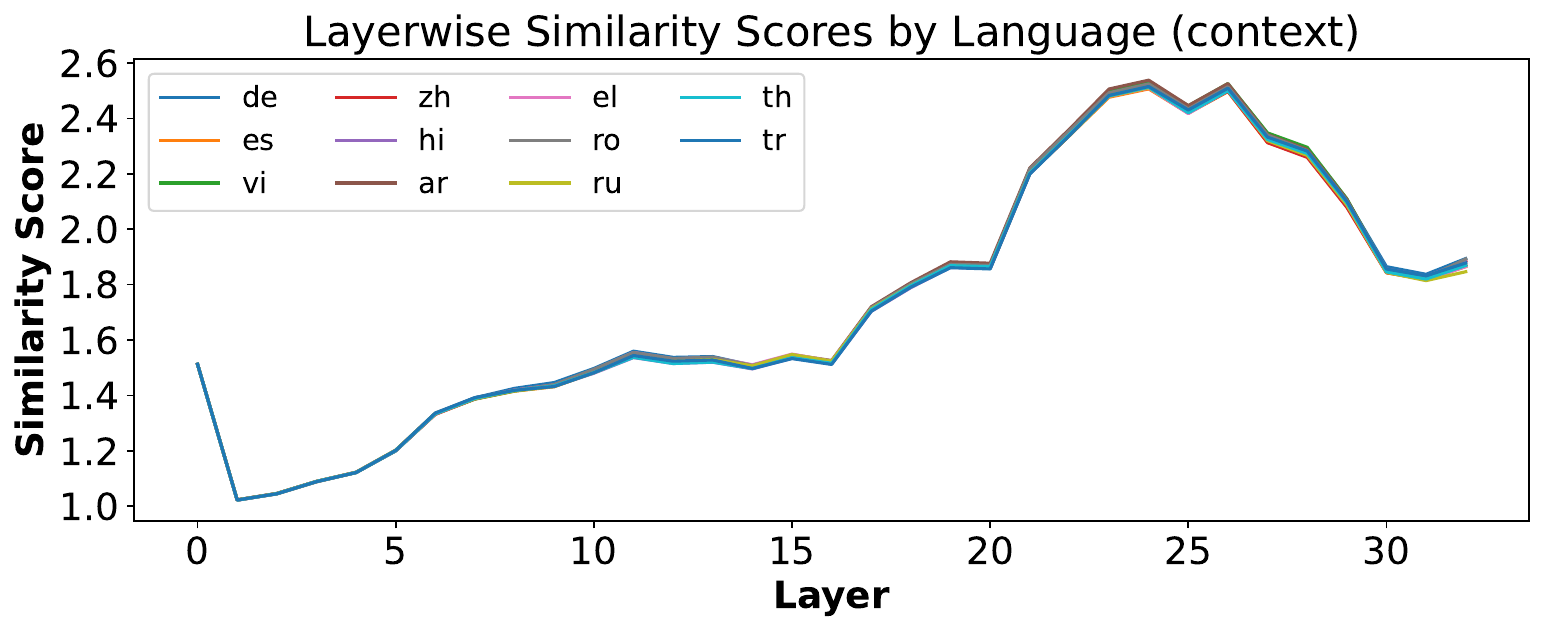}
        \caption{Context hidden state similarity for balanced samples.}
    \end{subfigure}
    \hfill
    \begin{subfigure}{0.45\textwidth}
        \includegraphics[width=\textwidth]{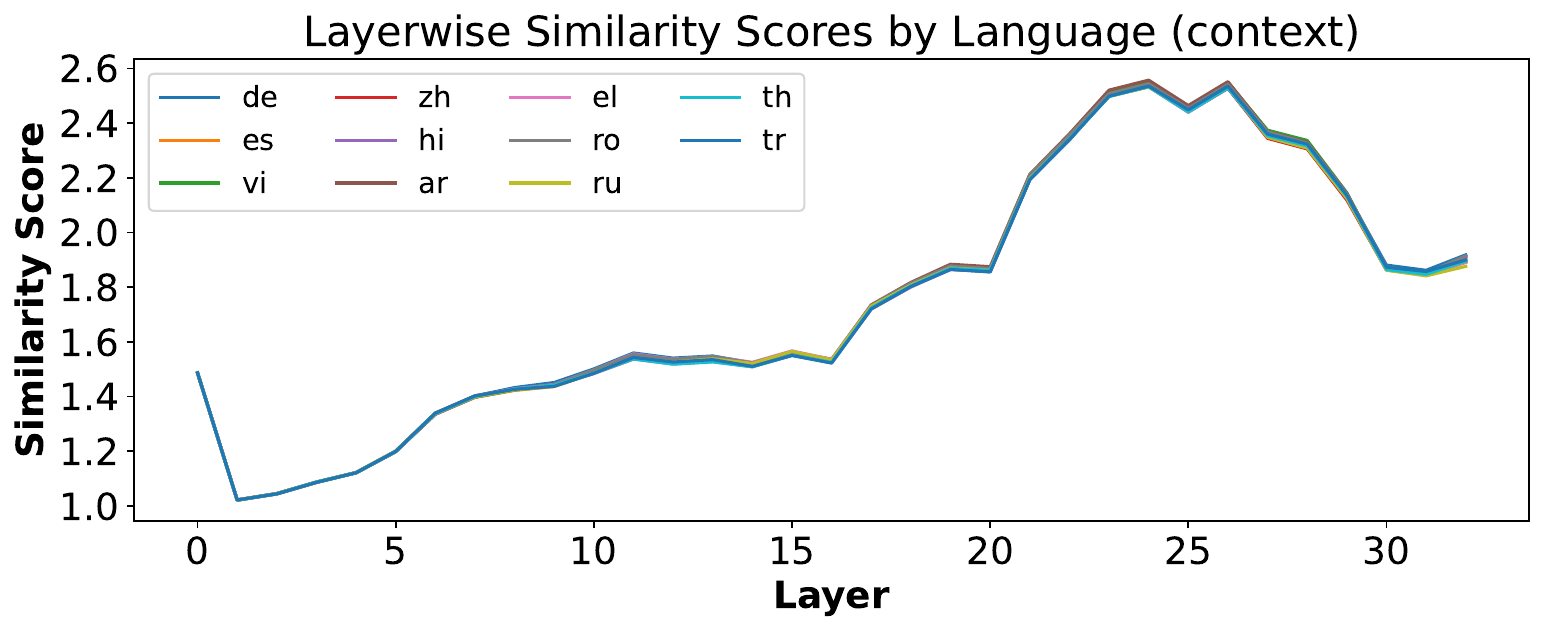}
        \caption{Context hidden state similarity for en-superior samples.}
    \end{subfigure}
    
    \vspace{0.5cm}
    
    \begin{subfigure}{0.45\textwidth}
        \includegraphics[width=\textwidth]{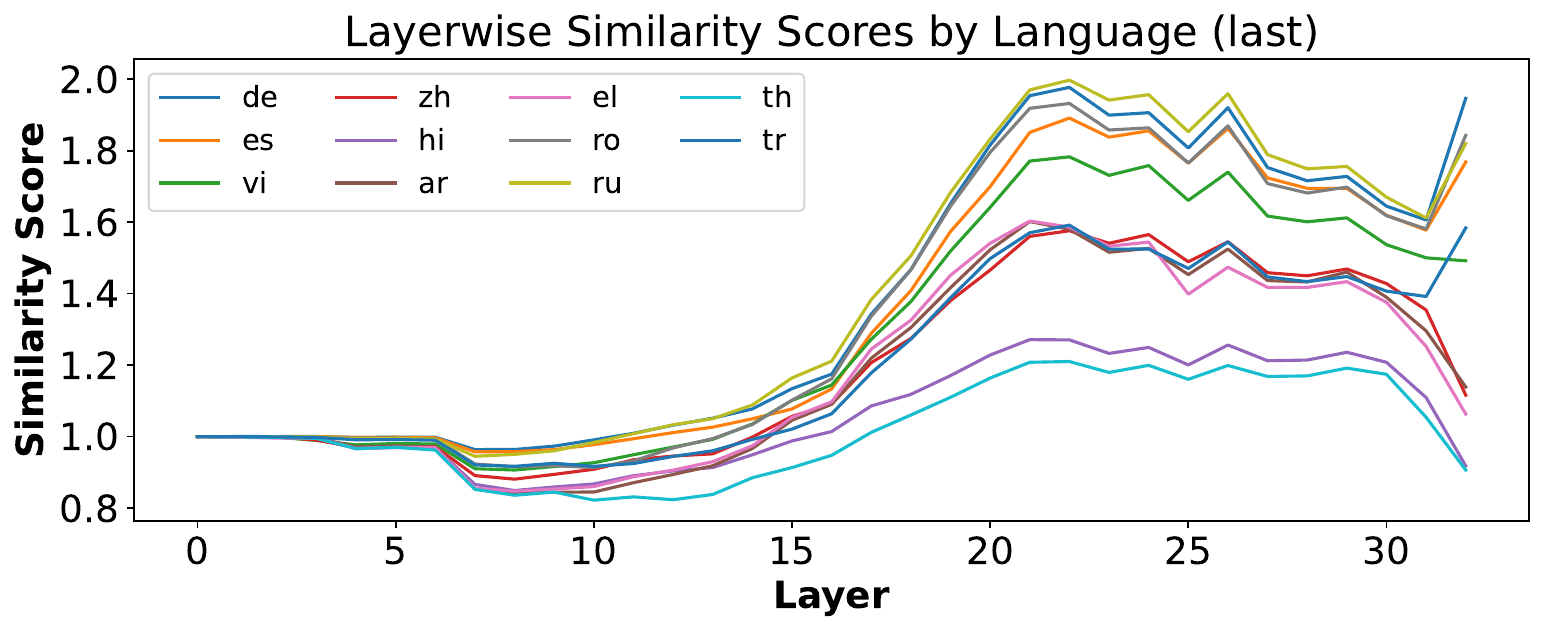}
        \caption{Last-input-token hidden state similarity for balanced samples.}
    \end{subfigure}
    \hfill
    \begin{subfigure}{0.45\textwidth}
        \includegraphics[width=\textwidth]{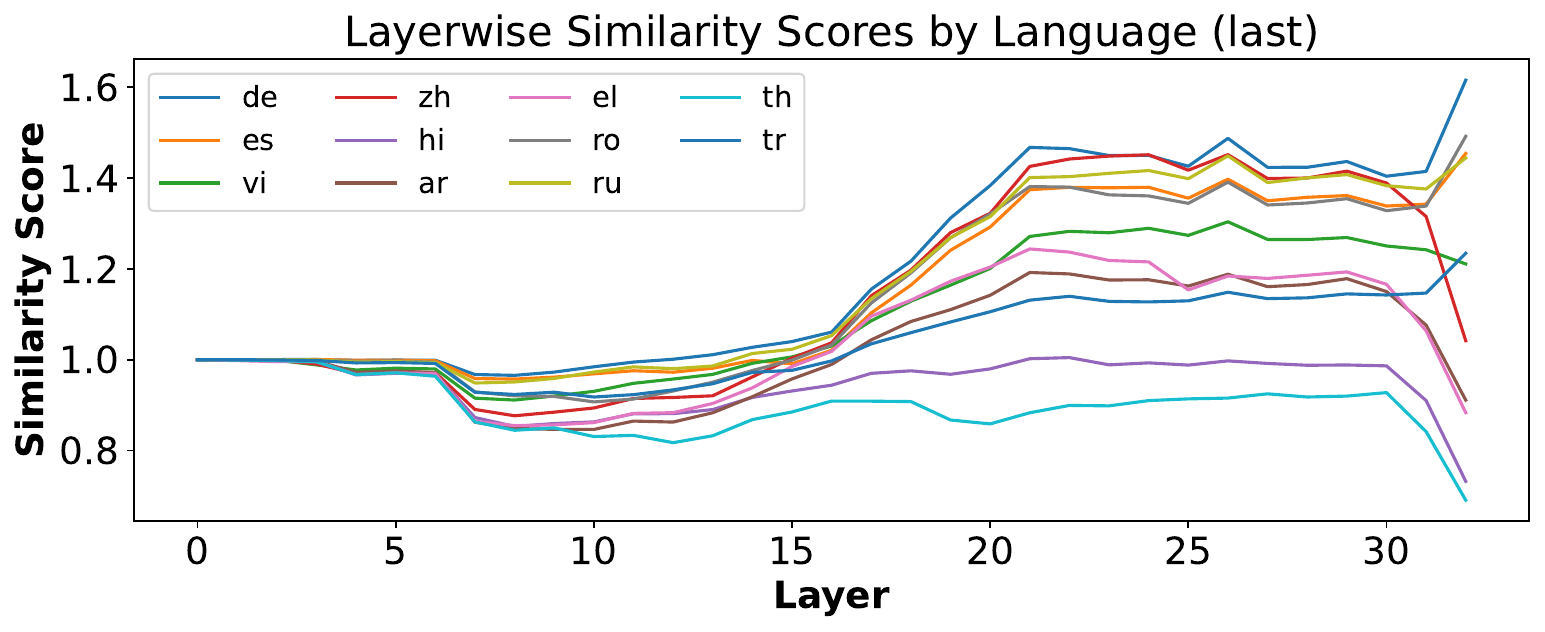}
        \caption{Last-input-token hidden state similarity for en-superior samples.}
    \end{subfigure}
    
    \caption{Hidden state similarity between English and other languages on different parts of the selected samples in each layer of the LLaMA-2-Chat-7B model.}
    \label{llama-2-chat-7b-2-shot-hidden-all}
\end{figure*}

Figures \ref{llama-3.1-base-8b-2-shot-hidden-all}–\ref{llama-2-chat-7b-2-shot-hidden-all} present the hidden state similarity results for additional LLaMA models, complementing the analysis of the LLaMA-3.1-Instruct-8B model discussed in the main body of the paper. 

\subsection{Training details and evaluation results of our finetuned LLaMA-3.1-8B}
\label{appendix:llama-tuned-params}
We tune the LLaMA-3.1-8B base model on TULU-V3 for 1 epoch with 8 * H800 GPUs for 15 hours using the LLaMA-Factory repository. The data cut-off length is 2048, batch size per device is 8, learning rate is 1.0e-5, and the warm-up ratio is 0.1 with cosine learning rate scheduling.

\begin{figure*}[ht]
    \centering
    \begin{subfigure}{0.48\textwidth}
        \includegraphics[width=\textwidth]{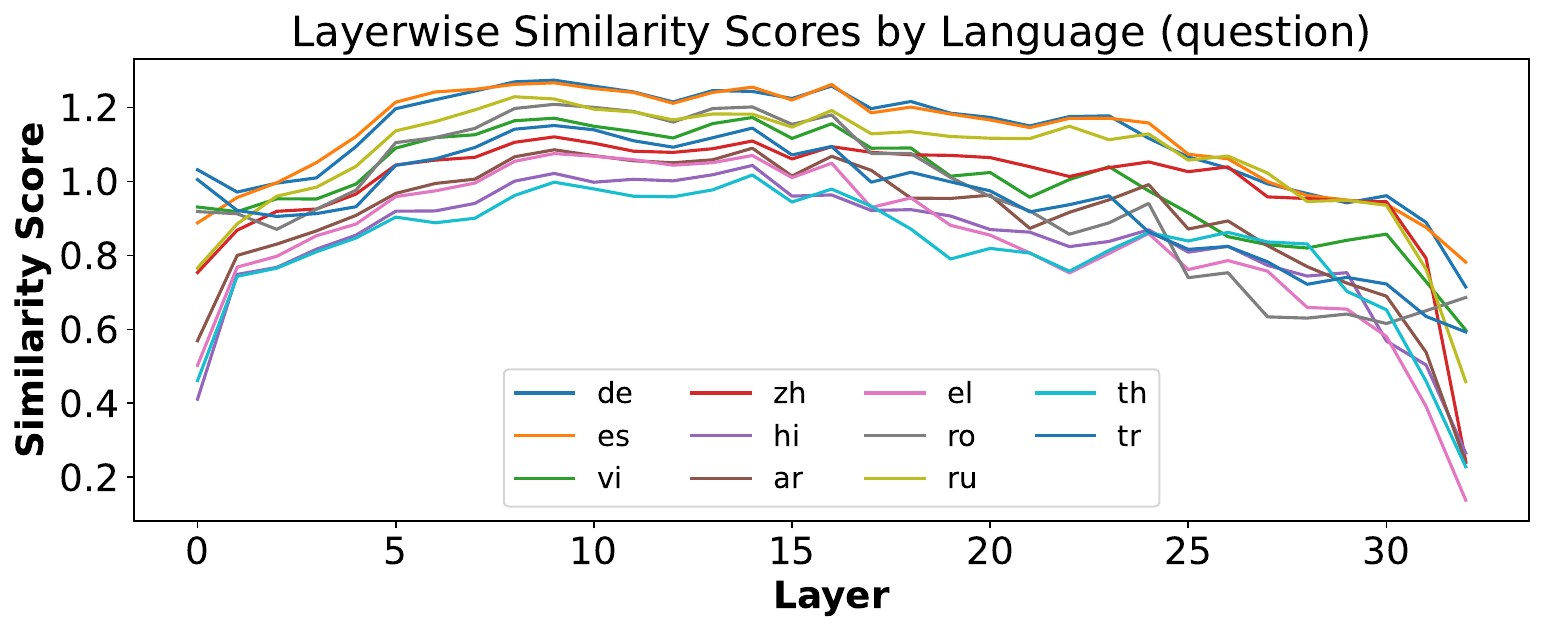}
        \caption{Question hidden state similarity.}
    \end{subfigure}
    \hfill
    \begin{subfigure}{0.48\textwidth}
        \includegraphics[width=\textwidth]{./assets/llama-3.1-tuned-8b-2-shot-balanced-c.pdf}
        \caption{Context hidden state similarity.}
    \end{subfigure}

    \vspace{0.5cm}

    \begin{subfigure}{0.48\textwidth}
        \centering
        \includegraphics[width=\textwidth]{./assets/llama-3.1-tuned-8b-2-shot-balanced-l.pdf}
        \caption{Last-input-token hidden state similarity.}
    \end{subfigure}

    \caption{Hidden state similarity between English and other languages on different parts of the balanced samples in each layer for our finetuned LLaMA-3.1-8B model.}
    \label{llama-3.1-tuned-8b-hidden-all}
\end{figure*}

Regarding evaluation, Table \ref{llama-3.1-tuned-8b-f1} summarizes the performance of our finetuned model on both en-x cross-lingual and x-x monolingual MRC tasks. Furthermore, Figure \ref{llama-3.1-tuned-8b-hidden-all} illustrates the hidden state similarity between English and other tested languages across layers, focusing on question, context, and last-input-token representations derived from balanced samples.

\begin{table*}[ht]
    \scriptsize
    \begin{center}
    \begin{subtable}[t]{\textwidth}
    \begin{tabular}{@{}cccclcccccccc@{}}
    \toprule
    {\textbf{}}                      & {\textbf{en-en}} & {\textbf{en-de}} & \textbf{en-es} & \textbf{en-vi}            & \textbf{en-zh} & \textbf{en-hi} & \textbf{en-ar} & \textbf{en-el} & \textbf{en-ro} & \textbf{en-ru} & \textbf{en-th} & \textbf{en-tr} \\ \midrule
    {\textbf{LLaMA-3.1-Tuned-8B}} & {78.80}          & {74.07}          & 69.85          & \multicolumn{1}{c}{72.12} & 71.03          & 69.23          & 67.57          & 69.35          & 71.54          & 71.86          & 71.12          & 71.02          \\ \bottomrule
    \end{tabular}
    \caption{en-x} 
    \end{subtable}
    \end{center}
    
    \begin{center}
    \begin{subtable}[t]{\textwidth}
    \begin{tabular}{cccccccccccc}
    \toprule
    {\textbf{}}                   & {\textbf{de-de}} & {\textbf{es-es}} & \textbf{vi-vi} & \textbf{zh-zh} & \textbf{hi-hi} & \textbf{ar-ar} & \textbf{el-el} & \textbf{ro-ro} & \textbf{ru-ru} & \textbf{th-th} & \textbf{tr-tr} \\ \midrule
    {\textbf{LLaMA-3.1-Tuned-8B}} & {69.28}          & {71.16}          & 73.73          & 63.71          & 67.48          & 64.35          & 61.84          & 73.03          & 65.66          & 62.34          & 62.63          \\ \bottomrule
    \end{tabular}
    \caption{x-x}
    \end{subtable}
    \end{center}
    \caption{2-shot F1 scores on en-x and x-x tasks for our finetuned LLaMA-3.1-8B.}
    \label{llama-3.1-tuned-8b-f1}
\end{table*}

\end{document}